\documentclass{article}

\usepackage{arxiv}

\usepackage[utf8]{inputenc} 
\usepackage[T1]{fontenc}    
\usepackage{hyperref}       
\usepackage{url}            
\usepackage{booktabs}       
\usepackage{amsfonts}       
\usepackage{nicefrac}       
\usepackage{microtype}      
\usepackage{graphicx}
\usepackage[super,sort&compress]{natbib}
\usepackage{doi}

\usepackage{amssymb}
\usepackage{amsmath}
\allowdisplaybreaks
\usepackage{booktabs}
\usepackage{multirow}
\usepackage{threeparttable}
\usepackage[ruled,vlined]{algorithm2e}
\usepackage{tcolorbox}
\usepackage{enumitem}
\tcbuselibrary{skins, breakable, fitting}
\usepackage[dvipsnames]{xcolor}
\usepackage{pifont}
\newcommand{\cmark}{\ding{51}}
\newcommand{\xmark}{\ding{55}}
\usepackage{fvextra}
\usepackage[T1]{fontenc}
\usepackage{courier}
\usepackage{ragged2e}
\usepackage[version=3]{mhchem}
\usepackage{rotating}
\usepackage{pdflscape}

\allowdisplaybreaks

\title{\emph{DASyR-LLM}: Domain-Aware Symbolic Regression with LLMs for Kinetic Model Discovery \thanks{Supplementary Information available: Detailed performance metrics for each case study, computational cost, confidence intervals for kinetic model parameters, extended table of related work, and detailed prompts used in this work. The complete source code repository is available at DOI: \href{https://doi.org/10.5281/zenodo.21793265}{\nolinkurl{10.5281/zenodo.21793265}}.}}

\author{\href{https://orcid.org/0009-0002-2589-6002}{\includegraphics[scale=0.06]{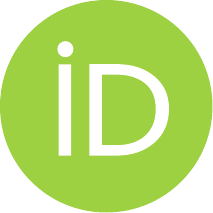}\hspace{1mm}Roberto Aliaga Medina}\\
    Department of Chemical Engineering\\ 
    Imperial College London\\
    South Kensington, London, SW7 2AZ, UK\\
    Department of Chemical Engineering, Biotechnology, and Materials\\
    University of Chile, Santiago, Chile\\
	\texttt{roberto.aliaga-medina25@imperial.ac.uk} \\
	\And
	\href{https://orcid.org/0000-0002-7717-0556}{\includegraphics[scale=0.06]{orcid.pdf}\hspace{1mm}Paulina Quintanilla} \\
	Department of Chemical Engineering\\
    University College London\\
    Gower Street, London, WC1E 6BT, UK\\
	\texttt{p.quintanilla@ucl.ac.uk} \\
	\And
    \href{https://orcid.org/0000-0003-0274-2852}{\includegraphics[scale=0.06]{orcid.pdf}\hspace{1mm}Antonio del Rio Chanona} \\
	Department of Chemical Engineering\\ 
    Imperial College London\\
    South Kensington, London, SW7 2AZ, UK\\
	\texttt{a.del-rio-chanona@imperial.ac.uk} \\
}

\date{}

\renewcommand{\headeright}{}
\renewcommand{\undertitle}{}
\renewcommand{\shorttitle}{LLM-guided Symbolic Regression for Kinetic Model Discovery}

\hypersetup{
pdftitle={LLM-Guided Symbolic Regression for Efficient and Interpretable Kinetic Model Discovery},
pdfsubject={LLM-Guided Symbolic Regression},
pdfauthor={Roberto Aliaga Medina, Paulina Quintanilla, Antonio del Rio Chanona},
pdfkeywords={LLM, symbolic regression, kinetic models},
}

\begin{document}
\maketitle

\begin{abstract}
	Kinetic model discovery is a central challenge in chemical engineering, as accurate rate expressions are essential for understanding and controlling chemical and biological processes. Symbolic regression (SR) has emerged as a powerful data-driven approach for identifying interpretable kinetic models, but usually operates without domain knowledge, often exploring physicochemically implausible models. Large language models (LLMs) offer a promising avenue for injecting domain expertise into this search.
    Here, we introduce an LLM-guided SR framework in which an LLM module is embedded within an iterative SR algorithm for automated kinetic model discovery. The LLM performs two roles at each iteration: (1) a qualitative physicochemical critique of the best SR candidates, and (2) the proposal of new candidate rate expressions guided by the SR-generated models and embedded chemical knowledge. Our framework is evaluated on four in silico case studies of increasing complexity, spanning heterogeneous catalysis and bioprocess systems.
    The results show that the LLM-guided framework reduces the number of iterations required to identify the ground-truth model by 41.7--79.3\% compared to a state-of-the-art SR framework, with the LLM directly proposing the correct model structure in more than half of the guided runs. In practical settings, where each iteration typically requires a new wet-lab experiment, this translates into a substantial reduction in experimental effort. Predictive performance on an independent validation dataset is equivalent between both approaches, with $R^2>0.98$ in all case studies. Ablation studies further indicate that both the SR component and the scale of the LLM contribute to this performance, with a reduced-size LLM largely retaining discovery efficiency. These findings demonstrate that LLMs can effectively inject domain knowledge into scientific model discovery, paving the way toward fully automated, domain-aware kinetic modelling pipelines.
\end{abstract}


\section{Introduction}

Mathematical modelling plays a central role in chemical engineering, enabling the analysis, optimisation, and control of complex chemical and biological systems. \cite{Rasmuson_Andersson_Olsson_Andersson_2014} Identifying dynamic kinetic models from experimental data remains a fundamental yet challenging task, due to nonlinear system behaviour, limited and noisy data, and the need to balance predictive accuracy with interpretability. \cite{D3DD00212H} This challenge is compounded by the cost and time typically required for data acquisition, \cite{Huan_2024} which makes efficient use of experimental resources a practical priority alongside identifying accurate model structures. Beyond fitting observations, model structures must reproduce the data while reflecting the underlying physical and chemical mechanisms of the system. Notably, this process is rarely a one-shot fitting exercise but an iterative loop in which candidate models inform the design of new experiments, making data acquisition an integral part of the discovery pipeline rather than a mere precursor to it. \cite{10.1039/d0re00232a}

Among data-driven model discovery approaches, symbolic regression (SR) has become a widely used framework due to its ability to recover explicit and interpretable mathematical expressions directly from data. \cite{cranmer2023interpretablemachinelearningscience, ROGERS2024120580, Makke2024} SR addresses the joint identification of functional form and parameters, most commonly through stochastic search strategies based on genetic programming (GP), \cite{devries2026symbolicdiscoverystochasticdifferential} which evolves candidate expressions through operations such as mutation and crossover. \cite{10.1007/978-3-031-82949-9_15} Despite their flexibility, SR and related approaches often suffer from excessively large search spaces, sensitivity to noise and initialisation, and limited mechanisms to enforce physically meaningful or domain-consistent structures. \cite{D3DD00212H,D5SC01473E,FORSTER2023108108} The simultaneous search for model structure and parameters is also combinatorially expensive, frequently leading to slow or unstable convergence. 
\cite{doi:10.1139/cgj-2025-0201} As a result, these methods may struggle to identify models that simultaneously accurate, interpretable, and physically plausible.

To mitigate these challenges, several approaches incorporate domain knowledge into data-driven discovery, either by constraining the model search to physically consistent structures \citep{WILSON2017785, doi:10.1021/acs.iecr.4c02981} or by guiding data acquisition through model-based design of experiments (MBDoE) as a core step within the discovery loop.\cite{D3DD00212H, ROGERS2024120580} Related efforts embed physical constraints directly into the SR search to narrow the effective search space and improve convergence, \citep{HAN2025734, keren2023physicsinformed, reinbold2021robust, ZHU20251, 10.1145/3679240.3734622} in some cases explicitly reducing the number of experiments required.\citep{servia2026physicsinformedsymbolicregressiondataefficient} However, these approaches typically rely on predefined structures or explicit assumptions about model form, limiting their flexibility when prior knowledge is incomplete or uncertain.

More recently, large language models (LLMs) have emerged as a promising tool for scientific discovery, \cite{10.1145/3735634} owing to two key properties. First, their strong reasoning capabilities and embedded scientific priors allow them to incorporate domain knowledge when proposing or refining candidate expressions, potentially guiding the search toward more meaningful model structures. \cite{Merler_2024, guo2025srllm, shojaee2025llmsrbench} Second, their generative nature enables the construction of novel candidates from previously explored solutions, which can help escape local optima without restarting the discovery process. \cite{bideh2026llmode, zhang2026llmmetasrincontextlearningevolving, pourcel2026selfimprovinglanguagemodelsevolutionary, gozeten2026evolutionarymultitaskoptimizationllmguided}

In this work, we propose DASyR-LLM, a framework for LLM-guided SR in which scientific reasoning is embedded into the model discovery loop. While agnostic to the underlying SR strategy, our framework builds on ADoK-S, developed by de Carvalho Servia \textit{et al.}, \cite{D3DD00212H} as the underlying SR backbone. We integrate an LLM module that guides the search toward physically meaningful candidate models, analysing high-performing candidate models and generating new structurally informed hypotheses, which are then evaluated within the same selection and experimental design pipeline.

This reduces the effective search space by embedding domain reasoning into the SR loop, while preserving the flexibility required to discover novel model structures from noisy data. Rather than relying on LLMs solely as equation generators or optimisation aids, we leverage them as knowledge-guided components that bias the search toward scientifically consistent models, with the ultimate goal of reducing the experimental effort required to arrive at validated kinetic models. To quantify the efficiency gains introduced by the LLM module, each case study is evaluated under two configurations: a baseline SR-only workflow, and the proposed LLM-guided framework (DASyR-LLM), run under identical experimental budgets for direct comparison

The rest of the article is organised as follows: Section \ref{sec:related} provides a review of existing model discovery frameworks and recent applications of LLMs in scientific model discovery; Section \ref{sec:methods} presents the proposed methodology; Section \ref{sec:cases} introduces the case studies used to evaluate the approach; Section \ref{sec:results} presents the results and discusses the performance of the proposed framework in terms of model quality, interpretability, and discovery efficiency; Section \ref{sec:ablation} provides ablation studies designed to assess the contribution of individual components of the framework; and, finally, Section \ref{sec:conclusions} summarises the main conclusions and outlines potential directions for future research.

\section{Related Work}
\label{sec:related}

\subsection{Symbolic Regression and Sparse Identification Methods}

Model discovery has been addressed through a variety of data-driven approaches. \citet{cranmer2023interpretablemachinelearningscience} developed PySR, an SR tool that searches for closed-form expressions via GP. Its main strength is interpretability, but the search is guided only by statistical fit, without domain knowledge to constrain candidate structures. \citet{NEUMANN2020123412} proposed a new SR formulation tailored to identifying physico-chemical laws from experimental data, though the search remains guided purely by statistical fit without embedded domain reasoning.

\citet{doi:10.1073/pnas.1517384113} introduced SINDy, which recovers governing equations via sparse regression over a library of candidate functions using sequential thresholding to identify the active terms, offering an efficient alternative to evolutionary SR. However, it does not accommodate nonlinear parameters and it is highly sensitive to noise in the estimated derivatives, particularly for reaction kinetics data. To mitigate this, \citet{doi:10.1021/acs.iecr.4c02981} introduced a derivative-free variant that explicitly incorporates domain information such as mass balances and chemistry constraints, though it still identifies a single global model per run, while \citet{LYU2025109265} coupled SINDy with iterative experimental design, identifiability analysis, and AIC-based selection (DoE-SINDy). 

In parallel, \citet{GUSMAO2023113701} proposed kinetics-informed neural networks (KINNs), embedding equality and inequality constraints to encode domain knowledge in forward (model construction) or inverse (parameter learning) mode, scaling well to complex reaction networks.

\subsection{Domain-Specific Frameworks for Kinetic Model Discovery}

\citet{WILSON2017785} introduced ALAMO, learning algebraic models via mixed-integer nonlinear programming with first-principles output constraints and error-maximisation sampling. Its strength is explicit constraint-handling, though its expressiveness is limited to pre-specified functional forms. \citet{SUN2020107103} proposed ALVEN, combining chemically/biologically motivated nonlinear features with sparse (elastic net) regression, achieving an efficient and interpretable approach.

More directly related to the present work, \citet{D3DD00212H} integrate SR with parameter estimation and AIC-based selection within an MBDoE-driven loop (ADoK). Its main limitation, directly addressed here, is that model selection relies purely on statistical criteria, without qualitative assessment of chemical plausibility. Building on this backbone, \citet{servia2026physicsinformedsymbolicregressiondataefficient} proposed PI-ADoK, which integrates physical constraints directly into the SR search to narrow the search space and reduce the number of experiments required for convergence, while incorporating Metropolis-Hastings-based uncertainty quantification. However, generation and selection remain guided by predefined physical constraints and statistical criteria, without an LLM to reason qualitatively about candidate plausibility. This SR-MBDoE structure was extended by Rogers \textit{et al.},\cite{ROGERS2024120580} coupling SR with MBDoE to jointly discriminate models and optimise a process flow diagram for formulated products, later enhanced via neural-network feature attribution. \citep{ROGERS2025109036} Both reduce experimental effort effectively for formulated product manufacturing, though neither is applied to reaction kinetics. Closest in domain, \citet{doi:10.1021/acs.jcim.5c03032} applied substructure-decomposed SR with MBDoE to constrain the search to mechanistically meaningful rate expressions, evaluated on methanol synthesis and an enzymatic system, discussing LLM-based augmented intelligence as future work. \citet{D5SC01473E} proposed SiMBA, targeting the identification of complete reaction mechanisms directly from data. This addresses a complementary but distinct problem to the one considered here: recovering the reaction network structure itself, rather than the rate expression for a known (assumed) network structure.

\subsection{Hybrid and Complementary Applications of SR}

\citet{NARAYANAN2022133032} proposed the Functional-Hybrid model, combining ranked domain-specific functional beliefs with SR to build interpretable hybrid dynamic models, validated across chemical, biochemical, ecological, and bioreactor case studies. Compared to a conventional ANN-based hybrid model, it achieves similar interpolation accuracy while offering interpretability, better performance with scarce data, and superior extrapolation. \citet{DICAPRIO2026122873} proposed SINDybrid, an MILP approach identifying epistemic uncertainty in a mechanistic backbone and compensating it with data-driven components. Related applications include Bayesian SR for surrogate-based flowsheet optimisation \citep{JOG2024108563} and SR-based process control. \citep{LIMA2025110350, LIMA2026103700} Complementing these, \citet{Tabrizi2025100276} developed MIDDoE, a software framework lowering the barrier to MBDoE adoption. More recently, \citet{ROSSI2026109634} introduced HyMech, which extends the SR search space with first-principles-derived function and variable pools to diagnose and correct process-model mismatch, embedding domain knowledge statically rather than through iterative qualitative reasoning, and without an MBDoE-driven data acquisition loop.

\subsection{LLMs in Scientific Model Discovery}

Recent work has explored several roles for LLMs, grouped here into four categories based by function. A first group employs LLMs as evolutionary operators: \citet{zhang2026llmmetasrincontextlearningevolving} use an LLM to design selection operators for an evolutionary SR algorithm, embedding domain knowledge directly into the generating prompt; \citet{du2024llm4edlargelanguagemodels} use the LLM as optimiser and evolutionary operator; \citet{bideh2026llmode} guide evolution using patterns from elite candidates. All three report efficiency gains over classical GP, but are evaluated on generic benchmarks rather than reaction kinetics, with purely quantitative selection.

A second group uses LLMs as generators of equation/code skeletons. \citet{shojaee2025llmsrscientificequationdiscovery} introduced LLM-SR, representing equations as executable programs optimised against data, drawing on broad scientific priors rather than domain-specific chemical reasoning. \citet{shen2026llmdmdlargelanguagemodelbased} adapted this to enforce algebraic constraints in power system dynamics (LLM-DMD), while \citet{ivanchik2025does} used the LLM as an oracle suggesting plausible PDE forms within an evolutionary framework. In all three, acceptance is governed purely by compilation success and fit to data, with no standalone reasoning about mechanistic plausibility.

A third group embeds LLMs within broader iterative discovery-calibration frameworks. \citet{NEURIPS2024_aea8bdc4} (G-Sim) combine LLM-driven structural design with likelihood-free calibration for simulator construction, and \citet{https://doi.org/10.1111/2041-210x.70244} (LEMMA) combine LLM-based equation synthesis with evolutionary optimisation for ecosystem modelling. Both achieve strong performance, but rely on statistical calibration alone to accept LLM-proposed structures, without an explicit plausibility check.\citet{wahl2026probabilisticframeworkllmbasedmodel} proposed ModelSMC, framing model discovery as probabilistic inference that weights LLM-proposed candidates by likelihood within a Sequential Monte Carlo scheme. This offers a principled alternative to heuristic LLM-agent workflows, though weighting remains purely statistical.

\begin{table*}[!h]
\centering
\footnotesize
\setlength{\tabcolsep}{4pt}
\renewcommand{\arraystretch}{0.1}
\begin{threeparttable}
\caption{Comparison of representative model/equation discovery approaches. \textbf{Mech.\ eval.}: incorporates qualitative mechanistic/physicochemical plausibility assessment of candidate models; \textbf{MBDoE}: uses model-based design of experiments to guide data acquisition; \textbf{Iter.}: model discovery proceeds through an iterative loop (as opposed to a single-shot fit); \textbf{Kinetics}: applied to reaction kinetics case studies; \textbf{LLM}: incorporates a large language model within the discovery pipeline}
\label{tab:related_work}
\begin{tabular}{@{}p{3.2cm}p{3.7cm}p{2.9cm}p{1.0cm}p{1.0cm}p{0.8cm}p{1.3cm}p{0.7cm}@{}}
\toprule
\textbf{Reference} & \textbf{Approach type} & \textbf{Domain knowledge} & \textbf{Mech. eval.} & \textbf{MBDoE} & \textbf{Iter.} & \textbf{Kinetics} & \textbf{LLM} \\
\midrule

SINDy; \cite{doi:10.1073/pnas.1517384113} DF-SINDy; \cite{doi:10.1021/acs.iecr.4c02981} DoE-SINDy \cite{LYU2025109265} &
Sparse regression over function library &
Implicit (library); explicit in DF-SINDy &
\xmark &
\xmark / \cmark\tnote{\textit{a}} &
\xmark / \cmark\tnote{\textit{a}} &
\xmark / \cmark\tnote{\textit{b}} &
\xmark \\
\addlinespace

KINNs \cite{GUSMAO2023113701} &
Physics-informed NNs for kinetics &
Yes, equality/inequality constraints &
\xmark &
\xmark &
\xmark &
\cmark &
\xmark \\
\addlinespace

ALAMO \cite{WILSON2017785} &
MINLP-based algebraic model learning &
Yes, first-principles constraints &
\xmark &
\cmark &
\cmark &
\cmark &
\xmark \\
\addlinespace

ALVEN \cite{SUN2020107103} &
Nonlinear features + elastic net &
Yes, predefined chem./bio.\ features &
\xmark &
\xmark &
\xmark &
Partial &
\xmark \\
\addlinespace

ADoK \cite{D3DD00212H} &
SR + parameter est.\ + AIC selection &
Minimal (baseline) &
\xmark &
\cmark &
\cmark &
\cmark &
\xmark \\
\addlinespace

SR-MbDoE (PFD); \cite{ROGERS2024120580, ROGERS2025109036} SR+MBDoE (kinetics) \cite{doi:10.1021/acs.jcim.5c03032} &
SR proposes candidates; MBDoE discriminates iteratively &
Yes, structural constraints &
\xmark &
\cmark &
\cmark &
\xmark / \cmark\tnote{\textit{c}} &
\xmark \\
\addlinespace

\midrule
\multicolumn{8}{l}{\textit{LLM-guided approaches}} \\
\addlinespace

LLM-Meta-SR; \cite{zhang2026llmmetasrincontextlearningevolving} LLM4ED; \cite{du2024llm4edlargelanguagemodels} LLM-ODE \cite{bideh2026llmode} &
LLM-guided evolutionary operators &
Generic, prompt-embedded &
\xmark &
\xmark &
\cmark &
\xmark &
\cmark \\
\addlinespace

LLM-SR; \cite{shojaee2025llmsrscientificequationdiscovery} LLM-DMD; \cite{shen2026llmdmdlargelanguagemodelbased} LLM-guided PDE \cite{ivanchik2025does} &
LLM generates equation/code skeletons &
Broad priors / none / plausibility oracle &
\xmark &
\xmark &
\cmark &
\xmark &
\cmark \\
\addlinespace

G-Sim; \cite{NEURIPS2024_aea8bdc4} LEMMA \cite{https://doi.org/10.1111/2041-210x.70244} &
LLM proposes structure; empirical calibration &
Yes, domain priors (causal / RAG-based) &
\xmark &
\xmark &
\cmark &
\xmark &
\cmark \\
\addlinespace

ModelSMC \cite{wahl2026probabilisticframeworkllmbasedmodel} &
Probabilistic inference (SMC) &
Implicit via LLM priors &
\xmark &
\xmark &
\cmark &
\xmark &
\cmark \\
\addlinespace

PiSR; \cite{Taskin2026} LaSR; \cite{grayeli2024symbolicregressionlearnedconcept} IGSR \cite{saveliev2026influenceguidedsymbolicregressionscientific} &
LLM as evaluator (loss/concept/MCTS)\tnote{\textit{d}} &
Generic/statistical criteria &
Partial &
\xmark &
\cmark &
\xmark &
\cmark \\
\addlinespace

\midrule
\textbf{This work} &
LLM embedded in SR--param.\ est.--AIC loop: generates and assesses candidates &
Yes, explicit physicochemical reasoning &
\cmark &
\cmark &
\cmark &
\cmark &
\cmark \\
\bottomrule
\end{tabular}
\begin{tablenotes}
\footnotesize
\item[\textit{a}] Applies to DoE-SINDy only.
\item[\textit{b}] DF-SINDy and DoE-SINDy applied to reaction kinetics; original SINDy is not domain-specific.
\item[\textit{c}] PFD: Rogers et al.\ couple SR with MBDoE to optimise a process flow diagram for formulated products. Kinetics: Kay \& Zhang apply substructure-decomposed SR with MBDoE to reaction kinetics (methanol synthesis and an enzymatic system).
\item[\textit{d}] MCTS: Monte Carlo Tree Search, used by IGSR for per-term influence-guided pruning of candidate expressions.
\end{tablenotes}
\end{threeparttable}
\end{table*}

A final group uses LLMs purely as evaluators. \citet{Taskin2026} integrates an LLM-based score (dimensional consistency, simplicity, physical realism) into the SR loss function. To our knowledge, this is the closest precedent to a qualitative plausibility check, but the criteria are fixed, domain-generic, and compressed into a single scalar, and the LLM does not propose candidates. \citet{grayeli2024symbolicregressionlearnedconcept} (LaSR) and \citet{saveliev2026influenceguidedsymbolicregressionscientific} (IGSR) instead guide search via abstracted concepts and per-term influence scores, enriching the feedback signal but without qualitative mechanistic reasoning.

\subsection{Summary and Research Gap}

Table~\ref{tab:related_work} summarises the main frameworks discussed above along the dimensions most relevant to this work (see \hyperref[ESI]{\textbf{Supplementary Information}\textsuperscript{$\*$}} for a more detailed comparison). Here, two patterns emerge. First, among frameworks combining SR with MBDoE, \cite{D3DD00212H,ROGERS2024120580,ROGERS2025109036,doi:10.1021/acs.jcim.5c03032} none incorporate an LLM to reason about candidate plausibility; selection relies solely on statistical criteria such as AIC. Second, among LLM-guided approaches, the LLM's role is limited to either candidate \emph{generation} (selected via purely numerical criteria) or candidate \emph{evaluation} (as a scalar term), but never both, and never combined with an MBDoE-driven loop. This gap aligns with recent perspectives in the catalysis kinetics literature, which envision AI-assisted, multimodal closed-loop model building as a key future direction. \citep{RANGARAJAN2026101240} The present work addresses this gap by combining both roles within an MBDoE-driven iterative loop applied specifically to reaction kinetics.

\section{Methodology}
\label{sec:methods}

The proposed approach uses ADoK-S as its underlying SR backbone, which combines SR, parameter estimation, model selection, and MBDoE to progressively refine candidate models using experimental data. ADoK-S is the strong-formulation variant of the Automated Discovery of Kinetic models (ADoK) framework, which estimates rate measurements from concentration data rather than bypassing this step as in the weak formulation (ADoK-W). \cite{D3DD00212H} Leveraging this framework, we embed an LLM module that guides the proposal of candidate models using domain-informed reasoning, as a core step within the same iterative loop.

\begin{figure}[h]
 \centering
 \includegraphics[height=8cm]{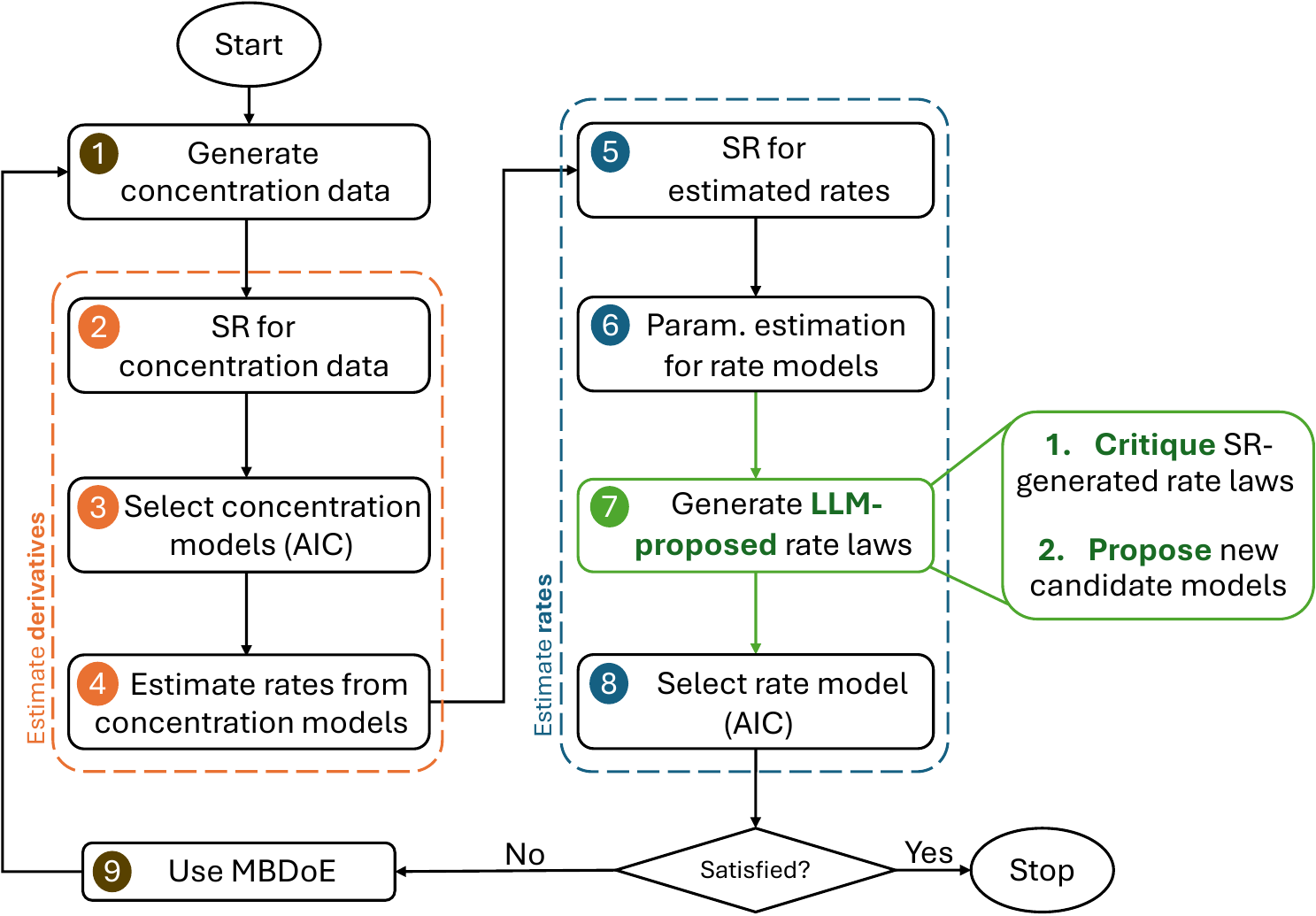}
\caption{Schematic of the LLM-guided SR workflow (DASyR-LLM), comprising nine steps. The LLM module (step 7) critiques the best SR candidates and proposes new rate expressions to guide the iterative model discovery process.}
 \label{general_workflow}
\end{figure}

Figure \ref{general_workflow} shows the proposed framework, comprising nine steps. It begins with the generation of candidate model structures via SR (steps 1-5), followed by parameter estimation (step 6). The LLM (step 7) then critiques the best-performing models at each iteration and proposes new candidate equations based on their structure, parameter values, and performance metrics. These LLM-proposed models are combined with those from SR and subjected to the same evaluation and selection procedure using statistical criteria (step 8), ensuring consistency within the framework. The selected models then guide the MBDoE step (step 9), which generates new experimental data. The framework iterates in this manner until the underlying model is recovered or the experimental budget is exhausted.

\subsection{Proposed Framework}

Building on ADoK-S, our framework relies on five main components: SR, parameter estimation, the LLM module, model selection (AIC), and MBDoE, integrated within the same iterative loop. The objective is to identify a symbolic model that maps system states (\textit{i.e.}, concentrations) to their corresponding time derivatives. Since direct measurements of the rates are unavailable, they are approximated from concentration data through an intermediate symbolic model fitted via SR, whose time derivatives provide rate estimates. This yields a nested structure, with model selection at the outer level and parameter estimation at the inner level, embedded within an iterative MBDoE loop that progressively enriches the dataset. 

This modular structure keeps each stage self-contained: the LLM module operates on the output of parameter estimation and feeds directly into model selection, without requiring changes to the underlying SR or experimental design procedures. Within this loop, the initial, data-driven generation of candidate models (either for concentrations or rates) is handled by SR, which constructs functional expressions $m \in \mathcal{M}$ through iterative composition of a predefined set of operators and elementary building blocks (\textit{i.e.}, input variables and constants), where $\mathcal{M}$ denotes the symbolic search space. The LLM subsequently expands this candidate set with knowledge-guided proposals (Section~\ref{sec:llm_module}). Our workflow consists of the following steps, numbered as in Figure~\ref{general_workflow}:

\begin{enumerate}
    \item Generation of concentration data from ground-truth system (Section~\ref{datagen}).
    
    \item Application of SR to identify candidate concentration models from the generated data (Section~\ref{section:SR}).

    \item Selection of the best concentration model for each species and experiment, based on AIC calculation (Section~\ref{modelsec}).
    
    \item Estimation of time derivatives from concentration profiles (Section~\ref{section:SR}).
    
    \item Application of SR to the estimated derivatives in order to identify candidate rate expressions (Section~\ref{section:SR}).

    \item Refinement of parameters for SR-generated rate models, using the values internally estimated by SR as initial guesses, through ODE integration (Section~\ref{paramest}).

    \item Critique of the best-performing rate models and proposal of new candidate expressions by the LLM, followed by parameter initialisation and then refinement as in step 6 (Section~\ref{sec:llm_module}).
    
    \item Calculation of the AIC and ranking of candidate models based on simulations obtained through ODE integration against the original concentration data (Section~\ref{modelsec}).
\end{enumerate}

If the ground-truth model is not identified after step 8, a ninth MBDoE step is performed to generate an additional experiment (Section~\ref{mbdoe}), which is incorporated into the experimental data set. These steps are repeated iteratively until the ground-truth model is found or the experimental budget (12 iterations, corresponding to 11 additional experiments) is fully spent. We now analyse the individual components of our framework:

\subsubsection{Data Generation.}
\label{datagen}

As no physical experiments were available for this study, synthetic data were generated by simulating a known ground-truth model for each case study. In the kinetic systems considered in this work, the system state is represented by the vector of species concentrations $\mathbf{C}(t)\in\mathbb{R}^{n_s}$, where $n_s$ is the total number of species. The system dynamics are described by
\begin{equation}
\dot{\mathbf{C}}(t)=\mathbf{f}(\mathbf{C}(t)), \qquad \mathbf{C}(t_0)=\mathbf{C}_0,
\end{equation}
where $\dot{\mathbf{C}}(t)$ denotes the vector of concentration time derivatives over the interval $t\in[t_0,t_f]$, and $\mathbf{f}$ maps the system state to these derivatives through the (unknown) ground-truth reaction rate and the stoichiometry of the system. The available dataset consists of $n_t$ sampled concentration measurements,
\begin{equation}
\mathcal{D} = \left\{ \left(t^{(i)},\mathbf{C}^{(i)}\right)\right\}_{i=1}^{n_t},
\end{equation}
where $\mathbf{C}^{(i)}$ denotes the noisy measured concentration vector at time $t^{(i)}$, obtained by adding measurement noise to the ground-truth trajectory simulated for each case study ($n_t$ grows over successive iterations as new experiments are incorporated via MBDoE, allowing candidate models to be refined as new information becomes available). These measurements constitute the input data used by the SR modules within our framework. For each case study, the initial dataset was generated by simulating the ground-truth model under two distinct initial conditions, representing two independent experiments. These initial conditions were randomly selected from a $2^k$ factorial design. \cite{mee2009} Gaussian noise with zero mean and standard deviation $\sigma = 0.1$ was subsequently added to all simulated concentration measurements throughout the discovery process, in order to emulate experimental uncertainty.

\subsubsection{Symbolic Regression Step.}
\label{section:SR}

\begin{figure*}[htbp]
    \centering
    \includegraphics[width=\linewidth]{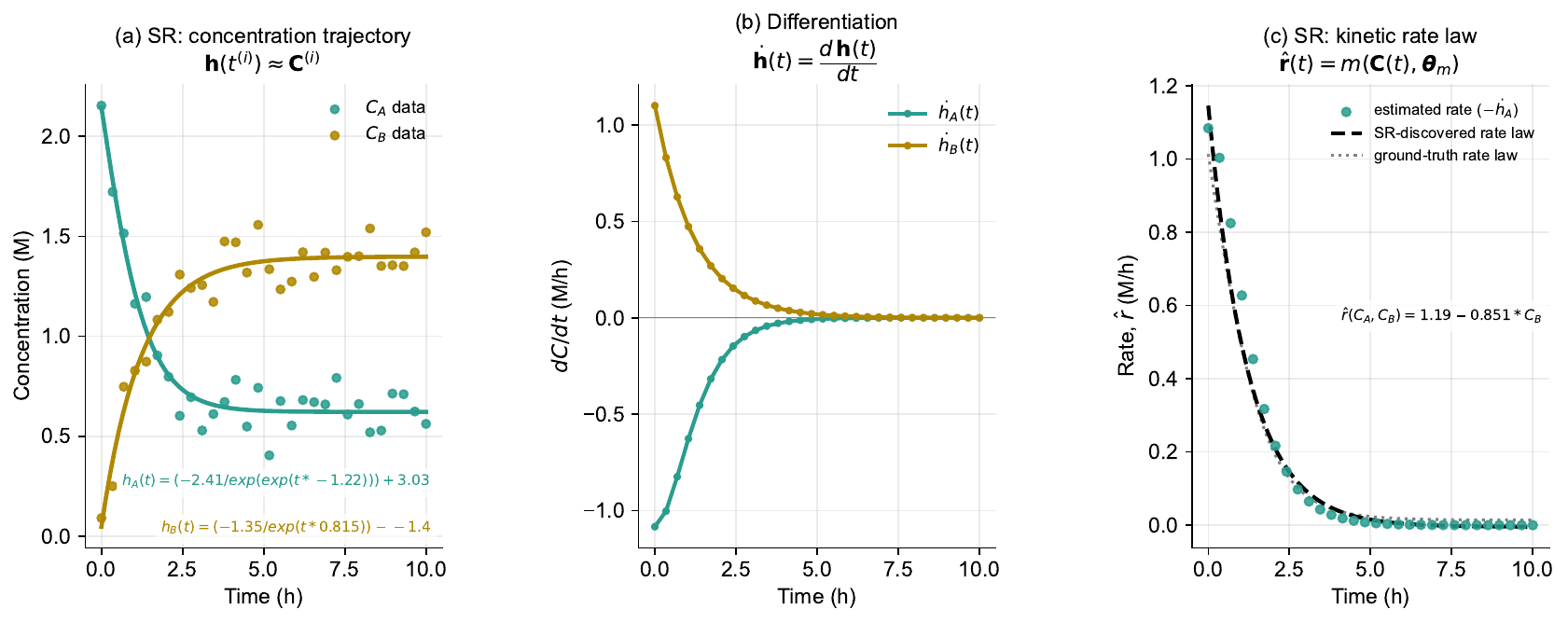}
    \caption{Illustration of the two-stage SR workflow using data from a single experiment (theoretical isomerisation case study). (a) SR fits an analytical trajectory $\mathbf{h}(t)$ to the noisy concentration measurements $\mathbf{C}^{(i)}$ for each species. (b) The fitted trajectories are differentiated to obtain estimates of the reaction rate, $\dot{\mathbf{h}}(t) = d\mathbf{h}(t)/dt$. (c) SR is applied using these estimated derivatives as the regression target to identify a candidate kinetic rate expression $\hat{\mathbf{r}}(t) = m(\mathbf{C}(t),\boldsymbol{\theta}_m)$, shown alongside the ground-truth rate law for reference. Note that this figure illustrates a single iteration of the workflow using a single experiment, for clarity. The converged rate law reported in Results (Section~\ref{sec:results}) is obtained after several iterations, incorporating additional MBDoE-generated experiments.}
    \label{fig:sr_pipeline}
\end{figure*}

SR was performed using \texttt{PySRRegressor}, an evolutionary SR framework developed by Cranmer. \cite{cranmer2023interpretablemachinelearningscience} Within our workflow, SR is applied at two different stages:

First, for each species, SR is used to identify an analytical approximation of the concentration trajectory:
\begin{equation}
\mathbf{h}(t^{(i)}) \approx \mathbf{C}^{(i)},
\end{equation}
where $\mathbf{h}$ is a vector of symbolic expressions, one per species, fitted so that its evaluation at each sampled time approximates the corresponding concentration measurement (illustrated in Figure~\ref{fig:sr_pipeline} using data from a single experiment). For each species, \texttt{PySRRegressor} returns a Pareto front of candidate expressions spanning different levels of complexity and accuracy, from which the best-fitting model is selected using the AIC (Section~\ref{modelsec}). Since $\mathbf{h}$ is differentiable, estimates of the corresponding time derivatives are obtained as
\begin{equation}
\dot{\mathbf{h}}(t) = \frac{d \mathbf{h}(t)}{dt},
\end{equation}
computed using \texttt{numdifftools.Derivative}. \cite{numdifftools} At this stage, the model search is carried out with a moderate computational budget and a standard set of arithmetic operators ($+$, $-$, $\times$, $\div$, $\exp$), with the objective of obtaining smooth and interpretable representations of the observed dynamics. 

Second, SR is applied using the estimated derivatives $\dot{\mathbf{h}}(t^{(i)})$ from the previous step as the regression target, and the measured concentrations $\mathbf{C}^{(i)}$ as input variables, in order to identify a candidate kinetic rate expression $\hat{\mathbf{r}}(t^{(i)}) \approx \dot{\mathbf{h}}(t^{(i)})$ of the form
\begin{equation}
\hat{\mathbf{r}}(t) = m(\mathbf{C}(t), \boldsymbol{\theta}_m),
\end{equation}
where $m$ represents the candidate kinetic model and $\boldsymbol{\theta}_m$ its associated parameter vector. The SR search budget is increased relative to the first stage, in which SR is applied directly to the concentration trajectories. This reflects the greater complexity of identifying mechanistically meaningful rate laws from estimated reaction rates, compared to fitting concentration profiles directly.

This two-stage strategy separates trajectory fitting from kinetic structure discovery, improving both the stability of the inferred derivatives and the expressiveness of the resulting rate models. It also avoids performing dynamic parameter estimation directly within the GP search: doing so would require solving the governing ODE system for every candidate expression at each generation, which is computationally impractical at the scale of SR.

\subsubsection{Parameter Estimation.}
\label{paramest}

Parameter fitting for models generated via SR is initially handled internally by \texttt{PySRRegressor}, which jointly optimises the symbolic expression and its associated parameter values during the search. For concentration trajectory models, these internally estimated parameters are used directly, and no additional calibration is required. For kinetic models, however, an explicit parameter refinement step is performed after SR. This refinement is applied both to kinetic rate expressions obtained via SR, using the parameter values provided by \texttt{PySRRegressor} as initial guesses, and to models proposed by the LLM. Since LLM-generated models provide only the functional structure, their parameters must first be initialised before undergoing the same refinement procedure. Given a candidate rate model $m$ with parameters $\boldsymbol{\theta}_m$, the corresponding concentration profile $\hat{\mathbf{C}}_m(t;\mathbf{C}_0,\boldsymbol{\theta}_m)$ is obtained by integrating $m$ forward in time from the experiment's initial condition, \textit{i.e.}, by solving
\begin{equation}
\dot{\hat{\mathbf{C}}}_m(t) = m\big(\hat{\mathbf{C}}_m(t),\boldsymbol{\theta}_m\big), \qquad \hat{\mathbf{C}}_m(t_0)=\mathbf{C}_0.
\end{equation}
That is, $\hat{\mathbf{C}}_m$ is the time integral of $m$, evaluated along the simulated trajectory rather than at the measured points $\mathbf{C}^{(i)}$ used to fit $m$. Parameters $\boldsymbol{\theta}_m$ are estimated by minimising the sum of squared errors (SSE) between $\hat{\mathbf{C}}_m$ and the experimental data:
\begin{equation}
\mathrm{SSE}(m) = \sum_{i=1}^{n_t} \left\| \mathbf{C}^{(i)} - \hat{\mathbf{C}}_m(t^{(i)};\mathbf{C}_0,\boldsymbol{\theta}_m) \right\|_Q^2.
\end{equation}
where, for an arbitrary vector $\mathbf{x}$, $\left\|\mathbf{x}
\right\|_Q^2=\mathbf{x}^{\top}Q\mathbf{x}$, and $Q$ is a weighting matrix, typically related to the inverse covariance of the measurements. In this work, all measurements within a given experiment are assumed to share the same error variance, so $Q=\mathbb{I}$, reducing the expression to the standard (unweighted) sum of squared errors. Parameter refinement is performed using the L-BFGS-B algorithm within a multistart framework to reduce sensitivity to local minima. Twenty optimisation runs are carried out from randomly generated initial values sampled uniformly within the prescribed parameter bounds, and the solution with the lowest SSE is retained.

\subsubsection{LLM-Guided Module.}
\label{sec:llm_module}

\begin{figure*}[htbp]
 \centering
 \includegraphics[width=\textwidth]{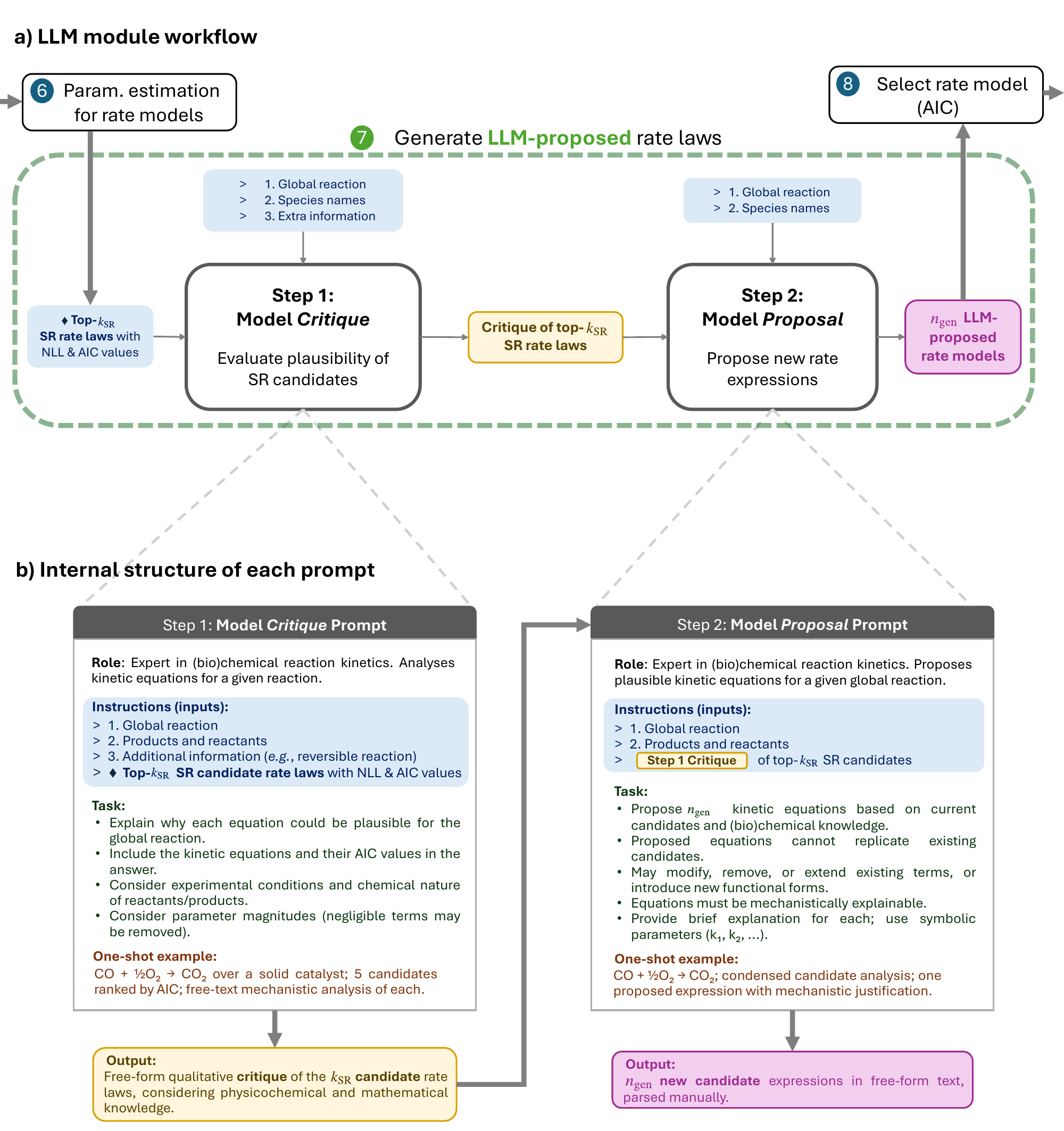}
 \caption{Structure of the LLM module used for model discovery guidance. \textbf{(a)} Two-step LLM workflow for model discovery guidance, shown within its context in our workflow (steps 6-8, numbered as in Fig. \ref{general_workflow}). Step 6 (parameter estimation) yields the top-$k_{\mathrm{SR}}$ candidate rate laws that feed into the LLM module. In the first step, the LLM critiques these best candidate models obtained by SR. The resulting critique is then used as input for the second step, in which the LLM proposes $n_{\mathrm{gen}}$ new candidate models guided by its embedded chemical and physical knowledge. These new candidates, together with the original top-$k_{\mathrm{SR}}$, are passed to step 8 (selection of rate model with AIC). \textbf{(b)} Structure of the prompts used in each step of the LLM module; dashed lines indicate that each prompt shown corresponds to the Step 1 and Step 2 boxes in (a). Step 1 receives the global reaction, species names, the top-$k_{\mathrm{SR}}$ SR candidates with their NLL and AIC values, and any additional context (\textit{e.g.}, reaction reversibility), and returns a qualitative physicochemical assessment. Step 2 receives the global reaction, species names, and the same top-$k_{\mathrm{SR}}$ SR candidates, together with the Step 1 critique, and proposes $n_{\mathrm{gen}}$ structurally distinct rate expressions with mechanistic justification. Complete prompt templates are provided in \hyperref[ESI]{\textbf{Supplementary Information}\textsuperscript{$\*$}}.}
 \label{LLM_workflow}
\end{figure*}

The LLM module corresponds to step 7 of our workflow, as shown in Figure~\ref{general_workflow}. Specifically, the LLM performs two main tasks, as detailed in Figure~\ref{LLM_workflow}a.

First, it critiques the $k_{SR}$ best-performing candidate models generated by SR (five in this work) at each iteration using their symbolic structure, number of parameters, parameter values, and associated performance metrics (negative log-likelihood (NLL) and AIC). Based on this information, it performs a qualitative assessment of each candidate, evaluating their physicochemical plausibility in terms of reaction context, parameter magnitudes, and underlying chemical and physical considerations. For this task, the LLM receives a structured prompt containing the symbolic expressions, fitted parameter values, and performance metrics of the candidate models, formatted consistently across iterations to ensure reproducibility. As shown in Figure \ref{LLM_workflow}b, the prompt includes the global reaction, species names, and a structured representation of the $k_{SR}$ best candidate rate laws with their AIC values, alongside a few-shot example to guide the response format.

Second, leveraging this assessment, the LLM proposes $n_{gen}$ new candidate equations (three in this work) informed by its prior critique and embedded domain knowledge, effectively acting as a knowledge-guided generator that directs the search toward physically consistent and interpretable models. For this task, the output of the first step is passed as additional context, together with explicit instructions to propose structurally distinct expressions and avoid reproducing existing candidates. The complete prompt templates are provided in \hyperref[ESI]{\textbf{Supplementary Information}\textsuperscript{$\*$}}.

This two-step design was motivated by preliminary evaluations indicating that separating model assessment from candidate proposal provides more effective guidance than prompting the LLM to propose new candidates from the initial SR results. By first critiquing existing solutions, the LLM can incorporate explicit feedback from the current search state before proposing new model structures. Both tasks are returned as free-form text. The proposed expressions are written in terms of the species names and symbolic parameter labels (\textit{e.g.}, $k_1$, $k_2$) provided in the prompt, and are parsed manually into a structured format compatible with the parameter estimation pipeline. We opted against structured output (\textit{e.g.}, JSON) for automated parsing, as free-form text preserves reasoning flexibility and manual parsing remained manageable at this study's scale. The entire LLM module is executed $n_{LLM}$ times in parallel (three in this work) from the same $k_{SR}$ SR candidates, yielding at most $n_{LLM} \times n_{gen}$ LLM-proposed candidates per iteration. Executions with empty or invalid output are repeated, while individual syntactically incompatible candidates (\textit{e.g.}, exponential or logarithmic forms) within an otherwise valid execution are discarded.

Subsequently, the LLM-proposed candidates are subjected to parameter estimation and evaluated using the AIC alongside those obtained from SR. The two best-performing models are then selected to guide the MBDoE step, enabling the next iteration of the algorithm. Both tasks were performed using Qwen3-14B, an open-source LLM, with its thinking mode enabled and the developer-recommended configuration (temperature of 0.6, top-p value of 0.95 and top-k value of 20). \cite{yang2025qwen3technicalreport} Its performance is further examined through ablation studies presented in Section~\ref{sec:ablation}. The complete workflow, integrating all components described above, is summarised in Algorithm \ref{alg:llm_adoks}.

\definecolor{sectioncolor}{RGB}{153,76,0}
\begin{algorithm}[!ht]
\caption{LLM-guided Symbolic Regression}
\label{alg:llm_adoks}
\KwIn{Ground-truth system; initial experimental dataset $\mathcal{D}$ 
      ($n_{\mathrm{exp}}=2$ experiments, $n_p = 30$ datapoints per experiment);
      experimental budget $n_B$ (iterations);
      number of top SR candidates critiqued by the LLM, $k_{\mathrm{SR}}$;
      number of parallel LLM instances, $n_{\mathrm{LLM}}$;
      number of new candidate expressions proposed per LLM instance, $n_{\mathrm{gen}}$.
      \\ \textit{In this work: $n_B=12$, $k_{\mathrm{SR}}=5$, $n_{\mathrm{LLM}}=3$, $n_{\mathrm{gen}}=3$.}}
\KwOut{Best kinetic model $m^*$}
\While{ground-truth model not identified \textbf{and} iterations $\leq n_B$}{
    \tcp{\textcolor[HTML]{E97132}{Concentration modelling}}
    Apply SR to $\mathcal{D}$ to identify candidate concentration models\;
    Select the best concentration model per species and experiment via AIC\;
    Estimate time derivatives $\dot{\hat{\mathbf{C}}}(t)$ from concentration profiles\;
    
    \tcp{\textcolor[HTML]{156082}{Rate model generation via SR}}
    Apply SR to $\dot{\hat{\mathbf{C}}}(t)$ to identify candidate rate expressions\;
    Refine parameters for selected SR-generated rate models via ODE integration\;
    
    \tcp{\textcolor[HTML]{196B24}{LLM module}}
    \For{$j = 1, \dots, n_{\mathrm{LLM}}$ \textbf{in parallel}}{
        \textbf{Step 1:} Given the top-$k_{\mathrm{SR}}$ SR candidates and, for each, its fitted parameters, AIC and NLL, perform a qualitative assessment of model structure and fit quality (\textit{e.g.}\ identify systematic residual trends or missing mechanistic terms)\;
        \textbf{Step 2:} Based on this assessment, propose $n_{\mathrm{gen}}$ new candidate rate expressions\;
        Convert proposed expressions into a structured symbolic representation\;
    }
    Perform parameter estimation for all LLM-proposed candidates via ODE integration\;
    
    \tcp{\textcolor[HTML]{156082}{Model selection}}
    Compute AIC for all candidates (SR \& LLM-proposed)\;
    \If{ground-truth model identified}{
        \Return best model $m^*$\;
    }
    Select top-2 models $m_1, m_2$ by AIC\;
    
    \tcp{\textcolor{sectioncolor}{MBDoE}}
    Compute $\mathbf{C}_0^{(\mathrm{new})}$ by maximizing discrepancy between $m_1$ and $m_2$\;
    Generate new experiment and append to $\mathcal{D}$\;
}
\Return best model $m^*$\;
\end{algorithm}

\subsubsection{Model Selection.}
\label{modelsec}

For model selection, candidate models (whether concentration trajectory models or kinetic rate models) are ranked using the AIC, computed from the NLL evaluated at the optimised parameter values $\boldsymbol{\theta}_m^*$ and the number of model parameters $d$. While parameter estimation (Section~\ref{paramest}) assumes a fixed weighting matrix $Q=\mathbb{I}$ across all measurements, likelihood evaluation instead uses a variance $\sigma_i^2$ estimated independently for each species, analogous to a diagonal weighting matrix whose entries reflect the inverse variance of each species. Under the assumption of independent Gaussian measurement errors with unknown variance, the NLL is computed as
\begin{equation}
\mathrm{NLL(\boldsymbol{\theta}_m^*|\mathcal{D})}=\sum_{i=1}^{n_r} \left[ \frac{(y_i-\hat{y}_i(\boldsymbol{\theta}_m^*))^2}{2\sigma_i^2}+\frac{1}{2}\ln\left(2\pi\sigma_i^2\right) \right],
\label{nll}
\end{equation}
where $y_i$ and $\hat{y}_i(\boldsymbol{\theta}_m^*)$ denote, respectively, the individual scalar components of the measurement vector $\mathbf{C}^{(i)}$ and of the corresponding model prediction $\hat{\mathbf{C}}_m(t^{(i)};\mathbf{C}_0,\boldsymbol{\theta}_m^*)$, stacked across all species and sampled times. The summation extends over all observations from all experiments, with $n_r=n_t\times n_s$ denoting the total number of residuals (the product of the number of measurement time points and the number of observed species), and $\sigma_i^2$ denotes the estimated measurement variance associated with the corresponding observation. The variance is estimated independently for each measured species based on the residual sum of squares. Note that $\sigma_i^2$ is an estimated quantity used for likelihood evaluation and model selection, and should be distinguished from the fixed noise level $\sigma$ introduced in Section \ref{datagen} to generate the synthetic measurement data. The AIC is then computed as
\begin{equation}
    \mathrm{AIC} = 2\,\mathrm{NLL}(\boldsymbol{\theta}_m^*|\mathcal{D})+2d.
    \label{aic}
\end{equation}
Given a set of candidate models, the preferred model is the one with the lowest AIC, representing the best trade-off between goodness of fit and model complexity. \cite{10.1098/rspb.2023.1261,https://doi.org/10.1002/wics.1460} If the top-ranked model matches the ground-truth structure, or is an equivalent expression differing only by negligible terms, the discovery process terminates. Otherwise, the two models with the lowest AIC values are selected for subsequent MBDoE. Although an AIC-difference threshold could serve as an early-stopping rule, we do not adopt one here, as our goal is to compare discovery efficiency across methods under a fixed, consistent experimental budget. Once the discovery process concludes, a final \textit{post-hoc} evaluation step assesses the generalisation capability of the identified models on an independent validation dataset, not involved in the iterative discovery loop, as detailed in Section~\ref{validation}.

\subsubsection{Model-Based Design of Experiments (MBDoE).}
\label{mbdoe}

MBDoE is used to generate informative experiments when the currently available data do not provide sufficient discrimination between candidate models. At each iteration, the two best-performing models according to the selection criterion (AIC) are used to design experimental conditions that maximise the discrepancy between their predicted system trajectories, using the simulated concentration profiles $\hat{\mathbf{C}}_m$ introduced in Section~\ref{paramest}.

The optimisation criterion is based on the SSE between model predictions over the experimental time horizon, following the formulation proposed by Hunter and Reiner: \cite{Hunter01081965}
\begin{equation}
    \mathbf{C}_0^{(new)} = \underset{\mathbf{C}_0}{\operatorname{arg \max}} \int_{t_0}^{t_f} \left\|
\hat{\mathbf{C}}_{m_1}(\tau;\mathbf{C}_0,\boldsymbol{\theta}_{m_1}^*)
-
\hat{\mathbf{C}}_{m_2}(\tau;\mathbf{C}_0,\boldsymbol{\theta}_{m_2}^*)
\right\|_2^2
\,d\tau,
\end{equation}
where
    \[
\left\|
\hat{\mathbf{C}}_{m_1}
-
\hat{\mathbf{C}}_{m_2}
\right\|_2
=
\sqrt{\sum_{j=1}^{n_s}
\left(
\hat{C}_{m_1,j}
-
\hat{C}_{m_2,j}
\right)^2}
.\]

Here, the design variable is restricted to the initial conditions $\mathbf{C}_0$. In principle, the same MBDoE formulation could be generalised to optimise over any other controllable process conditions (\textit{e.g.}, temperature or pressure), depending on the degrees of freedom available in a given experimental setup. The optimisation is constrained to $\mathbf{C}_0 \in [\mathbf{C}_0^{\min}, \mathbf{C}_0^{\max}]$, the same bounds used to generate the initial factorial design (Section~\ref{datagen}), which we refer to as the training domain (reported in Table~\ref{tab:case_studies}). The resulting initial conditions are used to generate new data, which are subsequently incorporated into the training dataset for the next iteration of the model discovery loop. In this way, the MBDoE step continuously enriches the available information and enhances discrimination between competing candidate models.

\subsection{Parameter Confidence Intervals \& Significance Analysis}

To assess the statistical reliability of the identified models, confidence intervals and significance tests were computed \textit{post-hoc} for the parameters of the best-performing model in each run, applied to the final model selected at convergence (Section~\ref{modelsec}). Here, a run refers to an independent repetition of the full discovery process for a given case study, starting from the same initial conditions.

Parameter uncertainty is quantified using the covariance matrix of the nonlinear least-squares estimator, derived from the Jacobian of the model residuals evaluated at the optimal parameter vector ${\boldsymbol{\theta}}^*$. The Jacobian is computed numerically via central finite differences. The residual standard deviation is estimated from the model fit as
\begin{equation}
\hat{\sigma} = \sqrt{\frac{\mathrm{SSE}(\boldsymbol{\theta}^*)}{n_r - d}},
\end{equation}
where $n_r$ denotes the total number of residuals, as defined in Section~\ref{modelsec}, and $d$ is the number of model parameters. The covariance matrix is then approximated as
\begin{equation}
\hat{\Sigma} = \hat{\sigma}^2 \left( J^\top J \right)^{-1},
\end{equation}
where $J$ denotes the Jacobian matrix evaluated at $\boldsymbol{{\theta}}^*$. Standard errors for each parameter are obtained as
\begin{equation}
\mathrm{SE}(\theta_j) =  \sqrt{\left[\hat{\Sigma}\right]_{jj}},
\end{equation}
where $\left[\hat{\Sigma}\right]_{jj}$ denotes the $j$-th diagonal element of $\hat{\Sigma}$ (\textit{i.e.}, the estimated variance of $\theta_j$), and approximate $(1-\alpha)$ confidence intervals are constructed as
\begin{equation}
\theta_j \in \left[ \theta_j^* - t_{\alpha/2,\, n_r - d} \cdot \mathrm{SE}(\theta_j);\, \theta_j^* + t_{\alpha/2,\, n_r - d} \cdot \mathrm{SE}(\theta_j) \right],
\end{equation}
where $t_{\alpha/2,\, n_r - d}$ denotes the critical value of the Student's $t$-distribution with $n_r - d$ degrees of freedom at significance level $\alpha$. In this work, a 95\% confidence level is used ($\alpha = 0.05$). 

A parameter is considered statistically significant if its associated $p$-value is below the significance threshold $\alpha = 0.05$, corresponding to a two-sided $t$-test of the null hypothesis $H_0: \theta_j = 0$. The $t$-statistic for each parameter is computed as
\begin{equation}
t_j = \frac{\theta_j^*}{\mathrm{SE}(\theta_j)},
\end{equation}
and the corresponding $p$-value is obtained from the Student's $t$-distribution with $n_r - d$ degrees of freedom. Parameters with $p > 0.05$ are considered statistically non-significant and are flagged in the results (see Section \ref{sec:results}). This approach relies on a local linear approximation of the model around the optimal parameter estimates, and therefore yields exact confidence intervals only for models that are linear in their parameters. For the nonlinear kinetic models considered here, these intervals should be interpreted as approximate, and may not fully capture asymmetries in the true confidence regions arising from nonlinearities.

\subsection{Validation Dataset and Out-of-Sample Evaluation}
\label{validation}

To assess generalisation capability beyond the experimental data used during discovery, an independent validation dataset was constructed for each case study before running the algorithm, generated by simulating the ground-truth model under 120 initial conditions, sampled both within and beyond the training domain using a fixed random seed for reproducibility.

Specifically, 30 initial conditions were sampled for each of four categories: interior points (uniformly within the training domain), boundary points (with at least one variable forced within 5\% of its lower or upper bound), moderate extrapolation (with one to three variables extended up to 20\% beyond the upper training bound), and aggressive extrapolation (up to 50\% beyond the training bounds). The training domain matched the initial condition bounds used in MBDoE for each case study (Table~\ref{tab:case_studies}), and Gaussian noise with the same standard deviation as the training data ($\sigma = 0.1$) was added to all validation measurements.

The predictive performance of each identified model was evaluated on this validation dataset after the completion of the iterative discovery process. Three metrics were computed: mean absolute error (MAE), mean squared error (MSE), and coefficient of determination ($R^2$), all calculated against the noisy validation measurements. This out-of-sample evaluation provides an external measure of model generalisation and supports a more robust comparison between the LLM-guided and baseline SR approaches.

The four sampling strategies are defined as follows. Interior points are sampled uniformly:
\begin{equation}
\mathbf{C}_0^{(i)} \sim \mathcal{U}(\mathbf{C}_0^{\min}, \mathbf{C}_0^{\max}), \quad i = 1, \ldots, 30.
\end{equation}
Boundary points are sampled such that at least one variable $j$ is forced within a fraction $\delta = 0.05$ of its lower or upper bound:
\begin{equation}
C_{0,j}^{(i)} \sim \begin{cases} 
\mathcal{U}(C_{0,j}^{\min},\, C_{0,j}^{\min} + \delta \cdot w_j) & \text{with probability } 0.5, \\
\mathcal{U}(C_{0,j}^{\max} - \delta \cdot w_j,\, C_{0,j}^{\max}) & \text{with probability } 0.5,
\end{cases}
\end{equation}
where $w_j = C_{0,j}^{\max} - C_{0,j}^{\min}$ denotes the width of the training domain for species $j$. Moderate extrapolation points extend one to three randomly selected variables up to a fraction $\epsilon = 0.20$ beyond the upper training bound:
\begin{equation}
C_{0,j}^{(i)} \sim \mathcal{U}(C_{0,j}^{\max},\, C_{0,j}^{\max} + \epsilon \cdot w_j), 
\quad j \in \mathcal{J}_{\text{out}},
\end{equation}
where $\mathcal{J}_{\text{out}}$ is a randomly selected subset of species indices. Remaining variables are sampled from the training domain. Aggressive extrapolation points follow the same 
structure with $\epsilon = 0.50$.

\subsection{Computational Setup}

All experiments were conducted sequentially, with the LLM module (Qwen3-14B) run remotely on the Imperial College London High Performance Computing cluster and all other framework components (SR, parameter estimation, and MBDoE) executed on a local workstation equipped with an Intel Core i9-12900 processor (16 cores, 24 threads) and 16 GB of RAM, running Linux Mint. Sequential execution was adopted to ensure consistent and reproducible wall-clock time measurements across all runs.

\section{Case Studies}
\label{sec:cases}

Four kinetic case studies are used to evaluate the proposed framework, spanning different levels of mechanistic complexity. The first three correspond to catalytic reaction systems from the chemical engineering literature: (1) the hydrodealkylation of toluene, \cite{fogler2016} (2) the decomposition of nitrous oxide, \cite{levenspiel1998} and (3) a theoretical isomerisation reaction, \cite{marin2019kinetics} which were also used in the original ADoK-S study. \cite{D3DD00212H} The fourth case study represents a bacterial protein production bioprocess, adapted from \citet{FORSTER2023108108}, and introduces additional complexity through a coupled ODE system with Monod-type growth kinetics. The experimental settings for all case studies are summarised in Table \ref{tab:case_studies}. 

\begin{table*}[h]
\centering
\caption{Initial conditions (IC) for the two starting experiments used to construct the initial dataset $\mathcal{D}$ for each case study, and the bounds ($\mathbf{C}_0^{\min}$, $\mathbf{C}_0^{\max}$) of the training domain. All case studies use 30 sampling points per experiment and a noise level of $\sigma = 0.1$}
\label{tab:case_studies}
\begin{tabular}{llllll}
\hline
\textbf{Case Study} & \textbf{Variables} & \textbf{IC1} & \textbf{IC2} & \textbf{$\mathbf{C}_0^{\min}$} & \textbf{$\mathbf{C}_0^{\max}$} \\
\hline
Hydrodealkylation (M) &$(C_T, C_H, C_B, C_M)$ & $(1, 8, 2, 3)$& $(5, 8, 0, 0.5)$ & $(1, 3, 0, 0.5)$ & $(5, 8, 2, 3)$ \\
Decomp. of $N_2O$ (M) &$(C_{N_2O}, C_{N_2}, C_{O_2})$ &$(5, 0, 0)$& $(10, 0, 0)$ & $(0, 0, 0)$ & $(10, 2, 3)$ \\
Isomerisation (M)& $(C_A, C_B)$ & $(2, 0)$ & $(10, 1)$ & $(0, 0)$ & $(10, 10)$ \\
Bioprocess (-)& $(B, S, P)$ &$(1, 8, 1)$& $(0.5, 10, 0)$ & $(0.1, 2, 0)$ & $(2, 20, 5)$ \\
\hline
\end{tabular}
\end{table*}

\subsection{The Hydrodealkylation of Toluene}

The hydrodealkylation of toluene is a well-established catalytic benchmark reaction from the chemical engineering literature. \cite{fogler2016} In this reaction, toluene reacts with hydrogen over a solid catalyst to produce benzene and methane,
\begin{equation*}
    \ce{C6H5CH3 + H2 <=> C6H6 + CH4}.
\end{equation*}
The ground-truth rate law follows Langmuir-Hinshelwood kinetics, derived experimentally and originally expressed in terms of partial pressures. For this work, the rate law is reformulated in terms of species concentrations, assuming ideal gas behaviour, where $C_T$, $C_H$, $C_B$, $C_M$ denote the concentrations of toluene, hydrogen, benzene, and methane, respectively. The kinetic parameters are defined as $k_A=2$ M$^{-1}$ h$^{-1}$, $k_B=9$ M$^{-1}$, and $k_C=5$ M$^{-1}$,

\begin{equation}
    r = -\frac{dC_T}{dt} = -\frac{dC_H}{dt} = \frac{dC_B}{dt} = \frac{dC_M}{dt} = \frac{k_AC_TC_H}{1+k_BC_B+k_CC_T}.
\end{equation}

\subsection{The Decomposition of Nitrous Oxide}

The decomposition of nitrous oxide is a catalytic reaction taken from the chemical engineering literature, \cite{levenspiel1998}
\begin{equation*}
    \ce{2N2O <=> 2N2 + O2}.
\end{equation*}
The ground-truth rate law exhibits a nonlinear dependence on the concentration of nitrous oxide, with inhibition in the denominator, where $C_{N_2O}$, $C_{N_2}$, and $C_{O_2}$ denote the concentrations of nitrous oxide, nitrogen, and oxygen, respectively. The kinetic parameters are defined as $k_1=2$ M$^{-1}$ h$^{-1}$ and $k_2=5$ M$^{-1}$,
\begin{equation}
    r = -\frac{1}{2}\frac{dC_{N_2O}}{dt} = \frac{1}{2}\frac{dC_{N_2}}{dt} = \frac{dC_{O_2}}{dt} = \frac{k_1 C_{N_2O}^2}{1+k_2C_{N_2O}}.
\end{equation}

\subsection{The Theoretical Isomerisation Reaction}

The isomerisation reaction is a theoretical case study taken from the kinetics literature, \cite{marin2019kinetics}
\begin{equation*}
    \ce{A <=> B}.
\end{equation*}
It represents a reversible transformation between two species A and B, governed by a rate law that incorporates both forward and reverse concentration terms in the numerator and a linear combination of concentrations in the denominator, making it structurally more complex than standard power-law expressions. The kinetic parameters are defined as $k_A=7$ M h$^{-2}$, $k_B=3$ M h$^{-2}$, $k_C=4$ h$^{-1}$, $k_D=2$ h$^{-1}$, and $k_E=6$ M h$^{-1}$.

\begin{equation}
    r=-\frac{dC_A}{dt}=\frac{dC_B}{dt}=\frac{k_AC_A-k_BC_B}{k_CC_A + k_DC_B+k_E}.
\end{equation}

\subsection{The Bacterial Production of Protein}

This case study corresponds to a batch bacterial protein production process adapted from \citet{FORSTER2023108108}. The state variables are the biomass concentration $B$, substrate concentration $S$, and product concentration $P$, whose dynamics are described by eqns \ref{biomass}, \ref{substrate}, and \ref{product}, respectively:
\begin{equation}
    \frac{dB}{dt}=\phi\cdot B,
    \label{biomass}
\end{equation}
\begin{equation}
    \frac{dS}{dt}=-\Sigma\cdot B,
    \label{substrate}
\end{equation}
\begin{equation}
    \frac{dP}{dt}=\pi\cdot B,
    \label{product}
\end{equation}
where $\phi$, $\Sigma=\phi/Y_{B,S}$, and $\pi=(Y_{P,S}/Y_{B,S})\cdot\phi$ denote the specific cell growth, substrate consumption, and product formation rates, respectively, with
\begin{equation}
    \phi = \phi_{max}\cdot \frac{S}{k_S+S}\cdot\left(1-\frac{B}{k_\phi+B}\right).
    \label{cellrate}
\end{equation}
The model structure follows \citet{FORSTER2023108108}, but the kinetic parameters were modified to $\phi_{max} = 0.5$, $k_S=2$, $k_\phi = 7$, $Y_{B,S}=5$, and $Y_{P,S}=2$ to obtain dynamics suitable for algorithmic evaluation. As this case study serves as a numerically challenging benchmark, state variables and parameters are treated as dimensionless and were chosen to produce dynamically rich trajectories rather than literature-realistic magnitudes. Additionally, for simplicity, we assume an isothermal operation. Thus, only an expression for $\frac{dB}{dt}$ needs to be identified, as the remaining ODEs are linked through yield coefficients assumed to be known. It is worth noting that, after substituting Eq.~\ref{cellrate} into the biomass balance, the target expression to be identified is
\begin{equation}
    \frac{dB}{dt}=\frac{k_1\cdot S\cdot B}{(k_S+S)\cdot(k_\phi+B)}, \qquad k_1 = \phi_{max}\cdot k_\phi,
\end{equation}
since SR represents parameters as fitted numerical constants rather than named physical parameters, and therefore recovers $\phi_{max}$ and $k_\phi$ only as their combined numerator constant $k_1$.

\section{Results and Discussion}
\label{sec:results}

\subsection{Model Discovery Performance}

Table \ref{Complete_results} summarises the results of all 32 experimental runs (4 case studies $\times$ 2 algorithms $\times$ 4 independent runs).

\begin{table*}[htbp]
\small
\begin{threeparttable}
\caption{Comparison of baseline and LLM-guided SR across case studies}
\label{Complete_results}

\begin{tabular*}{\textwidth}{@{\extracolsep{\fill}}lllll}
\hline
Case study & Run & Algorithm & Iterations \tnote{\textit{a}} & Model found \tnote{\textit{b}} \\
\hline

\multirow{8}{*}{Hydrodealkylation of Toluene}
 & 1  & Baseline SR      & 7   & $\frac{C_H \cdot C_T}{4.463 \cdot C_B + 2.633 \cdot C_T + 0.420}$\\
 & 2  & Baseline SR      & 8   & $\frac{0.416\cdot C_H \cdot C_T}{1.777 \cdot C_B + C_T + 0.513}$\\
 & 3  & Baseline SR      & 9   & $\frac{0.222\cdot C_H \cdot C_T}{C_B + 0.556 \cdot C_T + 0.147}$\\
 & 4  & Baseline SR      & 12  & $\frac{-0.024\cdot C_B + 0.352\cdot C_H \cdot C_T + 0.019 \cdot C_T}{1.508 \cdot C_B + C_T}$\\
 & 5  & LLM-guided SR    & 3(1) & $\frac{C_H \cdot C_T}{4.521 \cdot C_B + 2.576 \cdot C_T + 0.557}$\\
 & 6  & LLM-guided SR    & 4(1) & $\frac{0.378 \cdot C_H \cdot C_T}{1.536 \cdot C_B + C_T + 0.526}$\\
 & 7  & LLM-guided SR  & 5(3) \tnote{\textit{c}} & $\frac{0.218 \cdot C_H \cdot C_T}{0.112 + C_B + 0.552 \cdot C_T}$\\
 & 8  & LLM-guided SR    & 6(1) & $\frac{C_H \cdot C_T}{4.503 \cdot C_B + 2.549 \cdot C_T + 0.554}$\\
\hline

\multirow{8}{*}{Decomposition of Nitrous Oxide}
 & 9  & Baseline SR & 6  & $\frac{C_{N_2O}\cdot(0.406\cdot C_{N_2O} - 0.037)}{C_{N_2O} + 0.125}$\\
 & 10 & Baseline SR & 8  & $\frac{C_{N_2O}\cdot(0.405\cdot C_{N_2O} \textcolor{Orange}{-0.0016\cdot C_{N_2}} - 0.0195)}{C_{N_2O} + 0.133}$\\
 & 11 & Baseline SR & 12 \tnote{\textit{d}} & $0.399\cdot C_{N_2O} - 0.005 \cdot C_{N_2}-0.035$\\
 & 12 & Baseline SR & 12 \tnote{\textit{d}} & $0.399\cdot C_{N_2O} - 0.005 \cdot C_{N_2}-0.029$\\
 & 13 & LLM-guided SR & 1(1) \tnote{\textit{c}} & $\frac{0.408\cdot C_{N_2O}}{\frac{0.232}{C_{N_2O}} + 1}$\\
 & 14 & LLM-guided SR & 2(1) & $\frac{C_{N_2O}\cdot(0.403\cdot C_{N_2O} \textcolor{Orange}{- 0.019)}}{C_{N_2O} + 0.130}$\\
 & 15 & LLM-guided SR & 3(1) & $\frac{C_{N_2O}^2\cdot(0.424 \textcolor{Orange}{+ 0.003\cdot C_{N_2O}})}{C_{N_2O}  \textcolor{Orange}{+0.283}}$\\
 & 16 & LLM-guided SR & 4(1) & $\frac{C_{N_2O}\cdot(0.406\cdot C_{N_2O} \textcolor{Orange}{- 0.014})}{C_{N_2O} + 0.181}$\\
\hline

\multirow{8}{*}{Theoretical Isomerisation}
 & 17 & Baseline SR & 2 & $\frac{2.357\cdot C_A - C_B}{1.474\cdot C_A + 0.767\cdot C_B + 0.994}$\\
 & 18 & Baseline SR & 2 & $\frac{1.707\cdot C_A - 0.726 \cdot C_B \textcolor{Orange}{+ 0.005}}{C_A + 0.466\cdot C_B + 1.506}$\\
 & 19 & Baseline SR & 4 & $\frac{3.484\cdot C_A - 1.487\cdot C_B \textcolor{Orange}{+ 0.091}}{2.056\cdot C_A + C_B + 2.492}$\\
 & 20 & Baseline SR & 4 & $\frac{1.996\cdot C_A - 0.826 \cdot C_B}{ C_A + 0.677\cdot C_B + 2.667}$\\
 & 21 & LLM-guided SR & 1(1) \tnote{\textit{c}} & $\frac{1.288\cdot C_A - 0.543 \cdot C_B}{ 1+0.769 \cdot C_A + 0.382\cdot C_B }$\\
 & 22 & LLM-guided SR & 1(1) \tnote{\textit{c}} & $\frac{1.169\cdot C_A - 0.518 \cdot C_B}{ 1+0.699 \cdot C_A + 0.267\cdot C_B }$\\
 & 23 & LLM-guided SR & 2(1) \tnote{\textit{c}} & $\frac{1.176\cdot C_A - 0.518 \cdot C_B}{ 1+0.686 \cdot C_A + 0.350\cdot C_B } \textcolor{Orange}{+ 0.014}$\\
 & 24 & LLM-guided SR & 3(1) & $\frac{1.708\cdot C_A - 0.7306 \cdot C_B}{ C_A + 0.491\cdot C_B + 1.361 }$\\
\hline

\multirow{8}{*}{Bacterial Production of Protein}
 & 25 & Baseline SR & 3  & $\frac{3.744\cdot B\cdot S}{(B +7.939)\cdot(S+1.945)}$\\
 & 26 & Baseline SR & 8  & $\frac{ B\cdot S}{(0.277\cdot B+2.088)\cdot(S+1.823)}$\\
 & 27 & Baseline SR & 8  & $\frac{3.327\cdot B\cdot S}{B+S\cdot(B+6.284)+18.066}$\\
 & 28 & Baseline SR & 10 & $\frac{ B\cdot S }{(0.272\cdot B+2.113)\cdot(S+1.823)}$\\
 & 29 & LLM-guided SR & 1(1) \tnote{\textit{c}} & $\frac{3.586\cdot B\cdot S}{(B + 7.152)\cdot(S+2.100)}$\\
 & 30 & LLM-guided SR  & 1(1) \tnote{\textit{c}} & $\frac{3.585\cdot B\cdot S}{(B + 7.151)\cdot(S+2.098)}$\\
 & 31 & LLM-guided SR  & 1(1) \tnote{\textit{c}} & $\frac{3.585\cdot (B-0.063)\cdot S}{(B + 6.563)\cdot(S+2.383)}$\\
 & 32 & LLM-guided SR & 3(2) \tnote{\textit{c}}& $\frac{3.507\cdot B\cdot (S+0.024)}{(B + 7.032)\cdot(S+2.013)}$\\
\hline

\end{tabular*}

\begin{tablenotes}
\footnotesize
\item[\textit{a}] Numbers in parentheses indicate the number of iterations in which at least one LLM-proposed model was among the two best-performing candidates by AIC, when the MBDoE step was applied.
\item[\textit{b}] Terms in orange are not statistically significant at the 95\% confidence level.
\item[\textit{c}] The final model was directly suggested by the LLM.
\item[\textit{d}] The maximum number of iterations was reached without converging to the ground-truth model (or an equivalent target model).
\end{tablenotes}

\end{threeparttable}
\end{table*}

\begin{itemize}
    \item For the hydrodealkylation of toluene, both algorithms consistently recovered Langmuir--Hinshelwood-type expressions of the correct form, with the LLM-guided approach converging in 3--6 iterations compared to 7--12 for the baseline. 
    \item In the decomposition of nitrous oxide, baseline SR failed to converge within the experimental budget in two of four runs (11 and 12), recovering instead linear approximations lacking the nonlinear inhibition structure of the ground truth.
    In a real setting, such non-convergence would consume the full experimental budget without yielding a validated model, underscoring the cost of inefficient search strategies in resource-limited wet-lab campaigns.
    \item For the theoretical isomerisation, both algorithms converged rapidly, though the LLM-guided approach reached equivalent models in 1--3 iterations versus 2--4 for the baseline, with two of four LLM-guided runs identifying the ground-truth structure directly in the first iteration.
    \item The bacterial protein production case study showed the most dramatic contrast: baseline SR required 3--10 iterations, while three of four LLM-guided runs converged in a single iteration, with the LLM directly proposing the final model in each case (note (\textit{c})), underscoring its role as both search accelerator and active generator of correct structures.
\end{itemize}

\begin{figure*}[htbp]
    \centering
    \includegraphics[width=\linewidth]{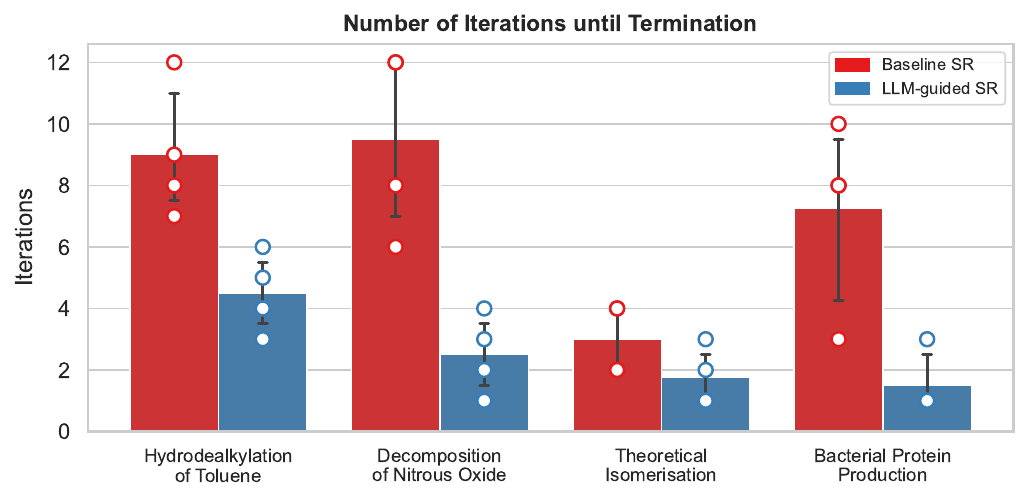}
    \caption{Number of iterations until termination (convergence or exhaustion of the experimental budget) for all runs, grouped by case study and algorithm. Bars represent the mean across four independent runs, error bars show standard deviation, and individual runs are shown as open circles. The decomposition of nitrous oxide shows the largest gap in absolute iteration count, as two baseline SR runs failed to converge within the budget, inflating its mean. Meaningful reductions are also observed for the hydrodealkylation of toluene and the bacterial protein production case, while the theoretical isomerisation shows the smallest difference, consistent with its lower intrinsic complexity and rapid convergence of both algorithms.}
    \label{iterations_grouped}
\end{figure*}

Figure \ref{iterations_grouped} consolidates these results across all runs and case studies. The LLM-guided framework consistently requires fewer iterations than the baseline SR, with the most pronounced differences observed in the nitrous oxide decomposition and bacterial production cases. The two non-converging baseline runs in the nitrous oxide case (reaching the maximum budget of 12 iterations) contribute to this gap, reinforcing that the LLM accelerates convergence and improves the likelihood of success within a fixed experimental budget.

In this framework, the LLM evaluates SR candidate models for physical plausibility, parameter relevance, and consistency with domain knowledge, allowing it to propose candidates that explore regions of the model space typically not reached by standard evolutionary search, thereby improving the interpretability of the discovered models. Incorporating the LLM into the iterative SR algorithm reduces the iterations required to recover the ground-truth model, or a near-equivalent model differing only by negligible parameter contributions, by 41.7 to 79.3\% compared to SR alone. Since each iteration corresponds to a new MBDoE-designed experiment, this translates directly into fewer experiments required to reach a validated kinetic model, a critical advantage in wet-lab settings where experimental time and resources are often the primary bottleneck. \cite{10.1039/d4dd00225c,Noh2018Pareto,Bosten2024AssistedAL}

\subsubsection{Illustrative Case (Run 32).}

We illustrate the LLM's role within the discovery loop with run 32 (Bacterial Production of Protein, LLM-guided SR), which converged to the ground-truth model in three iterations, with LLM intervention in two. In the first iteration, SR identified a set of candidate rate expressions from the initial dataset. The LLM critiqued the five best-performing candidates:
\begin{subequations}
\begin{align}
    r_1 &= \frac{k_1 \cdot B \cdot (k_2 \cdot B - k_3 \cdot P + k_4 \cdot S + k_5)}{(B-P+k_6)}\\
    r_2 &= \frac{k_1 \cdot B}{(B +k_2)}\\
    r_3 &= \frac{k_1 \cdot B \cdot (k_2\cdot B + (B^{k_3}+k_4)\cdot(k_5\cdot S + k_6))}{(B^{k_7}+k_8)}\\
    r_4 &= \frac{B \cdot (P+S)}{(k_1 \cdot B + k_2)}\\
    r_5 &= \frac{k_1 \cdot (B-k_2) \cdot ((k_3 \cdot B + k_4 \cdot S) \cdot (B+k_5) + k_5)}{(B+k_7)}
\end{align}
\end{subequations}
and, guided by its embedded knowledge, proposed expressions incorporating product inhibition, a mechanism absent from the SR candidates at this stage. The following two best-performing LLM-proposed expressions, selected based on their AIC, were passed to the MBDoE step
\begin{subequations}
\begin{align}
    r_1 &= \frac{k_1 \cdot B \cdot S \cdot (1-k_2\cdot P)}{(B + k_3)(S + k_4)}\\
    r_2 &= \frac{k_1 \cdot B^{k_2} \cdot S}{S + k_3}
\end{align}
\end{subequations}
Both are mechanistically plausible: the first extends Monod-type growth with product feedback suppression, while the second adopts a simpler power-law dependence on biomass and substrate. In the second iteration, new MBDoE-generated data allowed SR to identify improved candidate structures. As none of the LLM-proposed candidates outperformed the best SR models under AIC, the MBDoE step proceeded with two SR-generated expressions:
\begin{subequations}
\begin{align}
    r_1 &= \frac{k_1 \cdot S + (S - k_2)(k_3 B - k_4 P + k_5)(B + P + k_6) + k_7}{(S - k_8)(B + P + k_9)}, \\
    r_2 &= \frac{k_1 \cdot S + (S - k_2)(k_3 B - k_4 P + k_5)(B + k_6 P + k_7) + k_8}{(S - k_9)(B + k_{10} P + k_{11})}\end{align}
\end{subequations}
This illustrates that the LLM does not always improve upon SR. In practice, its contribution is conditional on the quality and diversity of the candidates available at each iteration. In the third and final iteration, the LLM proposed a rate expression of the form:
\begin{equation}
r = \frac{k_1 \cdot B \cdot (S - k_5)}{(B + k_2)(S + k_3)},
\end{equation}
which closely matches the ground-truth model $\frac{\phi_{max} \cdot k_\phi \cdot S \cdot B}{(k_S + S)(k_\phi + B)}$. The offset term $(S - k_5)$ in the numerator was found to be statistically significant at the 95\% confidence level, representing a minor structural discrepancy with respect to the ground-truth. Nevertheless, the identified model captures the essential Monod-type dependence on both biomass and substrate, and the algorithm was considered to have converged to a near-equivalent representation of the ground-truth kinetics. The following excerpt illustrates the reasoning process of the LLM in the third iteration of run 32, showing how the critique of SR candidates (Step 1) informed the proposal of the final model structure (Step 2):

\definecolor{headercolor}{RGB}{127,128,127}

\begin{tcolorbox}[
    enhanced,
    colback=gray!5,
    colframe=headercolor,
    boxrule=0.6pt,
    arc=3pt,
    title={\textbf{LLM output excerpt — Run 32, Iteration 3}},
    coltitle=white,
    colbacktitle=headercolor,
    fonttitle=\small\bfseries,
    left=4pt, right=4pt, top=4pt, bottom=4pt,
]
\small
\textbf{Step 1 — Critique of SR candidates (excerpt):}

\smallskip
\textit{\ttfamily The best-performing SR candidate} $r_1 = B(p_0 \cdot S - p_1) / [S(B + p_2)]$ \textit{\ttfamily introduces an inverse dependency on substrate concentration in the denominator, which contradicts classical Monod kinetics. Furthermore, the threshold term} $(p_0 \cdot S - p_1)$ \textit{\ttfamily allows the rate to become negative when substrate is scarce, which is physically inadmissible for biomass growth. Despite its strong statistical fit (AIC = $-278.5$), the model lacks mechanistic grounding. The exclusion of product inhibition in the top-performing candidates represents a potential gap, as product accumulation is known to suppress growth in fermentation systems.}

\smallskip
\textbf{Step 2 — Proposed rate expression:}
\begin{equation*}
r = \frac{k_1 \cdot B \cdot (S - K_{\min})}{(K_d + B)(K_m + S)}
\end{equation*}
\textit{\ttfamily This rate law avoids the problematic inverse dependency on $S$ observed in the SR candidates. The denominator uses Monod-type terms for both biomass and substrate, representing surface-site limitations at high cell densities. The threshold term $(S - K_{\min})$ ensures the reaction proceeds only above a minimum substrate concentration.}
\end{tcolorbox}

This example demonstrates the capacity of the LLM to identify structural limitations in SR-generated expressions and propose mechanistically sounder alternatives, even when the statistical fit of the existing candidates is strong. Taken together, run 32 highlights two key aspects of the LLM-guided framework: it can introduce mechanistically meaningful structures, such as product inhibition,  not readily accessible to standard evolutionary SR; and its contribution varies across iteration, sometimes directly guiding the search towards the correct model structure and other times leaving SR as the dominant driver. The interplay between these two components is central to the efficiency gains observed across all case studies.

Beyond accelerating convergence, the LLM module contributes to interpretability. At each iteration, it provides a qualitative assessment of the candidate rate laws, evaluating their physicochemical plausibility in terms of reaction mechanism, parameter magnitudes, and consistency with known chemistry. This critique effectively filters out candidates with implausible structures (such as positive product contributions to growth rates, as seen in the bacterial case) and directs the search towards mechanistically grounded expressions. Run 32 exemplifies this: while the SR candidates explored in the second iteration took the form of high-order rational expressions with limited mechanistic transparency, the LLM's final proposal in the third iteration recovered a compact Monod-type structure closely matching the ground truth, explicitly rejecting the top SR candidate on mechanistic grounds despite its competitive statistical fit. As shown in Section~\ref{cost_pred}, this gain in interpretability is not achieved at the expense of predictive accuracy, which remains comparable between the LLM-guided and baseline models across all case studies. Predictive performance ($R^2$) on the validation dataset was comparable between algorithms across all runs. See Section~S.1 in the \hyperref[ESI]{\textbf{Supplementary Information}\textsuperscript{$\*$}} for detailed predictive performance metrics for all case studies.

\subsubsection{Statistical Significance of Identified Parameters.}

The statistical significance analysis of the identified model parameters provides additional insight into the quality of the discovered expressions. Terms highlighted in orange in Table \ref{Complete_results} are not statistically significant at the 95\% confidence level, indicating negligible contributions to the rate expression. In the decomposition of nitrous oxide, three of four LLM-guided runs identified models containing a single non-significant term which, upon removal, yields the ground-truth rate law directly. This confirms that the LLM-guided framework consistently converges to the correct functional form, with the non-significant terms representing minor artefacts of the regression rather than meaningful mechanistic contributions. A similar pattern appears in other case studies, where near-equivalent models differ from the ground truth only by such negligible terms. Run~32 is a notable exception: the offset term $(S-k_5)$ was found statistically significant, yet numerically small, with the model preserving the essential Monod-type dependence on biomass and substrate. This shows that convergence to a near-equivalent model does not always imply statistical non-significance of the additional terms, but rather that the identified expression captures the dominant kinetic behaviour. Full parameter estimates with their 95\% confidence intervals for all converged runs are provided in the \hyperref[ESI]{\textbf{Supplementary Information}\textsuperscript{$\*$}}.

\subsubsection{Iteration Reduction Across Case Studies.}

\begin{figure}[h]
    \centering
    \includegraphics[width=0.47\linewidth]{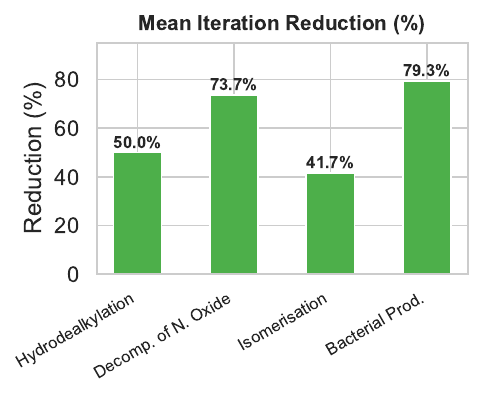}
    \caption{Mean iteration reduction (\%) achieved by the LLM-guided SR framework relative to the baseline SR, computed as the percentage decrease in mean iterations until termination. The largest reductions are observed for the bacterial protein production and decomposition of nitrous oxide case studies, where the baseline SR either required many iterations or failed to converge within the experimental budget. More moderate reductions are observed for the hydrodealkylation of toluene and the theoretical isomerisation, consistent with the lower complexity of these systems and the faster convergence of the baseline algorithm.}
    \label{fig:iteration_reduction}
\end{figure}

Figure \ref{fig:iteration_reduction} quantifies the mean reduction in iterations across case studies. The largest reductions occur for the nitrous oxide decomposition and bacterial production cases, where baseline SR struggled most, either failing to converge or requiring many iterations. The isomerisation case shows a more modest reduction, consistent with its lower intrinsic complexity and the rapid convergence of both algorithms. These results suggest that the LLM guidance benefits most where the ground-truth model structure is less accessible to purely data-driven search, and where domain knowledge adds the greatest additional value.

\subsection{Computational Cost and Predictive Performance}
\label{cost_pred}

\begin{figure}[h]
    \centering
    \includegraphics[width=0.61\linewidth]{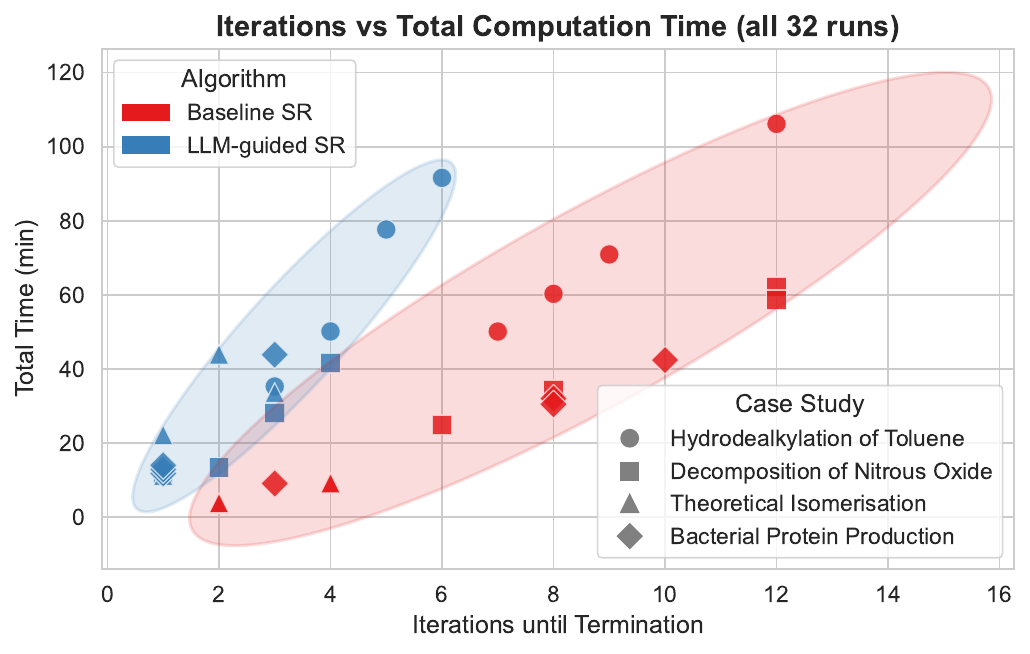}
    \caption{Wall-clock time versus number of iterations until termination for all 32 experimental runs. Each point represents a single run; colour indicates the algorithm (red: Baseline SR, blue: LLM-guided SR) and marker shape indicates the case study. The LLM-guided runs form a steeper cluster closer to the origin, reflecting a higher per-iteration cost but fewer iterations required. The baseline SR runs are displaced towards higher iteration counts with a flatter slope, indicating lower per-iteration cost. Shaded ellipses are included solely as visual guides to facilitate comparison of the clustering of runs associated with each algorithm.}
    \label{fig:iterations_vs_total_time}
\end{figure}

Figure~\ref{fig:iterations_vs_total_time} shows the relationship between the number of iterations and total wall-clock time for all 32 runs, with colour indicating the algorithm and marker shape the case study. Two distinct clusters are visible: LLM-guided runs form a cloud with a steeper slope closer to the origin, while baseline SR runs form a flatter cloud displaced towards higher iteration counts. This reflects the fundamental trade-off between the two approaches: baseline SR requires more iterations but a lower cost per iteration, since it lacks the LLM inference and additional parameter estimation steps, while the LLM-guided framework converges faster at a higher per-iteration cost. The hydrodealkylation of toluene shows the highest iteration counts and longest total times for both algorithms, reflecting the greater complexity of its rate law. Where the LLM-guided approach achieves substantial iteration reductions, the lower per-iteration efficiency is compensated by the savings from avoided physical experiments. This is particularly favourable in wet-lab experimental design, where an additional physical experiment can involve reagents, instrument time, setup, and operator effort. Where the reduction is more modest, total wall-clock time may be comparable to or slightly higher than the baseline.

\begin{figure}[!h]
    \centering
    \includegraphics[width=0.85\linewidth]{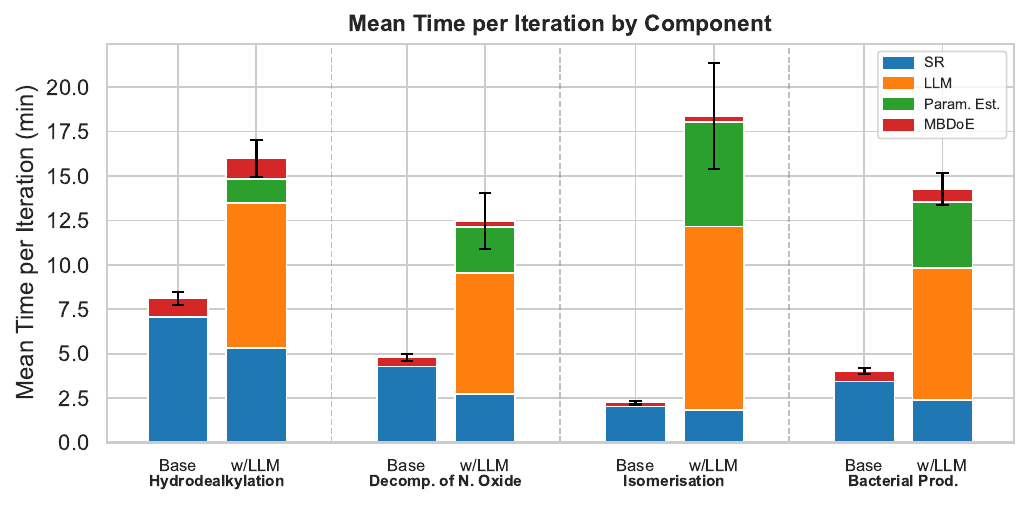}
    \caption{Mean time per iteration broken down by component (SR, LLM inference, parameter estimation, and MBDoE), for each case study and algorithm (Base: Baseline SR; w/LLM: LLM-guided SR). The total mean time per iteration is consistently higher for the LLM-guided framework, driven primarily by the LLM inference and parameter estimation steps. Error bars represent the standard error of the mean (SEM) of the total time per iteration computed across independent runs.}
    \label{fig:time_breakdown}
\end{figure}

Figure~\ref{fig:time_breakdown} shows the mean time per iteration broken down by component for each case study. In the baseline workflow, the dominant cost is the SR step itself, with MBDoE a smaller fraction. In the LLM-guided workflow, the LLM inference adds a consistent overhead that exceeds the SR cost in all case studies. The SR component is slightly lower on average here, likely because earlier convergence limits the number of iterations over which the dataset (and hence SR processing time) grows in the LLM-guided framework. MBDoE cost depends more on the case-study complexity than on the algorithm, being higher for the hydrodealkylation of toluene and bacterial protein production, and lower for the nitrous oxide decomposition and isomerisation. Overall, the per-iteration cost of the LLM-guided framework is consistently higher than the baseline, but this is compensated by fewer total iterations and experiments, a trade-off justified whenever experimental costs (\textit{e.g.}, time, materials, resources) far exceed computational costs. Detailed per-run computational cost breakdowns are provided in the \hyperref[ESI]{\textbf{Supplementary Information}\textsuperscript{$\*$}}.

A key practical consideration is the parameter estimation step, which presents a trade-off between estimation quality and computational cost. A multistart nonlinear optimisation strategy with 20 random starting points was used for all case studies. While this reduces sensitivity to local minima, it does not guarantee global optimality. Parameter estimation can represent a significant fraction of the total wall-clock time per iteration, exceeding the SR step in two of the four case studies on average.

This trade-off is particularly relevant for LLM-proposed candidates: unlike SR-generated models, which return both a symbolic expression and associated parameter estimates, these candidates provide only the functional structure without initial parameter values, making them more susceptible to convergence issues. In practice, the quality of the initial guesses can determine whether a candidate ranks competitively under AIC and is selected for MBDoE. A poorly initialised optimisation may cause a structurally correct model to be discarded in favour of a less accurate one, potentially increasing the iterations required for convergence. Improving the parameter estimation strategy, for example through physics-informed parameter estimation, \cite{nielsen2025physicsinformedregressionparameterestimation} surrogate-based optimisation, \cite{weglarztomczak2020populationbasedoptimizationkineticparameter} or increased number of starting points, \cite{math12081201} represents a promising direction for enhancing the robustness and efficiency of the framework.

\begin{figure*}[h]
    \centering
    \includegraphics[width=\linewidth]{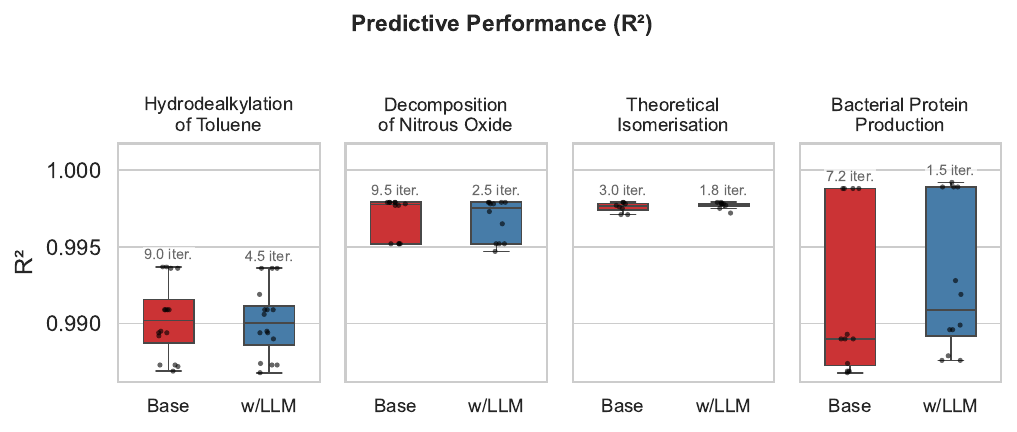}
    \caption{Predictive performance ($R^2$) of the final models identified by each algorithm on an independent validation dataset. Each point represents the $R^2$ value computed for a single species concentration profile in a single run. The number above each box indicates the mean number of iterations until termination for that algorithm-case study combination. Both algorithms achieve consistently high $R^2$ values across all case studies, with similar distributions between algorithms.}
    \label{fig:r2_comparison}
\end{figure*}

Figure \ref{fig:r2_comparison} compares the predictive performance of the final models, measured by $R^2$ on the independent validation dataset described in Section~\ref{validation}, which spans interior, boundary, and extrapolation regions of the input space. Across all case studies and species, both algorithms achieve consistently high, similarly distributed $R^2$ values, suggesting that the iteration reduction achieved by the LLM-guided framework does not come at the cost of predictive accuracy. 

The $R^2$ values are over $0.98$ across all case studies, with differences between case studies reflecting their underlying complexity rather than algorithmic differences. The theoretical isomerisation shows the highest and most tightly clustered $R^2$ values for both algorithms, consistent with its simpler rate law and rapid convergence. The hydrodealkylation of toluene and nitrous oxide decomposition show intermediate values, while the bacterial protein production case exhibits the most variability, reflecting the added complexity of its Monod-type kinetics relative to the simpler forms of the other case studies.

While the current framework still requires human intervention in parsing LLM-proposed expressions and introduces additional computational cost per iteration, these findings suggest that LLMs can systematically inject domain knowledge into SR, acting as effective knowledge-informed operators in model discovery. Crucially, this additional computational cost, while non-negligible (Figure~\ref{fig:time_breakdown}), does not involve physical resources such as reagents or instrument time, and may therefore be an acceptable trade-off in wet-lab settings where experimental resources are limiting. The results open a path toward fully automated, knowledge-driven pipelines, where manual parsing could be eliminated through structured output or code-generating LLMs, and where per-iteration LLM overhead may shrink as models improve in speed and efficiency.

\section{Ablation Studies}
\label{sec:ablation}

To isolate the individual contributions of the framework's components, two ablation studies were conducted on the bacterial protein production case study, the case study exhibiting the largest efficiency gains from LLM guidance (Section~\ref{sec:results}). The first ablation, \textit{LLM-Only Discovery}, removes the SR component entirely, evaluating whether the LLM (Qwen3-14B) can independently recover kinetic model structures without the data-driven guidance provided by SR. The second, \textit{Reduced-Size LLM}, replaces the backbone LLM (Qwen3-14B) with a smaller model from the same family (Qwen3-8B), assessing the sensitivity of the framework's performance to LLM size. Given the reduced scope of these studies relative to the main experimental campaign (Section~\ref{sec:results}), comparison is restricted to two metrics: the number of iterations until convergence and the coefficient of determination ($R^2$) of the final model on the validation dataset described in Section~\ref{validation}. Results for all ablation runs are summarised in Table~\ref{ablation_results}.

\begin{table*}[!ht]
\small
\begin{threeparttable}
\caption{Ablation study results for the bacterial protein production case study}
\label{ablation_results}
\begin{tabular*}{\textwidth}{@{\extracolsep{\fill}}llcccccl}
\hline
& & & \multicolumn{3}{c}{$R^2$} & & \\
\cmidrule(lr){4-6}
Configuration & Run & Iterations \tnote{\textit{a}} & $B$ & $S$ & $P$ & & Model found \\
\hline
\multirow{2}{*}{LLM-Only Discovery -- without data} & 33 & 6(6)\tnote{\textit{b}} & 0.988 & 0.999 & 0.990 & & $\frac{0.498\cdot B\cdot S}{(1.981+S)\cdot (1+0.142\cdot B)}$ \\
 & 34 & 12(12)\tnote{\textit{c}} & 0.209 & 0.953 & 0.351 & & $\frac{0.725\cdot B\cdot S}{(0.890+S)\cdot (1+1.450\cdot P^{0.634})}$ \\
\hline
\multirow{2}{*}{LLM-Only Discovery -- with data} & 35 & 1(1) \tnote{\textit{b}} & 0.988 & 0.999 & 0.990 & & $\frac{0.501\cdot B \cdot S}{(2.099 + S) \cdot (1+0.140 \cdot B)}$ \\
 & 36 & 6(6) & 0.988  & 0.999 & 0.990 & & $\frac{0.499\cdot B \cdot S}{(1.979 + S) \cdot (1+0.143 \cdot B)}$ \\
\hline
\multirow{4}{*}{Reduced-Size LLM (Qwen3-8B)} & 37 & 2(1)\tnote{\textit{b}} & 0.988 & 0.999 & 0.990 & & $\frac{0.249\cdot B \cdot S}{(1 + 0.500\cdot S) \cdot (1+0.142 \cdot B)}$ \\
& 38 & 3(2)\tnote{\textit{b}} & 0.988 & 0.999 & 0.990 & & $\frac{3.507\cdot B \cdot S}{(1.988 + S) \cdot (7.020 + B)}$ \\
& 39 & 3(2)\tnote{\textit{b}} & 0.988 & 0.999 & 0.990 & & $\frac{1.766\cdot B \cdot S}{(0.504 + S) \cdot (7.022 + B)}$ \\
& 40 & 5(5)\tnote{\textit{b}} & 0.988 & 0.999 & 0.990 & & $\frac{3.511\cdot B \cdot S}{(2.000 + S) \cdot (7.015 + B)}$ \\
\hline
\end{tabular*}

\begin{tablenotes}
\footnotesize
\item[\textit{a}] Numbers in parentheses indicate the number of iterations in which at least one LLM-proposed model was among the two best-performing candidates by AIC.
\item[\textit{b}] The final model was directly suggested by the LLM.
\item[\textit{c}] The maximum number of iterations was reached without converging to the ground-truth model (or an equivalent target model).
\end{tablenotes}
\end{threeparttable}
\end{table*}

\subsection{LLM-Only Discovery}

In this ablation, the SR component is removed from the discovery loop, and the LLM (Qwen3-14B) is solely responsible for generating candidate rate expressions. An initial LLM-only generation step (denoted as iteration 1) was introduced to initialise the discovery loop: the LLM is prompted to propose $k_{SR}$ kinetic equations for the biomass rate directly from the global reaction and general chemical/mathematical knowledge, without access to any SR-generated candidates. The prompt used for this seeding step mirrors the structure of the Step 2 prompt used in the full framework (Section~\ref{sec:llm_module}), here requesting $k_{SR}$ symbolically parameterised, structurally justified candidate equations. The resulting expressions are then subjected to the same parameter estimation and AIC-based ranking procedure as in the full framework (Sections~\ref{paramest} and~\ref{modelsec}), and the top-performing candidates are passed to iteration 2, where the LLM proceeds as in the standard workflow: critiquing the best candidates from the previous iteration and proposing new ones (Section~\ref{sec:llm_module}).

Two variants of this ablation were evaluated, differing in whether the LLM has access to the experimental data. Four independent runs were performed in total, split evenly between the two variants (two runs each), matching the number of runs used for the main case studies (Table~\ref{Complete_results}) to allow direct comparison. All runs used the same initial experimental dataset (Table~\ref{tab:case_studies}), sampling protocol, noise level, and experimental budget as the main bacterial production runs. Experiments beyond iteration 1 were generated via the same MBDoE procedure described in Section~\ref{mbdoe}.

\begin{itemize}
    \item \textbf{Without data:} The LLM receives only the candidate models' symbolic structure, fitted parameter values, and performance metrics (AIC, NLL, number of parameters) at each iteration, identical to the prompting scheme used in the full framework (Section~\ref{sec:llm_module}). It never has direct access to the underlying concentration measurements. Of the two runs, one failed to converge to the ground-truth model structure within the experimental budget, while the other converged after 6 iterations.
    \item \textbf{With data:} The LLM additionally receives the raw concentration-time data for all available experiments, provided as a string listing each variable's measured values over time. This gives the LLM access to information that is intentionally withheld in the original framework, where the LLM operates only on symbolic candidates and their fitted performance. This discrepancy should be considered when interpreting the results, as it may bias the LLM's proposals in ways not directly comparable to its role in the full framework. Of the two runs, one converged to the ground-truth model directly at iteration 1, with the LLM proposing the correct structure from the global reaction and raw data alone, and the second converged after 6 iterations.
\end{itemize}

\subsection{Reduced-Size LLM}

To assess the sensitivity of the framework to LLM capacity, the backbone model (Qwen3-14B) is replaced with Qwen3-8B, a smaller model from the same family, while keeping the SR component, prompting scheme, and all other framework settings unchanged. The smaller model is run with the same configuration as Qwen3-14B (thinking mode enabled, temperature of 0.6, top-p of 0.95, top-k of 20) \cite{yang2025qwen3technicalreport}, isolating the effect of model size from other confounding factors. All four runs converged to the ground-truth model structure, in 2, 3, 3, and 5 iterations, with the LLM-proposed model directly recovering the target structure in every case. This is a slight upward shift relative to the full framework using Qwen3-14B on this case study (1--3 iterations, Table~\ref{Complete_results}), plausibly linked to the reduced parameter count, but remains substantially lower than the 3--10 iterations required by baseline SR for this case study. This suggests that a smaller backbone LLM can retain most of the framework's efficiency gains over baseline SR, even if not fully matching the full-size model.
\subsection{Discussion}

Removing SR from the discovery loop increases variability in convergence and may require additional iterations to recover the target kinetic structure. The \textit{LLM-Only Discovery} variant without data access failed to converge in one of two runs, and required six iterations in the other, considerably more than the 1--3 iterations observed for the full framework on this case study (Table~\ref{Complete_results}). This suggests that SR contributes meaningfully to the framework's efficiency, likely by providing the LLM with data-grounded candidate structures to reason over, rather than relying solely on prior chemical knowledge. 

In contrast, providing the LLM with raw experimental data in the absence of SR allowed the correct model structure to be proposed directly at iteration 1 in one of two runs, and within 6 iterations in the other. This result suggests that direct access to experimental data can partially compensate for the absence of SR-derived candidates. However, this configuration provides the LLM with additional information that is not available in the proposed framework and should therefore be interpreted as an upper-bound scenario rather than a direct comparison. Importantly, the converged runs across both the ablation configurations and the complete framework achieved comparable predictive performance ($R^2>0.98$ for all state variables; full predictive performance metrics are provided in the \hyperref[ESI]{\textbf{Supplementary Information}\textsuperscript{$\*$}} indicating that the main advantage of the proposed approach lies in the efficiency of model discovery rather than in predictive accuracy after convergence.

Reducing the backbone LLM from 14B to 8B parameters did not prevent successful recovery of the ground-truth model in any of the four runs, with convergence times (2--5 iterations) largely overlapping the range observed for the full framework (Table~\ref{Complete_results}), suggesting that a smaller model can retain much of the reasoning capability required for this task. This finding is limited to a single case study, however, and evaluation on additional systems would be needed to confirm the generality of this robustness to model size.

A potential limitation is that three of the four case studies correspond to established kinetic systems from the chemical engineering literature and may have been encountered by the LLM during pretraining. The extent of such exposure cannot be determined. However, the framework never provides the LLM with the case identity or target expression, only the global reaction, symbolic candidates, and their performance metrics, and the synthetic data are generated from parameter values specific to this study. The ablation studies, conducted on the bacterial production case (an adapted system less likely to appear verbatim in pretraining data), show that the LLM alone does not consistently recover the target model, failing to converge in one of two runs and requiring six iterations in the other without data access. This suggests the observed improvements stem from the interaction between SR-derived candidates and LLM reasoning rather than memorised retrieval, though a role for prior exposure in the three literature-derived cases cannot be fully excluded.

\section{Conclusions}
\label{sec:conclusions}

We proposed DASyR-LLM, an LLM-guided SR framework for accelerated and interpretable kinetic model discovery, integrating an LLM module as a core component of the iterative discovery loop, built on ADoK-S as the underlying SR backbone. The LLM performs two complementary roles at each iteration: a qualitative physicochemical critique of the best-performing SR candidates, and the proposal of new candidate rate expressions guided by embedded domain knowledge. The framework was evaluated on four kinetic case studies of increasing complexity, spanning catalytic reactions and a bioprocess system, across 32 independent experimental runs.

The results show that the LLM-guided framework consistently reduces the iterations required to identify the ground-truth model or a near-equivalent expression, achieving mean reductions of 41.7--79.3\% compared to the baseline. Since each iteration corresponds to an MBDoE-designed experiment, this translates directly into fewer physical experiments required to reach a validated kinetic model, which is particularly valuable in wet-lab settings where experimental time and resources are the limiting factor.

The most pronounced improvements occurred in the more complex systems (nitrous oxide decomposition and bacterial protein production), where domain knowledge provided the greatest value. In 9 out of 16 LLM-guided runs, the final model was directly proposed by the LLM, highlighting its role as a search accelerator and active generator of correct model structures. In 7 runs, the LLM guided the search towards the correct functional form, subsequently recovered by SR. Predictive performance ($R^2$ on an independent validation dataset) was consistently over $0.98$ and comparable between both algorithms across all case studies, confirming that the reduction in iterations and experiments does not come at the cost of model quality.

Beyond efficiency, the framework contributes to interpretability: at each iteration, the LLM provides mechanistic justifications for the proposed rate expressions, evaluating their plausibility in terms of reaction context, parameter magnitudes, and consistency with established kinetic theory. This critique effectively filters out physically inconsistent candidates and directs the search towards mechanistically grounded expressions, a capability unavailable in purely data-driven SR approaches.

Ablation studies conducted on the bacterial production case study further clarified the role of each component. Removing SR entirely (\textit{LLM-Only Discovery}) increased variability and reduced reliability of convergence relative to the full framework, particularly when the LLM had no access to the underlying experimental data, indicating that SR-derived candidates provide essential data-grounded structure for the LLM to reason over. Reducing the backbone LLM's parameter count (Qwen3-8B instead of Qwen3-14B) did not prevent successful recovery of the ground-truth model across all four runs tested, suggesting that the framework's benefits are not exclusively tied to large-scale LLMs; further evaluation on additional case studies would help confirm the generality of this finding. Across all ablation configurations, predictive performance on the validation dataset remained comparable to the full framework, reinforcing that the primary contribution of LLM guidance lies in discovery efficiency rather than final model accuracy.

The framework also has limitations. Its per-iteration computational cost is higher than the baseline, driven mainly by LLM inference and the additional parameter estimation required for LLM-proposed candidates. The current implementation requires manual parsing of LLM-proposed expressions, introducing human intervention in an otherwise automated pipeline. As with any multistart nonlinear optimisation, the parameter estimation step does not guarantee global convergence, a limitation inherent to this class of methods \cite{Marti2018} rather than specific to the LLM-guided approach. Nonetheless, parameter-estimate quality can still influence model selection and convergence, \cite{https://doi.org/10.1002/app.45137} particularly for more complex LLM-proposed candidates. In addition, as with any data-driven approach, performance depends on the quality and completeness of the underlying data: the synthetic case studies considered here assume that all species are measurable and that measurement errors are well described by zero-mean Gaussian noise, without systematic bias or missing observations, assumptions that may not hold in real wet-lab settings.

Several directions for future work emerge from these findings. The manual parsing step could be eliminated through structured output or code-generating LLMs, enabling a fully automated pipeline. Parameter estimation could be improved through physics-informed initialisation or surrogate-based optimisation, reducing both computational cost and sensitivity to local minima. A stopping criterion for real-world applications, where the ground-truth model is unknown, remains an open challenge; convergence of the AIC across successive iterations or stabilisation of the selected model structure are promising approaches, and would be particularly valuable in wet-lab deployments to prevent unnecessary experimentation. 

The competitive performance of the data-informed LLM-Only variant (Section~\ref{sec:ablation}) also suggests that alternative configurations of LLM-driven generation, ranging from a complementary role alongside SR to a fully LLM-driven discovery loop with direct access to experimental data, merit further investigation across a broader range of case studies. Finally, the framework could be extended to larger reaction networks, systems governed by partial differential equations, or non-kinetic modelling domains, and its performance evaluated with alternative LLMs or prompt strategies to establish its generality and robustness.

Despite these limitations and open directions, this work demonstrates that embedding LLM-driven reasoning within an iterative SR framework substantially accelerates kinetic model discovery without compromising predictive accuracy, offering a practical route toward more efficient, knowledge-aware model building in chemical and biological process development.

\section*{Notation}
\begin{tabular}{lp{0.7\linewidth}}
AIC & Akaike Information Criterion \\
$\mathbf{C}(t)$              & Vector of species concentrations, $\mathbf{C}(t) \in \mathbb{R}^{n_s}$ \\
$\dot{\mathbf{C}}(t)$        & Vector of concentration time derivatives \\
$\hat{\mathbf{C}}_m(t;\mathbf{C}_0,\boldsymbol{\theta}_m)$ & Model-predicted concentration vector, obtained by integrating candidate rate model $m$ \\
$\mathbf{C}^{(i)}$           & Measured concentration vector at time $t^{(i)}$ \\
$\mathbf{C}_0$               & Initial concentration vector \\
$\mathbf{C}_0^{(\mathrm{new})}$ & New initial conditions from MBDoE \\
$\mathbf{C}_0^{\min}$, $\mathbf{C}_0^{\max}$ & Lower and upper bounds of the training domain for initial conditions \\
$d$                          & Number of model parameters \\
$\mathcal{D}$                & Experimental dataset \\
$\mathbf{f}$                 & Function mapping system state to concentration derivatives, through the (unknown) ground-truth reaction rate and stoichiometry \\
$\mathbf{h}$                 & Symbolic concentration trajectory model, fitted via SR \\
$\dot{\mathbf{h}}(t)$        & Estimated concentration time derivatives, obtained by differentiating $\mathbf{h}(t)$ \\
$J$                          & Jacobian matrix of residuals \\
$\mathcal{J}_{\text{out}}$   & Randomly selected subset of species indices for extrapolation \\
$k_{\mathrm{SR}}$            & Number of top SR candidates critiqued by the LLM \\
$m$                          & Candidate kinetic model, $m \in \mathcal{M}$ \\
$\mathcal{M}$                & Symbolic search space \\
MAE                          & Mean absolute error \\
MSE                          & Mean squared error \\
$n_B$ & Experimental budget (maximum number of iterations) \\
$n_s$                        & Number of species \\
$n_t$                        & Number of sampled measurements currently available across all experiments (grows over successive iterations) \\
$n_r$                        & Total number of residuals, $n_r = n_t \cdot n_s$ \\
$n_{\mathrm{exp}}$           & Number of initial experiments \\
$n_p$                        & Number of datapoints per experiment \\
$n_{\mathrm{LLM}}$           & Number of parallel LLM instances \\
$n_{\mathrm{gen}}$           & Number of new candidate expressions proposed per LLM instance \\
NLL                          & Negative log-likelihood function \\
$R^2$                        & Coefficient of determination \\
$\hat{\mathbf{r}}(t)$        & Estimated kinetic rate vector \\
$\mathrm{SE}(\theta_j)$      & Standard error of parameter $\theta_j$ \\
$\mathrm{SSE}(m)$            & Sum of squared errors of model $m$ \\
$t^{(i)}$                    & $i$-th measurement time \\
$t_0, t_f$                   & Initial and final times \\
$t_j$                        & $t$-statistic for parameter $\theta_j$ \\
$t_{\alpha/2,\, n_r - d}$    & Student's $t$ critical value at significance level $\alpha$ \\
$\mathcal{U}(a, b)$          & Uniform distribution on the interval $[a, b]$ \\
$w_j$                        & Width of the training domain for species $j$, $w_j = C_{0,j}^{\max} - C_{0,j}^{\min}$ \\
$\alpha$                     & Significance level for hypothesis testing ($\alpha = 0.05$) \\
$\delta$                     & Boundary fraction for near-boundary sampling ($\delta = 0.05$) \\
$\epsilon$                   & Extrapolation fraction beyond training domain ($\epsilon = 0.20$ or $0.50$) \\
$\sigma$                     & Noise standard deviation \\
$\hat{\sigma}$                & Estimated residual standard deviation \\
$\sigma_i^2$                  & Estimated measurement variance for observation $i$ \\
$\hat{\Sigma}$                & Estimated parameter covariance matrix \\
$\boldsymbol{\theta}_m$      & Parameter vector of model $m$ \\
$\theta_j$                   & Individual model parameter $j$ \\
$\boldsymbol{\theta}^*$      & Optimal parameter vector \\
\end{tabular}

\section*{Author contributions}

R.A.M.: Conceptualization, Methodology, Software, Data curation, Formal analysis, Investigation, Validation, Visualization, Writing -- original draft, Writing -- review \& editing.
P.Q.: Conceptualization, Methodology, Visualization, Writing -- review \& editing, Supervision.
A.d.R.C.: Conceptualization, Methodology, Visualization, Writing -- review \& editing, Supervision.

\section*{Conflicts of interest}
The authors declare no conflicts of interest.

\section*{Data availability}

\urlstyle{same}

All results, figures, and source code used in this work are publicly available at \url{https://github.com/RobertoAliaga/LLM-Guided-Symbolic-Regression/} (archived on Zenodo, DOI: \href{https://doi.org/10.5281/zenodo.21793265}{\nolinkurl{10.5281/zenodo.21793265}}). Full logs of the inputs and outputs generated by the LLM at each iteration of the framework are provided in the associated code repository, organised by case study and iteration. Representative prompts are also included in the \hyperref[ESI]{\textbf{Supplementary Information}\textsuperscript{$\*$}}. As Qwen3-14B is an open-source model, its configuration is reported in Section \ref{sec:llm_module} to support reproducibility.

\section*{Acknowledgements}

The authors acknowledge computational resources and support provided by the Imperial College Research Computing Service (http://doi.org/10.14469/hpc/2232). R. A. M. gratefully acknowledges financial support from the Student Exchange Programme of the Department of Chemical Engineering, Biotechnology, and Materials, University of Chile, which partially funded the research visit during which this work was carried out, and also acknowledges support from the National Agency for Research and Development (ANID), Human Capital Subdirectorate, National Master's Scholarship 2025 (Scholarship No. 22252213).

\bibliographystyle{rsc} 
\bibliography{llms} 

\clearpage

\section*{Supplementary Information}
\label{ESI}

\setcounter{figure}{0}
\renewcommand{\thefigure}{S\arabic{figure}}
\setcounter{table}{0}
\renewcommand{\thetable}{S\arabic{table}}
\setcounter{equation}{0}
\renewcommand{\theequation}{S\arabic{equation}}

\begin{landscape}
\subsection*{S1. Performance Metrics}
\label{complete_metrics}

\begin{table}[h]
\centering
\caption{Performance metrics for the Hydrodealkylation of Toluene case study}
\label{tab:algorithm_metrics_HoT}
\begin{tabular}{llcccccccccccc}
\hline
Run & Algorithm
& \multicolumn{3}{c}{$C_{T}$}
& \multicolumn{3}{c}{$C_{H}$}
& \multicolumn{3}{c}{$C_{B}$}
& \multicolumn{3}{c}{$C_{M}$} \\
\cmidrule(lr){3-5}
\cmidrule(lr){6-8}
\cmidrule(lr){9-11}
\cmidrule(lr){12-14}
&
& MAE & RMSE & $R^2$
& MAE & RMSE & $R^2$
& MAE & RMSE & $R^2$
& MAE & RMSE & $R^2$ \\
\hline
1 & Baseline SR & 0.0882 & 0.1144& 0.9909 & 0.1059 & 0.1370&0.9937 & 0.1023& 0.1301& 0.9895& 0.1161&0.1480&0.9873\\
2 & Baseline SR & 0.0883& 0.1143 & 0.9909 & 0.1052 & 0.1362& 0.9937& 0.1040& 0.1322&0.9892 &0.1173& 0.1499&0.9869\\
3 & Baseline SR & 0.0878&0.1139 & 0.9909&0.1060 & 0.1371& 0.9936& 0.1026& 0.1305& 0.9894& 0.1160& 0.1477 & 0.9873\\
4 & Baseline SR & 0.0884& 0.1141& 0.9909& 0.1067& 0.1375& 0.9936& 0.1027& 0.1308&0.9894 & 0.1166&0.1484 & 0.9872\\
\hline
5 & LLM-guided & 0.0882& 0.1145& 0.9909&0.1065 &0.1377 & 0.9936& 0.1026& 0.1306& 0.9894& 0.1161&0.1476 & 0.9873\\
6 & LLM-guided & 0.0892& 0.1158& 0.9906& 0.1197 & 0.1502& 0.9919& 0.1044& 0.1329& 0.9890& 0.1180&0.1505 & 0.9868\\
7 & LLM-guided & 0.0881 & 0.1144 & 0.9909 & 0.1068 & 0.1379& 0.9936& 0.1027& 0.1306 & 0.9894 & 0.1161 & 0.1475 & 0.9874\\
8 & LLM-guided & 0.0879& 0.1140& 0.9909& 0.1060& 0.1371& 0.9936& 0.1024& 0.1303& 0.9895& 0.1160&0.1477 &0.9873\\
\hline
\end{tabular}
\end{table}

\end{landscape}

\begin{table*}[htbp]
\centering
\caption{Performance metrics for the Decomposition of Nitrous Oxide case study}
\label{tab:algorithm_metrics_DoNO}
\begin{tabular}{llccccccccc}
\hline
Run & Algorithm
& \multicolumn{3}{c}{$C_{N_{2}O}$}
& \multicolumn{3}{c}{$C_{N_2}$}
& \multicolumn{3}{c}{$C_{O_2}$} \\
\cmidrule(lr){3-5}
\cmidrule(lr){6-8}
\cmidrule(lr){9-11}
&
& MAE & MSE & $R^2$
& MAE & MSE & $R^2$
& MAE & MSE & $R^2$ \\
\hline
9 & Baseline SR & 0.0819& 0.1038& 0.9978& 0.1242& 0.1568& 0.9979& 0.1088& 0.1379& 0.9952\\
10 & Baseline SR & 0.0809& 0.1027& 0.9979& 0.1231& 0.1557& 0.9979& 0.1084& 0.1373& 0.9952\\
11 & Baseline SR & 0.0831& 0.1054& 0.9977& 0.1228&0.1560 & 0.9979& 0.1082& 0.1371& 0.9952\\
12 & Baseline SR & 0.0830& 0.1054& 0.9977& 0.1232& 0.1565& 0.9979& 0.1082& 0.1371& 0.9952\\
\hline
13 & LLM-guided SR & 0.0815& 0.1036& 0.9978& 0.1245& 0.1574& 0.9979& 0.1088& 0.1378& 0.9952\\
14 & LLM-guided SR & 0.1053& 0.1319& 0.9965& 0.1426& 0.1799& 0.9973& 0.1137& 0.1441& 0.9947\\
15 & LLM-guided SR & 0.0813 & 0.1032& 0.9978& 0.1228& 0.1560& 0.9979& 0.1085& 0.1373& 0.9952\\
16 & LLM-guided SR & 0.0806& 0.1023& 0.9979& 0.1236& 0.1563& 0.9979& 0.1086& 0.1375& 0.9952\\
\hline
\end{tabular}
\end{table*}

\begin{table*}[htbp]
\centering
\caption{Performance metrics for the Theoretical Isomerisation case study}
\label{tab:algorithm_metrics_SI}
\begin{tabular}{llccccccccc}
\hline
Run & Algorithm
& \multicolumn{3}{c}{$C_A$}
& \multicolumn{3}{c}{$C_B$} \\
\cmidrule(lr){3-5}
\cmidrule(lr){6-8}
&
& MAE & RMSE & $R^2$
& MAE & RMSE & $R^2$ \\
\hline
17 & Baseline SR & 0.0985& 0.1266& 0.9975& 0.1032& 0.1314& 0.9976 \\
18 & Baseline SR & 0.0929& 0.1185& 0.9978& 0.0972& 0.1238& 0.9979 \\
19 & Baseline SR & 0.0940& 0.1202& 0.9977& 0.0978&0.1249 & 0.9979 \\
20 & Baseline SR & 0.1070& 0.1366& 0.9971& 0.1130& 0.1455& 0.9971 \\
\hline
21 & LLM-guided SR & 0.0945& 0.1210& 0.9977& 0.0996& 0.1269& 0.9978 \\
22 & LLM-guided SR & 0.1053& 0.1335&0.9972 & 0.1056& 0.1352& 0.9975 \\
23 & LLM-guided SR & 0.0929& 0.1189& 0.9978& 0.0969& 0.1238& 0.9979 \\
24 & LLM-guided SR & 0.0920& 0.1178& 0.9978& 0.0961& 0.1228& 0.9979 \\
\hline
\end{tabular}
\end{table*}

\begin{table*}[htbp]
\centering
\caption{Performance metrics for the Bacterial Production of Protein case study}
\label{tab:algorithm_metrics_Bio}
\begin{tabular}{llccccccccc}
\hline
Run & Algorithm
& \multicolumn{3}{c}{B}
& \multicolumn{3}{c}{S}
& \multicolumn{3}{c}{P} \\
\cmidrule(lr){3-5}
\cmidrule(lr){6-8}
\cmidrule(lr){9-11}
&
& MAE & MSE & $R^2$
& MAE & MSE & $R^2$
& MAE & MSE & $R^2$ \\
\hline
25 & Baseline SR & 0.4066& 0.8706& 0.9868& 0.1390& 0.2192& 0.9988& 0.1848& 0.3512& 0.9890\\
26 & Baseline SR & 0.4127& 0.8684& 0.9869& 0.1411& 0.2189& 0.9988& 0.1873& 0.3508& 0.9890\\
27 & Baseline SR & 0.3983& 0.8513& 0.9874& 0.1378& 0.2145& 0.9988& 0.1839& 0.3454& 0.9893\\
28 & Baseline SR & 0.4043& 0.8672& 0.9869& 0.1395& 0.2188& 0.9988& 0.1849& 0.3500& 0.9890\\
\hline
29 & LLM-guided SR & 0.3840& 0.8437& 0.9876& 0.1363& 0.2136& 0.9989& 0.1772& 0.3414& 0.9896 \\
30 & LLM-guided SR & 0.3837& 0.8435& 0.9876& 0.1363& 0.2136& 0.9989& 0.1771& 0.3413& 0.9896\\
31 & LLM-guided SR & 0.4393& 0.6850& 0.9919& 0.1365& 0.1840& 0.9992& 0.1973& 0.2836& 0.9928\\
32 & LLM-guided SR & 0.3646& 0.8330& 0.9879& 0.1365& 0.2135& 0.9989& 0.1703& 0.3368& 0.9899\\
\hline
\end{tabular}
\end{table*}

\clearpage
\newpage
\subsection*{S2. Computational Cost for Baseline SR}

\begin{table*}[htbp]
\centering
\caption{Computational time per iteration for all experimental runs using the Baseline SR workflow in the Hydrodealkylation of Toluene case study}
\label{tab:baseline_iteration_times_HoT}
\begin{tabular}{cccccc}
\hline
Run & Iteration &
SR Time (s) &
MBDoE Time (s) &
Iteration Total (s) &
Cumulative Time (s) \\
\hline

\multirow{7}{*}{1}
& 1 & 224.27 & 43.00 & 267.27 & 267.27 \\
& 2 & 271.44 & 80.79 & 352.23 & 619.50 \\
& 3 & 364.61 & 103.88 & 468.49 & 1087.99 \\
& 4 & 445.08 & 66.82 & 511.90 & 1599.89 \\
& 5 & 419.40 & 41.82 & 461.22 & 2061.11 \\
& 6 & 424.21 & 69.38 & 493.59 & 2554.70 \\
& 7 & 452.42 & -- & 452.42 & 3007.12 \\
\hline

\multirow{8}{*}{2}
& 1 & 243.15 & 62.58 & 305.73 & 305.73 \\
& 2 & 257.53 & 51.47 & 309.00 & 614.73 \\
& 3 & 298.91 & 43.05 & 341.96 & 956.69 \\
& 4 & 440.86 & 60.26 & 501.12 & 1457.81 \\
& 5 & 565.90 & 90.16 & 656.06 & 2113.87 \\
& 6 & 356.39 & 88.66 & 445.05 & 2558.92 \\
& 7 & 499.70 & 65.39 & 565.09 & 3124.01 \\
& 8 & 494.11 & -- & 494.11 & 3618.12 \\
\hline

\multirow{9}{*}{3}
& 1 & 262.73 & 75.93 & 338.66 & 338.66 \\
& 2 & 333.45 & 31.83 & 365.28 & 703.94 \\
& 3 & 569.14 & 107.95 & 677.09 & 1381.03 \\
& 4 & 310.77 & 61.85 & 372.62 & 1753.65 \\
& 5 & 387.88 & 35.93 & 423.81 & 2177.46 \\
& 6 & 478.29 & 44.31 & 522.60 & 2700.06 \\
& 7 & 388.51 & 52.92 & 441.43 & 3141.49 \\
& 8 & 463.33 & 103.66 & 566.99 & 3708.48 \\
& 9 & 547.20 & -- & 547.20 & 4255.68 \\
\hline

\multirow{12}{*}{4}
& 1 & 226.25 & 49.37 & 275.62 & 275.62 \\
& 2 & 304.70 & 41.65 & 346.35 & 621.97 \\
& 3 & 380.67 & 45.83 & 426.50 & 1048.47 \\
& 4 & 388.03 & 33.74 & 421.77 & 1470.24 \\
& 5 & 368.40 & 90.91 & 459.31 & 1929.55 \\
& 6 & 606.77 & 49.19 & 655.96 & 2585.51 \\
& 7 & 416.62 & 65.50 & 482.12 & 3067.63 \\
& 8 & 593.90 & 88.43 & 682.33 & 3749.96 \\
& 9 & 568.03 & 70.17 & 638.20 & 4388.16 \\
& 10 & 582.65 & 42.83 & 625.48 & 5013.64 \\
& 11 & 582.53 & 52.83 & 635.36 & 5649.00 \\
& 12 & 720.81 & -- & 720.81 & 6369.81 \\
\hline

\end{tabular}
\end{table*}

\begin{table*}[htbp]
\centering
\caption{Computational time per iteration for all experimental runs using the Baseline SR workflow in the Decomposition of Nitrous Oxide case study}
\label{tab:baseline_iteration_times_DoNO}
\begin{tabular}{cccccc}
\hline
Run & Iteration &
SR Time (s) &
MBDoE Time (s) &
Iteration Total (s) &
Cumulative Time (s) \\
\hline

\multirow{6}{*}{9}
& 1 & 156.06 & 13.33 & 169.39 & 169.39 \\
& 2 & 168.39 & 32.67 & 201.06 & 370.45 \\
& 3 & 258.60 & 36.09 & 294.69 & 665.14 \\
& 4 & 207.33 & 38.60 & 245.93 & 911.07 \\
& 5 & 241.05 & 60.48 & 301.53 & 1212.60 \\
& 6 & 283.06 & -- & 283.06 & 1495.66 \\
\hline

\multirow{8}{*}{10}
& 1 & 219.36 & 33.81 & 253.17 & 253.17 \\
& 2 & 167.31 & 25.71 & 193.02 & 446.19 \\
& 3 & 193.79 & 18.52 & 212.31 & 658.50 \\
& 4 & 209.45 & 49.84 & 259.29 & 917.79 \\
& 5 & 233.08 & 13.42 & 246.50 & 1164.29 \\
& 6 & 253.74 & 24.54 & 278.28 & 1442.57 \\
& 7 & 285.62 & 24.30 & 309.92 & 1752.49 \\
& 8 & 308.08 & -- & 308.08 & 2060.57 \\
\hline

\multirow{12}{*}{11}
& 1 & 145.88 & 22.21 & 168.09 & 168.09 \\
& 2 & 163.29 & 67.01 & 230.30 & 398.39 \\
& 3 & 190.05 & 7.69 & 197.74 & 596.13 \\
& 4 & 241.94 & 34.31 & 276.25 & 872.38 \\
& 5 & 239.99 & 19.20 & 259.19 & 1131.57 \\
& 6 & 263.82 & 16.91 & 280.73 & 1412.30 \\
& 7 & 311.80 & 78.10 & 389.90 & 1802.20 \\
& 8 & 312.69 & 45.45 & 358.14 & 2160.34 \\
& 9 & 338.89 & 20.24 & 359.13 & 2519.47 \\
& 10 & 380.39 & 46.50 & 426.89 & 2946.36 \\
& 11 & 375.78 & 15.95 & 391.73 & 3338.09 \\
& 12 & 394.22 & -- & 394.22 & 3732.31 \\
\hline

\multirow{12}{*}{12}
& 1 & 149.03 & 22.97 & 172.00 & 172.00 \\
& 2 & 156.97 & 12.22 & 169.19 & 341.19 \\
& 3 & 190.82 & 72.34 & 263.16 & 604.35 \\
& 4 & 205.01 & 19.98 & 224.99 & 829.34 \\
& 5 & 252.26 & 10.19 & 262.45 & 1091.79 \\
& 6 & 251.08 & 43.63 & 294.71 & 1386.50 \\
& 7 & 266.52 & 12.34 & 278.86 & 1665.36 \\
& 8 & 301.41 & 10.71 & 312.12 & 1977.48 \\
& 9 & 332.87 & 18.19 & 351.06 & 2328.54 \\
& 10 & 329.23 & 26.76 & 355.99 & 2684.53 \\
& 11 & 362.43 & 75.19 & 437.62 & 3122.15 \\
& 12 & 399.11 & -- & 399.11 & 3521.26 \\
\hline
\end{tabular}
\end{table*}

\begin{table*}[htbp]
\centering
\caption{Computational time per iteration for all experimental runs using the Baseline SR workflow in the Theoretical Isomerisation case study}
\label{tab:baseline_iteration_times_SI}
\begin{tabular}{cccccc}
\hline
Run & Iteration &
SR Time (s) &
MBDoE Time (s) &
Iteration Total (s) &
Cumulative Time (s) \\
\hline

\multirow{2}{*}{17}
& 1 & 95.45 & 3.63 & 99.08 & 99.08 \\
& 2 & 119.26 & -- & 119.26 & 218.34 \\
\hline

\multirow{2}{*}{18}
& 1 & 100.47 & 11.20 & 111.67 & 111.67 \\
& 2 & 122.44 & -- & 122.44 & 234.11 \\
\hline

\multirow{4}{*}{19}
& 1 & 102.28 & 7.45 & 109.73 & 109.73 \\
& 2 & 114.92 & 20.84 & 135.76 & 245.49 \\
& 3 & 143.21 & 18.49 & 161.70 & 407.19 \\
& 4 & 154.17 & -- & 154.17 & 561.36 \\
\hline

\multirow{4}{*}{20}
& 1 & 100.67 & 3.64 & 104.31 & 104.31 \\
& 2 & 112.91 & 16.45 & 129.36 & 233.67 \\
& 3 & 136.85 & 25.51 & 162.36 & 396.03 \\
& 4 & 156.69 & -- & 156.69 & 552.72 \\
\hline
\end{tabular}
\end{table*}

\begin{table*}[htbp]
\centering
\caption{Computational time per iteration for all experimental runs using the Baseline SR workflow in the Bacterial Production of Protein case study}
\label{tab:baseline_iteration_times_Bio}
\begin{tabular}{cccccc}
\hline
Run & Iteration &
SR Time (s) &
MBDoE Time (s) &
Iteration Total (s) &
Cumulative Time (s) \\
\hline

\multirow{3}{*}{25}
& 1 & 134.50 & 28.46 & 162.96 & 162.96 \\
& 2 & 154.60 & 53.19 & 207.79 & 370.75 \\
& 3 & 176.87 & -- & 176.87 & 547.62 \\
\hline

\multirow{8}{*}{26}
& 1 & 135.13 & 11.11 & 146.24 & 146.24 \\
& 2 & 154.56 & 48.88 & 203.44 & 349.68 \\
& 3 & 174.08 & 72.63 & 246.71 & 596.39 \\
& 4 & 189.78 & 29.10 & 218.88 & 815.27 \\
& 5 & 212.59 & 21.25 & 233.84 & 1049.11 \\
& 6 & 236.12 & 78.82 & 314.94 & 1364.05 \\
& 7 & 250.52 & 27.46 & 277.98 & 1642.03 \\
& 8 & 280.00 & -- & 280.00 & 1922.003 \\
\hline

\multirow{8}{*}{27}
& 1 & 136.00 & 6.13 & 142.13 & 142.13 \\
& 2 & 159.29 & 19.13 & 178.42 & 320.55 \\
& 3 & 179.95 & 22.69 & 202.64 & 523.19 \\
& 4 & 194.41 & 35.73 & 230.14 & 753.33 \\
& 5 & 215.55 & 48.20 & 263.75 & 1017.08 \\
& 6 & 231.89 & 18.79 & 250.68 & 1267.76 \\
& 7 & 247.13 & 41.11 & 288.24 & 1556.00 \\
& 8 & 274.82 & -- & 274.82 & 1830.82 \\
\hline

\multirow{10}{*}{28}
& 1 & 133.99 & 12.21 & 146.20 & 146.20 \\
& 2 & 151.19 & 59.84 & 211.03 & 357.23 \\
& 3 & 175.10 & 35.01 & 210.11 & 567.34 \\
& 4 & 194.35 & 9.22 & 203.57 & 770.91 \\
& 5 & 218.27 & 40.51 & 258.78 & 1029.69 \\
& 6 & 236.93 & 39.51 & 276.44 & 1306.13 \\
& 7 & 254.84 & 41.55 & 296.39 & 1602.52 \\
& 8 & 280.37 & 40.46 & 320.83 & 1923.35 \\
& 9 & 292.14 & 14.90 & 307.04 & 2230.39 \\
& 10 & 314.28 & -- & 314.28 & 2544.67 \\
\hline
\end{tabular}
\end{table*}

\clearpage
\newpage

\begin{landscape}
\subsection*{S3. Computational Cost for LLM-Guided SR}

\begin{table}[h]
\centering
\caption{Computational time per iteration for all experimental runs using the LLM-guided SR workflow in the Hydrodealkylation of Toluene case study}
\label{tab:baseline_iteration_times_HoTLLM}
\begin{tabular}{cccccccc}
\hline
Run & Iteration &
SR Time (s) &
LLM Time (s) &
Param. Est. Time (s) &
MBDoE Time (s) &
Iteration Total (s) &
Cumulative Time (s) \\
\hline

\multirow{3}{*}{5}
& 1 & 211.49 & 760 & 30.47 & 38.00 & 1039.96 & 1039.96 \\
& 2 & 284.67 & 384 & 54.97 & 50.69 & 774.33 & 1814.29 \\
& 3 & 303.04 & -- & -- & -- & 303.04 & 2117.33 \\

\hline

\multirow{4}{*}{6}
& 1 & 263.97 & 430 & 54.63 & 58.08 & 806.68 & 806.68 \\
& 2 & 386.96 & 477 & 72.10 & 50.84 & 986.90 & 1793.58 \\
& 3 & 319.82 & 453 & 42.56 & 77.59 & 892.97 & 2686.55\\
& 4 & 321.86 & -- & -- & -- & 321.86 & 3008.41 \\

\hline

\multirow{5}{*}{7}
& 1 & 236.43 & 689 & 39.27 & 47.09 & 1011.79 & 1011.79 \\
& 2 & 304.04 & 440 & 13.92 & 62.03 & 819.99 & 1831.78 \\
& 3 & 379.36 & 399 & 22.79 & 55.18 & 856.33 & 2688.11 \\
& 4 & 408.31 & 481 & 88.27 & 62.36 & 1039.94 & 3728.05 \\
& 5 & 335.95 & 524 & 71.62 & -- & 931.57 & 4659.62 \\

\hline

\multirow{6}{*}{8}
& 1 & 254.77 & 848 & 34.12 & 151.41 & 1288.30 & 1288.30 \\
& 2 & 287.52 & 356 & 78.93 & 139.99 & 862.44 & 2150.74 \\
& 3 & 318.37 & 401 & 270.18 & 60.58 & 1050.13 & 3200.87 \\
& 4 & 355.84 & 373 & 149.31 & 81.07 & 959.22 & 4160.09 \\
& 5 & 370.61 & 331 & 185.08 & 46.98 & 933.67 & 5093.76 \\
& 6 & 402.28 & -- & -- & -- & 402.28 & 5496.04 \\
\hline

\end{tabular}
\end{table}

\begin{table}[htbp]
\centering
\caption{Computational time per iteration for all experimental runs using the LLM-guided SR workflow in the Decomposition of Nitrous Oxide case study}
\label{tab:baseline_iteration_times_DoNOLLM}
\begin{tabular}{cccccccc}
\hline
Run & Iteration &
SR Time (s) &
LLM Time (s) &
Param. Est. Time (s) &
MBDoE Time (s) &
Iteration Total (s) &
Cumulative Time (s) \\
\hline

\multirow{1}{*}{13}
& 1 & 140.32 & 322 & 253.64 & -- & 715.96 & 715.96 \\

\hline

\multirow{2}{*}{14}
& 1 & 138.79 & 393 & 104.38 & 9.51 & 645.68 & 645.68 \\
& 2 & 158.59 & -- & -- & -- & 158.59 & 804.27\\

\hline

\multirow{3}{*}{15}
& 1 & 157.13 & 293 & 31.27 & 26.72 & 508.12 & 508.12\\
& 2 & 156.13 & 475 & 357.34 & 10.21 & 998.68 & 1506.80\\
& 3 & 180.95 & -- & -- & -- & 180.95 & 1687.75 \\

\hline

\multirow{4}{*}{16}
& 1 & 144.58 & 438 & 171.62 & 19.42 & 773.62 & 773.62 \\
& 2 & 162.37 & 528 & 136.94 & 25.47 & 852.78 & 1626.40 \\
& 3 & 176.80 & 421 & 36.50 & 20.28 & 654.58 & 2280.98 \\
& 4 & 213.51 & -- & -- & -- & 213.51 & 2494.49 \\
\hline
\end{tabular}
\end{table}

\begin{table}[htbp]
\centering
\caption{Computational time per iteration for all experimental runs using the LLM-guided SR workflow in the Theoretical Isomerisation case study}
\label{tab:baseline_iteration_times_SILLM}
\begin{tabular}{cccccccc}
\hline
Run & Iteration &
SR Time (s) &
LLM Time (s) &
Param. Est. Time (s) &
MBDoE Time (s) &
Iteration Total (s) &
Cumulative Time (s) \\
\hline

\multirow{1}{*}{21}
& 1 & 98.24 & 998 & 236.08 & -- & 1332.32 & 1332.32\\

\hline

\multirow{1}{*}{22}
& 1 & 101.20 & 407 & 167.84 & -- & 676.04 & 676.04\\

\hline

\multirow{2}{*}{23}
& 1 & 102.30 & 954 & 491.12 & 17.47 & 1564.89 & 1564.89\\
& 2 & 116.58 & 431 & 525.82 & -- & 1073.40 & 2638.29\\

\hline

\multirow{3}{*}{24}
& 1 & 99.50 & 533 & 431.56 & 35.59 & 1099.65 & 1099.65\\
& 2 & 123.42 & 397 & 262.49 & 7.93 & 790.84 & 1890.49\\
& 3 & 128.95 & -- & -- & -- & 128.95 & 2019.44\\
\hline
\end{tabular}
\end{table}

\begin{table}[htbp]
\centering
\caption{Computational time per iteration for all experimental runs using the LLM-guided SR workflow in the Bacterial Production of Protein case study}
\label{tab:baseline_iteration_times_BioLLM}
\begin{tabular}{cccccccc}
\hline
Run & Iteration &
SR Time (s) &
LLM Time (s) &
Param. Est. Time (s) &
MBDoE Time (s) &
Iteration Total (s) &
Cumulative Time (s) \\
\hline

\multirow{1}{*}{29}
& 1 & 134.10 & 384 & 192.32 & -- & 710.42 & 710.42 \\

\hline

\multirow{1}{*}{30}
& 1 & 131.94 & 394 & 249.14 & -- & 775.08 & 775.08 \\

\hline

\multirow{1}{*}{31}
& 1 & 134.91 & 397 & 308.23 & -- & 840.14 & 840.14 \\

\hline

\multirow{3}{*}{32}
& 1 & 133.31 & 446 & 126.63 & 18.17 & 724.11 & 724.11\\
& 2 & 153.22 & 400 & 219.42 & 71.74 & 844.38 & 1568.49\\
& 3 & 172.20 & 659 & 233.28 & -- & 1064.48 & 2632.97\\
\hline
\end{tabular}
\end{table}

\end{landscape}
\clearpage

\newpage
\subsection*{S4. Confidence Intervals and p-values}

\subsubsection*{S4.1. Hydrodealkylation of Toluene}

\begin{table}[htbp]
\centering
\caption{Parameter estimates for Run 1 (Baseline SR) --- Hydrodealkylation of Toluene}
\label{tab:params_HoT_run1}
\begin{threeparttable}
\begin{tabular}{lcccc}
\hline
Parameter & Value & 95\% CI & Uncertainty (\%) & $p$-value \\
\hline
$k_{1}$ & 4.4633 & [4.3575, 4.5692] & 2.37 & $<0.001$ \\
$k_{2}$ & 2.6332 & [2.5120, 2.7544] & 4.60 & $<0.001$ \\
$k_{3}$ & -0.4198 & [-0.7417, -0.0980] & 76.66 & 0.011 \\
\hline
\end{tabular}
\begin{tablenotes}
\small
\item Model equation: $\dfrac{C_H \cdot C_T}{4.463 \cdot C_B + 2.633 \cdot C_T + 0.420}$
\end{tablenotes}
\end{threeparttable}
\end{table}

\begin{table}[htbp]
\centering
\caption{Parameter estimates for Run 2 (Baseline SR) --- Hydrodealkylation of Toluene}
\label{tab:params_HoT_run2}
\begin{threeparttable}
\begin{tabular}{lcccc}
\hline
Parameter & Value & 95\% CI & Uncertainty (\%) & $p$-value \\
\hline
$k_{1}$ & 0.4163 & [0.3987, 0.4339] & 4.22 & $<0.001$ \\
$k_{2}$ & 1.7766 & [1.6693, 1.8839] & 6.04 & $<0.001$ \\
$k_{3}$ & 0.5125 & [0.3789, 0.6462] & 26.08 & $<0.001$ \\
\hline
\end{tabular}
\begin{tablenotes}
\small
\item Model equation: $\dfrac{0.416\cdot C_H \cdot C_T}{1.777 \cdot C_B + C_T + 0.513}$
\end{tablenotes}
\end{threeparttable}
\end{table}

\begin{table}[htbp]
\centering
\caption{Parameter estimates for Run 3 (Baseline SR) --- Hydrodealkylation of Toluene}
\label{tab:params_HoT_run3}
\begin{threeparttable}
\begin{tabular}{lcccc}
\hline
Parameter & Value & 95\% CI & Uncertainty (\%) & $p$-value \\
\hline
$k_{1}$ & 0.2219 & [0.2164, 0.2273] & 2.45 & $<0.001$ \\
$k_{2}$ & 0.5558 & [0.5246, 0.5870] & 5.61 & $<0.001$ \\
$k_{3}$ & 0.1467 & [0.0634, 0.2299] & 56.75 & $<0.001$ \\
\hline
\end{tabular}
\begin{tablenotes}
\small
\item Model equation: $\dfrac{0.222\cdot C_H \cdot C_T}{C_B + 0.556 \cdot C_T + 0.147}$
\end{tablenotes}
\end{threeparttable}
\end{table}

\begin{table}[htbp]
\centering
\caption{Parameter estimates for Run 4 (Baseline SR) --- Hydrodealkylation of Toluene}
\label{tab:params_HoT_run4}
\begin{threeparttable}
\begin{tabular}{lcccc}
\hline
Parameter & Value & 95\% CI & Uncertainty (\%) & $p$-value \\
\hline
$k_{1}$ & -0.0237 & [-0.0309, -0.0164] & 30.66 & $<0.001$ \\
$k_{2}$ & 0.3516 & [0.3388, 0.3645] & 3.65 & $<0.001$ \\
$k_{3}$ & -0.0193 & [-0.0332, -0.0054] & 72.03 & 0.007 \\
$k_{4}$ & 1.5084 & [1.4174, 1.5994] & 6.03 & $<0.001$ \\
\hline
\end{tabular}
\begin{tablenotes}
\small
\item Model equation: $\dfrac{-0.024\cdot C_B + 0.352\cdot C_H \cdot C_T + 0.019 \cdot C_T}{1.508 \cdot C_B + C_T}$
\end{tablenotes}
\end{threeparttable}
\end{table}

\begin{table}[htbp]
\centering
\caption{Parameter estimates for Run 5 (LLM-guided SR) --- Hydrodealkylation of Toluene}
\label{tab:params_HoT_run5}
\begin{threeparttable}
\begin{tabular}{lcccc}
\hline
Parameter & Value & 95\% CI & Uncertainty (\%) & $p$-value \\
\hline
$k_{1}$ & 4.5216 & [4.3453, 4.6980] & 3.90 & $<0.001$ \\
$k_{2}$ & 2.5757 & [2.5069, 2.6445] & 2.67 & $<0.001$ \\
$k_{3}$ & -0.5571 & [-1.0913, -0.0229] & 95.89 & 0.041 \\
\hline
\end{tabular}
\begin{tablenotes}
\small
\item Model equation: $\dfrac{C_H \cdot C_T}{4.521 \cdot C_B + 2.576 \cdot C_T + 0.557}$
\end{tablenotes}
\end{threeparttable}
\end{table}

\begin{table}[htbp]
\centering
\caption{Parameter estimates for Run 6 (LLM-guided SR) --- Hydrodealkylation of Toluene}
\label{tab:params_HoT_run6}
\begin{threeparttable}
\begin{tabular}{lcccc}
\hline
Parameter & Value & 95\% CI & Uncertainty (\%) & $p$-value \\
\hline
$k_{1}$ & 0.3780 & [0.3565, 0.3995] & 5.69 & $<0.001$ \\
$k_{2}$ & 1.5360 & [1.4051, 1.6669] & 8.52 & $<0.001$ \\
$k_{3}$ & 0.5264 & [0.3419, 0.7109] & 35.05 & $<0.001$ \\
\hline
\end{tabular}
\begin{tablenotes}
\small
\item Model equation: $\dfrac{0.378 \cdot C_H \cdot C_T}{1.536 \cdot C_B + C_T + 0.526}$
\end{tablenotes}
\end{threeparttable}
\end{table}

\begin{table}[htbp]
\centering
\caption{Parameter estimates for Run 7 (LLM-guided SR) --- Hydrodealkylation of Toluene}
\label{tab:params_HoT_run7}
\begin{threeparttable}
\begin{tabular}{lcccc}
\hline
Parameter & Value & 95\% CI & Uncertainty (\%) & $p$-value \\
\hline
$k_{1}$ & 0.2181 & [0.2121, 0.2242] & 2.78 & $<0.001$ \\
$k_{2}$ & 0.1119 & [0.0250, 0.1988] & 77.66 & 0.012 \\
$k_{3}$ & 0.5519 & [0.5175, 0.5863] & 6.23 & $<0.001$ \\
\hline
\end{tabular}
\begin{tablenotes}
\small
\item Model equation: $\dfrac{0.218 \cdot C_H \cdot C_T}{0.112 + C_B + 0.552 \cdot C_T}$
\end{tablenotes}
\end{threeparttable}
\end{table}

\begin{table}[htbp]
\centering
\caption{Parameter estimates for Run 8 (LLM-guided SR) --- Hydrodealkylation of Toluene}
\label{tab:params_HoT_run8}
\begin{threeparttable}
\begin{tabular}{lcccc}
\hline
Parameter & Value & 95\% CI & Uncertainty (\%) & $p$-value \\
\hline
$k_{1}$ & 4.5029 & [4.3716, 4.6342] & 2.92 & $<0.001$ \\
$k_{2}$ & 2.5488 & [2.4309, 2.6668] & 4.63 & $<0.001$ \\
$k_{3}$ & -0.5543 & [-0.8907, -0.2178] & 60.71 & 0.001 \\
\hline
\end{tabular}
\begin{tablenotes}
\small
\item Model equation: $\dfrac{C_H \cdot C_T}{4.503 \cdot C_B + 2.549 \cdot C_T + 0.554}$
\end{tablenotes}
\end{threeparttable}
\end{table}

\clearpage
\newpage
\subsubsection*{S4.2. Decomposition of Nitrous Oxide}

\begin{table}[htbp]
\centering
\caption{Parameter estimates for Run 9 (Baseline SR) --- Decomposition of Nitrous Oxide}
\label{tab:params_DoNO_run9}
\begin{threeparttable}
\begin{tabular}{lcccc}
\hline
Parameter & Value & 95\% CI & Uncertainty (\%) & $p$-value \\
\hline
$k_{1}$ & 0.4065 & [0.4021, 0.4109] & 1.08 & $<0.001$ \\
$k_{2}$ & 0.0371 & [0.0207, 0.0534] & 44.10 & $<0.001$ \\
$k_{3}$ & 0.1253 & [0.0572, 0.1935] & 54.39 & $<0.001$ \\
\hline
\end{tabular}
\begin{tablenotes}
\small
\item Model equation: $\dfrac{C_{N_2O}\cdot(0.406\cdot C_{N_2O} - 0.037)}{C_{N_2O} + 0.125}$
\end{tablenotes}
\end{threeparttable}
\end{table}

\begin{table}[htbp]
\centering
\caption{Parameter estimates for Run 10 (Baseline SR) --- Decomposition of Nitrous Oxide}
\label{tab:params_DoNO_run10}
\begin{threeparttable}
\begin{tabular}{lcccc}
\hline
Parameter & Value & 95\% CI & Uncertainty (\%) & $p$-value \\
\hline
$k_{1}$ & 0.4046 & [0.4008, 0.4084] & 0.94 & $<0.001$ \\
$k_{2}$ & 0.0016 & [-3.725e-05, 0.0033] & 102.29 & 0.055 \\
$k_{3}$ & 0.0195 & [0.0041, 0.0349] & 78.99 & 0.013 \\
$k_{4}$ & 0.1325 & [0.0580, 0.2071] & 56.25 & $<0.001$ \\
\hline
\end{tabular}
\begin{tablenotes}
\small
\item Model equation: $\dfrac{C_{N_2O}\cdot(0.405\cdot C_{N_2O} - 0.0016\cdot C_{N_2} - 0.0195)}{C_{N_2O} + 0.133}$
\end{tablenotes}
\end{threeparttable}
\end{table}

\begin{table}[htbp]
\centering
\caption{Parameter estimates for Run 11 (Baseline SR) --- Decomposition of Nitrous Oxide}
\label{tab:params_DoNO_run11}
\begin{threeparttable}
\begin{tabular}{lcccc}
\hline
Parameter & Value & 95\% CI & Uncertainty (\%) & $p$-value \\
\hline
$k_{1}$ & 0.3990 & [0.3957, 0.4022] & 0.82 & $<0.001$ \\
$k_{2}$ & 0.0045 & [0.0032, 0.0058] & 29.63 & $<0.001$ \\
$k_{3}$ & 0.0352 & [0.0305, 0.0400] & 13.37 & $<0.001$ \\
\hline
\end{tabular}
\begin{tablenotes}
\small
\item Model equation: $0.399\cdot C_{N_2O} - 0.005 \cdot C_{N_2}-0.035$
\end{tablenotes}
\end{threeparttable}
\end{table}

\begin{table}[htbp]
\centering
\caption{Parameter estimates for Run 12 (Baseline SR) --- Decomposition of Nitrous Oxide}
\label{tab:params_DoNO_run12}
\begin{threeparttable}
\begin{tabular}{lcccc}
\hline
Parameter & Value & 95\% CI & Uncertainty (\%) & $p$-value \\
\hline
$k_{1}$ & 0.3986 & [0.3961, 0.4011] & 0.63 & $<0.001$ \\
$k_{2}$ & 0.0045 & [0.0035, 0.0056] & 22.83 & $<0.001$ \\
$k_{3}$ & 0.0289 & [0.0243, 0.0336] & 16.11 & $<0.001$ \\
\hline
\end{tabular}
\begin{tablenotes}
\small
\item Model equation: $0.399\cdot C_{N_2O} - 0.005 \cdot C_{N_2}-0.029$
\end{tablenotes}
\end{threeparttable}
\end{table}

\begin{table}[htbp]
\centering
\caption{Parameter estimates for Run 13 (LLM-guided SR) --- Decomposition of Nitrous Oxide}
\label{tab:params_DoNO_run13}
\begin{threeparttable}
\begin{tabular}{lcccc}
\hline
Parameter & Value & 95\% CI & Uncertainty (\%) & $p$-value \\
\hline
$k_{1}$ & 0.4083 & [0.4012, 0.4154] & 1.74 & $<0.001$ \\
$k_{2}$ & 0.2320 & [0.1752, 0.2888] & 24.50 & $<0.001$ \\
\hline
\end{tabular}
\begin{tablenotes}
\small
\item Model equation: $\dfrac{0.408\cdot C_{N_2O}}{\dfrac{0.232}{C_{N_2O}} + 1}$
\end{tablenotes}
\end{threeparttable}
\end{table}

\begin{table}[htbp]
\centering
\caption{Parameter estimates for Run 14 (LLM-guided SR) --- Decomposition of Nitrous Oxide}
\label{tab:params_DoNO_run14}
\begin{threeparttable}
\begin{tabular}{lcccc}
\hline
Parameter & Value & 95\% CI & Uncertainty (\%) & $p$-value \\
\hline
$k_{1}$ & 0.4031 & [0.3962, 0.4099] & 1.69 & $<0.001$ \\
$k_{2}$ & 0.0197 & [-0.0024, 0.0417] & 111.97 & 0.080 \\
$k_{3}$ & 0.1297 & [0.0369, 0.2224] & 71.55 & 0.006 \\
\hline
\end{tabular}
\begin{tablenotes}
\small
\item Model equation: $\dfrac{C_{N_2O}\cdot(0.403\cdot C_{N_2O} - 0.019)}{C_{N_2O} + 0.130}$
\end{tablenotes}
\end{threeparttable}
\end{table}

\begin{table}[htbp]
\centering
\caption{Parameter estimates for Run 15 (LLM-guided SR) --- Decomposition of Nitrous Oxide}
\label{tab:params_DoNO_run15}
\begin{threeparttable}
\begin{tabular}{lcccc}
\hline
Parameter & Value & 95\% CI & Uncertainty (\%) & $p$-value \\
\hline
$k_{1}$ & 2.0000 & [1.8598, 2.1402] & 7.01 & $<0.001$ \\
$k_{2}$ & 0.4242 & [0.3353, 0.5131] & 20.96 & $<0.001$ \\
$k_{3}$ & 0.0033 & [-0.0029, 0.0095] & 187.87 & 0.296 \\
$k_{4}$ & 0.2832 & [-0.0156, 0.5819] & 105.50 & 0.063 \\
\hline
\end{tabular}
\begin{tablenotes}
\small
\item Model equation: $\dfrac{C_{N_2O}^2\cdot(0.424 + 0.003\cdot C_{N_2O})}{C_{N_2O} +0.283}$
\end{tablenotes}
\end{threeparttable}
\end{table}

\begin{table}[htbp]
\centering
\caption{Parameter estimates for Run 16 (LLM-guided SR) --- Decomposition of Nitrous Oxide}
\label{tab:params_DoNO_run16}
\begin{threeparttable}
\begin{tabular}{lcccc}
\hline
Parameter & Value & 95\% CI & Uncertainty (\%) & $p$-value \\
\hline
$k_{1}$ & 0.4055 & [0.4000, 0.4110] & 1.36 & $<0.001$ \\
$k_{2}$ & 0.0135 & [-0.0064, 0.0335] & 147.27 & 0.183 \\
$k_{3}$ & 0.1811 & [0.0964, 0.2658] & 46.78 & $<0.001$ \\
\hline
\end{tabular}
\begin{tablenotes}
\small
\item Model equation: $\dfrac{C_{N_2O}\cdot(0.406\cdot C_{N_2O} - 0.014)}{C_{N_2O} + 0.181}$
\end{tablenotes}
\end{threeparttable}
\end{table}

\clearpage
\newpage
\subsubsection*{S4.3. Theoretical Isomerisation}

\begin{table}[htbp]
\centering
\caption{Parameter estimates for Run 17 (Baseline SR) --- Theoretical Isomerisation}
\label{tab:params_SI_run17}
\begin{threeparttable}
\begin{tabular}{lcccc}
\hline
Parameter & Value & 95\% CI & Uncertainty (\%) & $p$-value \\
\hline
$k_{1}$ & 2.3575 & [2.3165, 2.3984] & 1.74 & $<0.001$ \\
$k_{2}$ & 1.4744 & [1.3990, 1.5499] & 5.12 & $<0.001$ \\
$k_{3}$ & 0.7674 & [0.7031, 0.8317] & 8.38 & $<0.001$ \\
$k_{4}$ & 0.9941 & [0.5306, 1.4575] & 46.62 & $<0.001$ \\
\hline
\end{tabular}
\begin{tablenotes}
\small
\item Model equation: $\dfrac{2.357\cdot C_A - C_B}{1.474\cdot C_A + 0.767\cdot C_B + 0.994}$
\end{tablenotes}
\end{threeparttable}
\end{table}

\begin{table}[htbp]
\centering
\caption{Parameter estimates for Run 18 (Baseline SR) --- Theoretical Isomerisation}
\label{tab:params_SI_run18}
\begin{threeparttable}
\begin{tabular}{lcccc}
\hline
Parameter & Value & 95\% CI & Uncertainty (\%) & $p$-value \\
\hline
$k_{1}$ & 1.7072 & [1.6184, 1.7961] & 5.20 & $<0.001$ \\
$k_{2}$ & 0.7260 & [0.6720, 0.7800] & 7.44 & $<0.001$ \\
$k_{3}$ & 0.0051 & [-0.1017, 0.1119] & 2100.90 & 0.925 \\
$k_{4}$ & 0.4665 & [0.4107, 0.5222] & 11.94 & $<0.001$ \\
$k_{5}$ & 1.5055 & [0.9629, 2.0481] & 36.04 & $<0.001$ \\
\hline
\end{tabular}
\begin{tablenotes}
\small
\item Model equation: $\dfrac{1.707\cdot C_A - 0.726 \cdot C_B + 0.005}{C_A + 0.466\cdot C_B + 1.506}$
\end{tablenotes}
\end{threeparttable}
\end{table}

\begin{table}[htbp]
\centering
\caption{Parameter estimates for Run 19 (Baseline SR) --- Theoretical Isomerisation}
\label{tab:params_SI_run19}
\begin{threeparttable}
\begin{tabular}{lcccc}
\hline
Parameter & Value & 95\% CI & Uncertainty (\%) & $p$-value \\
\hline
$k_{1}$ & 3.4843 & [3.2436, 3.7251] & 6.91 & $<0.001$ \\
$k_{2}$ & 1.4871 & [1.3714, 1.6027] & 7.78 & $<0.001$ \\
$k_{3}$ & 0.0911 & [-0.0355, 0.2177] & 138.93 & 0.158 \\
$k_{4}$ & 2.0560 & [1.9054, 2.2065] & 7.32 & $<0.001$ \\
$k_{5}$ & 2.4922 & [1.6818, 3.3027] & 32.52 & $<0.001$ \\
\hline
\end{tabular}
\begin{tablenotes}
\small
\item Model equation: $\dfrac{3.484\cdot C_A - 1.487\cdot C_B + 0.091}{2.056\cdot C_A + C_B + 2.492}$
\end{tablenotes}
\end{threeparttable}
\end{table}

\begin{table}[htbp]
\centering
\caption{Parameter estimates for Run 20 (Baseline SR) --- Theoretical Isomerisation}
\label{tab:params_SI_run20}
\begin{threeparttable}
\begin{tabular}{lcccc}
\hline
Parameter & Value & 95\% CI & Uncertainty (\%) & $p$-value \\
\hline
$k_{1}$ & 1.9961 & [1.8665, 2.1257] & 6.49 & $<0.001$ \\
$k_{2}$ & 0.8262 & [0.7682, 0.8842] & 7.02 & $<0.001$ \\
$k_{3}$ & 0.6773 & [0.5631, 0.7914] & 16.85 & $<0.001$ \\
$k_{4}$ & 2.6667 & [2.0807, 3.2526] & 21.97 & $<0.001$ \\
\hline
\end{tabular}
\begin{tablenotes}
\small
\item Model equation: $\dfrac{1.996\cdot C_A - 0.826 \cdot C_B}{C_A + 0.677\cdot C_B + 2.667}$
\end{tablenotes}
\end{threeparttable}
\end{table}

\begin{table}[htbp]
\centering
\caption{Parameter estimates for Run 21 (LLM-guided SR) --- Theoretical Isomerisation}
\label{tab:params_SI_run21}
\begin{threeparttable}
\begin{tabular}{lcccc}
\hline
Parameter & Value & 95\% CI & Uncertainty (\%) & $p$-value \\
\hline
$k_{1}$ & 1.2883 & [0.9214, 1.6551] & 28.48 & $<0.001$ \\
$k_{2}$ & 0.5426 & [0.3742, 0.7110] & 31.04 & $<0.001$ \\
$k_{3}$ & 0.7688 & [0.5137, 1.0239] & 33.18 & $<0.001$ \\
$k_{4}$ & 0.3821 & [0.2071, 0.5571] & 45.79 & $<0.001$ \\
\hline
\end{tabular}
\begin{tablenotes}
\small
\item Model equation: $\dfrac{1.288\cdot C_A - 0.543 \cdot C_B}{1+0.769 \cdot C_A + 0.382\cdot C_B}$
\end{tablenotes}
\end{threeparttable}
\end{table}

\begin{table}[htbp]
\centering
\caption{Parameter estimates for Run 22 (LLM-guided SR) --- Theoretical Isomerisation}
\label{tab:params_SI_run22}
\begin{threeparttable}
\begin{tabular}{lcccc}
\hline
Parameter & Value & 95\% CI & Uncertainty (\%) & $p$-value \\
\hline
$k_{1}$ & 1.1686 & [1.1094, 1.2279] & 5.07 & $<0.001$ \\
$k_{2}$ & 0.5176 & [0.4886, 0.5465] & 5.59 & $<0.001$ \\
$k_{3}$ & 0.6985 & [0.6936, 0.7035] & 0.71 & $<0.001$ \\
$k_{4}$ & 0.2670 & [0.1303, 0.4037] & 51.19 & $<0.001$ \\
\hline
\end{tabular}
\begin{tablenotes}
\small
\item Model equation: $\dfrac{1.169\cdot C_A - 0.518 \cdot C_B}{1+0.699 \cdot C_A + 0.267\cdot C_B}$
\end{tablenotes}
\end{threeparttable}
\end{table}

\begin{table}[htbp]
\centering
\caption{Parameter estimates for Run 23 (LLM-guided SR) --- Theoretical Isomerisation}
\label{tab:params_SI_run23}
\begin{threeparttable}
\begin{tabular}{lcccc}
\hline
Parameter & Value & 95\% CI & Uncertainty (\%) & $p$-value \\
\hline
$k_{1}$ & 1.1766 & [0.7661, 1.5872] & 34.89 & $<0.001$ \\
$k_{2}$ & 0.5177 & [0.3671, 0.6682] & 29.09 & $<0.001$ \\
$k_{3}$ & 0.6857 & [0.4341, 0.9373] & 36.70 & $<0.001$ \\
$k_{4}$ & 0.3500 & [0.2052, 0.4949] & 41.38 & $<0.001$ \\
$k_{5}$ & 0.0139 & [-0.0569, 0.0847] & 510.16 & 0.699 \\
\hline
\end{tabular}
\begin{tablenotes}
\small
\item Model equation: $\dfrac{1.176\cdot C_A - 0.518 \cdot C_B}{1+0.686 \cdot C_A + 0.350\cdot C_B} + 0.014$
\end{tablenotes}
\end{threeparttable}
\end{table}

\begin{table}[htbp]
\centering
\caption{Parameter estimates for Run 24 (LLM-guided SR) --- Theoretical Isomerisation}
\label{tab:params_SI_run24}
\begin{threeparttable}
\begin{tabular}{lcccc}
\hline
Parameter & Value & 95\% CI & Uncertainty (\%) & $p$-value \\
\hline
$k_{1}$ & 1.7085 & [1.6347, 1.7823] & 4.32 & $<0.001$ \\
$k_{2}$ & 0.7307 & [0.6928, 0.7685] & 5.18 & $<0.001$ \\
$k_{3}$ & 0.4913 & [0.4553, 0.5274] & 7.34 & $<0.001$ \\
$k_{4}$ & 1.3606 & [0.9954, 1.7258] & 26.84 & $<0.001$ \\
\hline
\end{tabular}
\begin{tablenotes}
\small
\item Model equation: $\dfrac{1.708\cdot C_A - 0.7306 \cdot C_B}{C_A + 0.491\cdot C_B + 1.361}$
\end{tablenotes}
\end{threeparttable}
\end{table}

\clearpage
\newpage
\subsubsection*{S4.4. Bacterial Production of Protein}

\begin{table}[htbp]
\centering
\caption{Parameter estimates for Run 25 (Baseline SR) --- Bacterial Production of Protein}
\label{tab:params_Bio_run25}
\begin{threeparttable}
\begin{tabular}{lcccc}
\hline
Parameter & Value & 95\% CI & Uncertainty (\%) & $p$-value \\
\hline
$k_{1}$ & 3.7442 & [3.6921, 3.7963] & 1.39 & $<0.001$ \\
$k_{2}$ & 7.9387 & [7.7592, 8.1182] & 2.26 & $<0.001$ \\
$k_{3}$ & 1.9452 & [1.9124, 1.9781] & 1.69 & $<0.001$ \\
\hline
\end{tabular}
\begin{tablenotes}
\small
\item Model equation: $\dfrac{3.744\cdot B\cdot S}{(B +7.939)\cdot(S+1.945)}$
\end{tablenotes}
\end{threeparttable}
\end{table}

\begin{table}[htbp]
\centering
\caption{Parameter estimates for Run 26 (Baseline SR) --- Bacterial Production of Protein}
\label{tab:params_Bio_run26}
\begin{threeparttable}
\begin{tabular}{lcccc}
\hline
Parameter & Value & 95\% CI & Uncertainty (\%) & $p$-value \\
\hline
$k_{1}$ & 0.2772 & [0.2755, 0.2788] & 0.59 & $<0.001$ \\
$k_{2}$ & 2.0879 & [2.0771, 2.0988] & 0.52 & $<0.001$ \\
$k_{3}$ & 1.7436 & [1.7145, 1.7726] & 1.67 & $<0.001$ \\
\hline
\end{tabular}
\begin{tablenotes}
\small
\item Model equation: $\dfrac{B\cdot S}{(0.277\cdot B+2.088)\cdot(S+1.823)}$
\end{tablenotes}
\end{threeparttable}
\end{table}

\begin{table}[htbp]
\centering
\caption{Parameter estimates for Run 27 (Baseline SR) --- Bacterial Production of Protein}
\label{tab:params_Bio_run27}
\begin{threeparttable}
\begin{tabular}{lcccc}
\hline
Parameter & Value & 95\% CI & Uncertainty (\%) & $p$-value \\
\hline
$k_{1}$ & 3.3267 & [3.3081, 3.3452] & 0.56 & $<0.001$ \\
$k_{2}$ & 6.2837 & [6.2235, 6.3439] & 0.96 & $<0.001$ \\
$k_{3}$ & 18.0658 & [17.7951, 18.3364] & 1.50 & $<0.001$ \\
\hline
\end{tabular}
\begin{tablenotes}
\small
\item Model equation: $\dfrac{3.327\cdot B\cdot S}{B+S\cdot(B+6.284)+18.066}$
\end{tablenotes}
\end{threeparttable}
\end{table}

\begin{table}[htbp]
\centering
\caption{Parameter estimates for Run 28 (Baseline SR) --- Bacterial Production of Protein}
\label{tab:params_Bio_run28}
\begin{threeparttable}
\begin{tabular}{lcccc}
\hline
Parameter & Value & 95\% CI & Uncertainty (\%) & $p$-value \\
\hline
$k_{1}$ & 0.2715 & [0.2698, 0.2733] & 0.64 & $<0.001$ \\
$k_{2}$ & 2.1126 & [2.1027, 2.1224] & 0.47 & $<0.001$ \\
$k_{3}$ & 1.8228 & [1.8024, 1.8432] & 1.12 & $<0.001$ \\
\hline
\end{tabular}
\begin{tablenotes}
\small
\item Model equation: $\dfrac{B\cdot S}{(0.272\cdot B+2.113)\cdot(S+1.823)}$
\end{tablenotes}
\end{threeparttable}
\end{table}

\begin{table}[htbp]
\centering
\caption{Parameter estimates for Run 29 (LLM-guided SR) --- Bacterial Production of Protein}
\label{tab:params_Bio_run29}
\begin{threeparttable}
\begin{tabular}{lcccc}
\hline
Parameter & Value & 95\% CI & Uncertainty (\%) & $p$-value \\
\hline
$k_{1}$ & 3.5858 & [3.5392, 3.6323] & 1.30 & $<0.001$ \\
$k_{2}$ & 7.1516 & [7.1427, 7.1604] & 0.12 & $<0.001$ \\
$k_{3}$ & 2.0997 & [1.9717, 2.2278] & 6.10 & $<0.001$ \\
\hline
\end{tabular}
\begin{tablenotes}
\small
\item Model equation: $\dfrac{3.586\cdot B\cdot S}{(B + 7.152)\cdot(S+2.100)}$
\end{tablenotes}
\end{threeparttable}
\end{table}

\begin{table}[htbp]
\centering
\caption{Parameter estimates for Run 30 (LLM-guided SR) --- Bacterial Production of Protein}
\label{tab:params_Bio_run30}
\begin{threeparttable}
\begin{tabular}{lcccc}
\hline
Parameter & Value & 95\% CI & Uncertainty (\%) & $p$-value \\
\hline
$k_{1}$ & 3.5850 & [3.5385, 3.6315] & 1.30 & $<0.001$ \\
$k_{2}$ & 7.1510 & [7.1421, 7.1598] & 0.12 & $<0.001$ \\
$k_{3}$ & 2.0981 & [1.9701, 2.2262] & 6.10 & $<0.001$ \\
\hline
\end{tabular}
\begin{tablenotes}
\small
\item Model equation: $\dfrac{3.585\cdot B\cdot S}{(B + 7.151)\cdot(S+2.098)}$
\end{tablenotes}
\end{threeparttable}
\end{table}

\begin{table}[htbp]
\centering
\caption{Parameter estimates for Run 31 (LLM-guided SR) --- Bacterial Production of Protein}
\label{tab:params_Bio_run31}
\begin{threeparttable}
\begin{tabular}{lcccc}
\hline
Parameter & Value & 95\% CI & Uncertainty (\%) & $p$-value \\
\hline
$k_{1}$ & 3.5847 & [3.4481, 3.7213] & 3.81 & $<0.001$ \\
$k_{2}$ & 0.0631 & [0.0241, 0.1020] & 61.74 & 0.002 \\
$k_{3}$ & 6.5629 & [6.2145, 6.9113] & 5.31 & $<0.001$ \\
$k_{4}$ & 2.3832 & [2.1103, 2.6561] & 11.45 & $<0.001$ \\
\hline
\end{tabular}
\begin{tablenotes}
\small
\item Model equation: $\dfrac{3.585\cdot (B-0.063)\cdot S}{(B + 6.563)\cdot(S+2.383)}$
\end{tablenotes}
\end{threeparttable}
\end{table}

\begin{table}[htbp]
\centering
\caption{Parameter estimates for Run 32 (LLM-guided SR) --- Bacterial Production of Protein}
\label{tab:params_Bio_run32}
\begin{threeparttable}
\begin{tabular}{lcccc}
\hline
Parameter & Value & 95\% CI & Uncertainty (\%) & $p$-value \\
\hline
$k_{1}$ & 3.5070 & [3.4882, 3.5258] & 0.54 & $<0.001$ \\
$k_{2}$ & -0.0245 & [-0.0251, -0.0238] & 2.59 & $<0.001$ \\
$k_{3}$ & 7.0316 & [6.9192, 7.1439] & 1.60 & $<0.001$ \\
$k_{4}$ & 2.0125 & [1.9373, 2.0877] & 3.74 & $<0.001$ \\
\hline
\end{tabular}
\begin{tablenotes}
\small
\item Model equation: $\dfrac{3.507\cdot B\cdot (S+0.024)}{(B + 7.032)\cdot(S+2.013)}$
\end{tablenotes}
\end{threeparttable}
\end{table}

\clearpage
\newpage
\subsection*{S5. Comparison of Related Work (Detailed)}

\begin{table*}[!ht]
\centering
\footnotesize
\setlength{\tabcolsep}{4pt}
\renewcommand{\arraystretch}{1.15}
\begin{threeparttable}
\caption{Comparison of representative model/equation discovery approaches}
\label{tab:related_work_complete}
\begin{tabular}{@{}p{3.4cm}p{3.4cm}p{3.6cm}p{1.1cm}p{1.0cm}p{1.1cm}p{1.5cm}p{0.8cm}@{}}
\toprule
\textbf{Reference} & \textbf{Approach type} & \textbf{Domain knowledge} & \textbf{Mech. eval.} & \textbf{MBDoE} & \textbf{Iter.} & \textbf{Kinetics/ reactions} & \textbf{LLM} \\
\midrule

SINDy; \cite{doi:10.1073/pnas.1517384113} DF-SINDy; \cite{doi:10.1021/acs.iecr.4c02981} DoE-SINDy \cite{LYU2025109265} &
Sparse regression over function library; DF-SINDy: derivative-free; DoE-SINDy: + iterative DoE &
Implicit (library); explicit in DF-SINDy &
\xmark &
\xmark / \cmark\tnote{\textit{a}} &
\xmark / \cmark\tnote{\textit{a}} &
\xmark / \cmark\tnote{\textit{b}} &
\xmark \\
\addlinespace

KINNs \cite{GUSMAO2023113701} &
Physics-informed NNs for kinetics (forward/inverse) &
Yes, equality/inequality constraints &
\xmark &
\xmark &
\xmark &
\cmark &
\xmark \\
\addlinespace

ALAMO \cite{WILSON2017785} &
MINLP-based algebraic model learning &
Yes, first-principles output constraints &
\xmark &
\cmark &
\cmark &
\cmark &
\xmark \\
\addlinespace

ALVEN \cite{SUN2020107103} &
Nonlinear feature generation + elastic net &
Yes, predefined chem./bio.\ feature families &
\xmark &
\xmark &
\xmark &
Partial &
\xmark \\
\addlinespace

ADoK \cite{D3DD00212H} &
SR + parameter est.\ + AIC selection &
Minimal (baseline) &
\xmark &
\cmark &
\cmark &
\cmark &
\xmark \\
\addlinespace

SR-MbDoE; \cite{ROGERS2024120580, ROGERS2025109036} SR+MBDoE for reaction networks \cite{doi:10.1021/acs.jcim.5c03032} &
SR proposes candidate models (Pareto front / substructure-decomposed); MBDoE discriminates iteratively &
Yes, structural constraints (Rogers: PFD-oriented; Kay \& Zhang: substructure decomposition for mechanistic forms) &
\xmark &
\cmark &
\cmark &
\xmark / \cmark\tnote{\textit{c}} &
\xmark \\
\addlinespace

\midrule
\multicolumn{8}{l}{\textit{LLM-guided approaches}} \\
\addlinespace

LLM-Meta-SR; \cite{zhang2026llmmetasrincontextlearningevolving} LLM4ED; \cite{du2024llm4edlargelanguagemodels} LLM-ODE \cite{bideh2026llmode} &
LLM-guided evolutionary operators &
Generic, prompt-embedded &
\xmark &
\xmark &
\cmark &
\xmark &
\cmark \\
\addlinespace

LLM-SR; \cite{shojaee2025llmsrscientificequationdiscovery} LLM-DMD; \cite{shen2026llmdmdlargelanguagemodelbased} LLM-guided PDE discovery \cite{ivanchik2025does} &
LLM generates executable equation/code skeletons (not final closed-form equations); parameters optimised separately &
Broad scientific priors (LLM-SR); none stated (LLM-DMD); LLM as plausibility oracle for equation form (Ivanchik) &
\xmark &
\xmark &
\cmark &
\xmark &
\cmark \\

G-Sim; \cite{NEURIPS2024_aea8bdc4} LEMMA \cite{https://doi.org/10.1111/2041-210x.70244} &
LLM proposes/refines simulator structure or equations; empirical calibration (likelihood-/gradient-free for G-Sim; TMB + evolutionary optimisation for LEMMA) &
Yes: G-Sim uses domain knowledge to guide causal structure; LEMMA uses RAG over ecological literature for realistic parameterisation &
\xmark &
\xmark &
\cmark &
\xmark &
\cmark \\
\addlinespace

ModelSMC \cite{wahl2026probabilisticframeworkllmbasedmodel} &
Probabilistic inference (SMC) &
Implicit via LLM priors &
\xmark &
\xmark &
\cmark &
\xmark &
\cmark \\
\addlinespace

PiSR; \cite{Taskin2026} LaSR; \cite{grayeli2024symbolicregressionlearnedconcept} IGSR \cite{saveliev2026influenceguidedsymbolicregressionscientific} &
LLM as evaluator only, scored via loss term (PiSR, no equation proposal); concept abstraction (LaSR); per-term influence pruning via MCTS (IGSR) &
PiSR: explicit, generic physical criteria; LaSR/IGSR: generic/statistical &
Partial &
\xmark &
\cmark &
\xmark &
\cmark \\

\midrule
\textbf{This work} &
LLM embedded within SR--parameter estimation--AIC selection loop: generates candidate rate expressions + qualitative plausibility assessment &
Yes, explicit chemical/physicochemical reasoning (reaction context, parameter magnitudes, kinetic theory consistency) &
\cmark &
\cmark &
\cmark &
\cmark &
\cmark \\
\bottomrule
\end{tabular}
\begin{tablenotes}
\footnotesize
\item[\textit{a}] Applies to DoE-SINDy only.
\item[\textit{b}] DF-SINDy and DoE-SINDy applied to reaction kinetics; original SINDy is not domain-specific.
\item[\textit{c}] Rogers et al.\ apply to formulated products/PFD optimisation; Kay \& Zhang apply to reaction kinetics.
\end{tablenotes}
\end{threeparttable}
\end{table*}

\clearpage
\newpage
\subsection*{S6. Detailed Prompts}

\subsubsection*{S6.1. Model Critique Prompt}
\label{subsec:analysis}

\begin{Verbatim}[
    breaklines=true,
    breakanywhere=true,
    fontsize=\small,
    frame=single
]

    You are an expert in chemical reaction kinetics. Your job is to analyze kinetic equations for a given global reaction.

    * INSTRUCTIONS:

    - You will receive the following inputs: 
        1) The global reaction.
        2) Products and reactants names (if known).
        3) The current 5 best candidate rate laws found by Symbolic Regression, with their AIC value.

    * TASK:
    - Explain why these kinetic equations could be plausible for their global reaction.
    - Your answer should also include the kinetic equations and their AIC value.
    - Consider the experimental conditions and the chemical nature of the reactants and products in your analysis.
    - Consider also the magnitude of the parameters (e.g., if a term has a very little influence on the rate, it could be negligible and the equation could be simplified by removing it).

    * INPUTS:
    1) Global reaction: {info["balanced_reaction"]}.
    2) Products names: ({info["products"]}). Reactants names: ({info["reactants"]}).
    3) Current best candidate rate laws: {info["best_candidates"]}.
    4) Additional information: {info["additional_info"]}.

    * EXAMPLE:
    BEGINNING OF EXAMPLE
    1) Global reaction: CO + 1/2 O2 → CO2.

    2) Products names: ("Carbon dioxide (CO2): CO2").
       Reactants names: ("Carbon monoxide (CO): CO", "Oxygen (O2): O2").

    3) Current best candidate rate laws (AIC in parentheses):
    - (k1 * KCO * CO * KO2 * O2) / (1 + KCO * CO + KO2 * O2)^2          (AIC = 95.2)
    - (k2 * CO * O2) / (1 + KCO * CO)                                   (AIC = 101.8)
    - k3 * CO * O2                                                      (AIC = 128.4)
    - (k4 * CO) / (1 + KCO * CO)                                        (AIC = 110.6)
    - k5 * CO^2 * O2                                                    (AIC = 140.3)

    4) Additional information: Reaction carried out over a solid metal catalyst at constant temperature.

    Answer:

    The rate law r = (k1 * KCO * CO * KO2 * O2) / (1 + KCO * CO + KO2 * O2)^2 (AIC = 95.2) is highly plausible. This expression is characteristic of a Langmuir-Hinshelwood mechanism in which both CO and O2 adsorb onto the catalyst surface before reacting. The denominator accounts for competitive adsorption and site saturation effects. The squared term suggests that two surface sites are involved in the rate-determining step. Its lowest AIC value strongly supports this model statistically and mechanistically.
    The expression r = (k2 * CO * O2) / (1 + KCO * CO) (AIC = 101.8) is also mechanistically reasonable. It implies that CO adsorption significantly affects surface coverage, while oxygen adsorption may be less limiting. This could correspond to a scenario where O2 dissociation is rapid and surface coverage by CO controls the rate.
    The simple power-law model r = k3 * CO * O2 (AIC = 128.4) assumes elementary bimolecular kinetics without adsorption limitations. While this form may approximate behavior at low surface coverage, it neglects catalytic surface saturation effects and is therefore less physically representative of heterogeneous catalysis.
    The rate law r = (k4 * CO) / (1 + KCO * CO) (AIC = 110.6) suggests that oxygen is present in excess and that CO adsorption limits the rate. This could be plausible under oxygen-rich conditions, but it does not explicitly account for oxygen participation in the surface reaction step.
    Finally, r = k5 * CO^2 * O2 (AIC = 140.3) is the least plausible both statistically and mechanistically. A second-order dependence on CO would imply simultaneous interaction of two CO molecules in the rate-determining step, which is unlikely for this catalytic oxidation mechanism.
    Overall, the Langmuir-Hinshelwood-type expression provides the best balance between statistical support (lowest AIC) and mechanistic consistency with surface-mediated catalysis.
    END OF EXAMPLE

    * NOW, ANALYZE THE KINETIC EQUATIONS:
\end{Verbatim}

\subsubsection*{S6.2. Model Proposal Prompt}

\begin{Verbatim}[
    breaklines=true,
    breakanywhere=true,
    fontsize=\small,
    frame=single
]

You are an expert in chemical reaction kinetics. Your job is to propose plausible kinetic equations for a given global reaction.

    * INSTRUCTIONS:

    - You will receive the following inputs: 
        1) The global reaction.
        2) Products and reactants names (if known).
        3) The current 5 best candidate rate laws found by Symbolic Regression, with their AIC value and an explanation of why they could be plausible for the global reaction.

    * TASK:
    - Propose three kinetic equations for the following global reaction, considering the available experimental data, the current best candidates, and your chemical & mathematical knowledge.
    - The current candidates have valuable information about the reaction mechanism and the influence of each species on the rate, but they are not perfect. 
    - The proposed equations CANNOT be the same as any of the current best candidates. Consider that you can include new terms and parameters, remove existing terms, or change the exponent of the existing terms.
    - The proposed equations do not need to be more complex than the current candidates, but they need to be mechanistically explainable. The new expressions may modify, remove, or extend existing terms, or introduce new functional forms.
    - All exponents must be a fixed number (e.g., 0.5, 1, 2, etc.) and cannot be a free parameter to optimize.
    - Give a brief explanation of the proposed kinetic equation.
    - It is not necessary to provide numbers for the parameters, but the structure of the equation must be clear (e.g., if you p2ropose a term with a parameter, write it as k1, k2, etc.).

    * INPUTS:
    1) Global reaction: {info["balanced_reaction"]}.
    2) Products names: ({info["products"]}). Reactants names: ({info["reactants"]}).
    3) Analysis of current best candidate rate laws: {info["best_candidates_analysis"]}.

    * EXAMPLE:
    BEGINNING OF EXAMPLE
    1) Global reaction: CO + 1/2 O2 → CO2.

    2) Products names: ("Carbon dioxide (CO2): CO2").
       Reactants names: ("Carbon monoxide (CO): CO", "Oxygen (O2): O2").

    3) Analysis of current best candidate rate laws:
    - (k1 * KCO * CO * KO2 * O2) / (1 + KCO * CO + KO2 * O2)^2  (AIC = 95.2)  -> Plausible Langmuir-Hinshelwood mechanism with competitive adsorption.

    - (k2 * CO * O2) / (1 + KCO * CO)  (AIC = 101.8)  → CO adsorption controls surface coverage.

    - k3 * CO * O2  (AIC = 128.4)  → Simple bimolecular power-law behavior.

    - (k4 * CO) / (1 + KCO * CO)  (AIC = 110.6)  → Oxygen in excess, CO adsorption limiting.

    - k5 * CO^2 * O2  (AIC = 140.3)  → Second-order in CO, less mechanistically justified.

    Answer:

    A plausible new kinetic equation is r = (k6 * KCO * CO * (KO2 * O2)^(1/2)) / (1 + KCO * CO + KO2 * O2)

    This expression differs from all previous candidates because:
    - The denominator is first order (not squared).
    - Oxygen appears with a fractional exponent (1/2).
    - Both adsorption terms are retained but modified.

    Mechanistically, this rate law could correspond to a scenario in which oxygen dissociatively adsorbs on the catalyst surface, generating two adsorbed oxygen atoms per O2 molecule. The square-root dependence on O2 reflects equilibrium dissociation before the rate-determining surface reaction. 
    The single-power denominator suggests moderate surface coverage where saturation effects exist but do not require a squared site balance term. This could represent a regime where only one surface site is involved in the rate-determining step, while competitive adsorption still influences the kinetics.
    Thus, this new equation integrates features of Langmuir-Hinshelwood behavior while introducing dissociative oxygen adsorption, making it mechanistically distinct from the previous candidates.
    END OF EXAMPLE

    * NOW, GENERATE THE NEW KINETIC EQUATION:

\end{Verbatim}

\newpage

\subsubsection*{S6.3. LLM-Only Discovery: Seeding Prompt (Iteration 1) -- Without Data}

\begin{Verbatim}[
    breaklines=true,
    breakanywhere=true,
    fontsize=\small,
    frame=single
]

    You are an expert in (bio)chemical reaction kinetics. Your job is to propose plausible kinetic equations for a given global reaction.

    * INSTRUCTIONS:
    - You will receive the following inputs: 
        1) The global reaction.
        2) Products and reactants names (if known).
        3) Additional information.

    * TASK:
    - Propose five kinetic equations for the biomass change in time following global reaction, considering your (bio)chemical & mathematical knowledge.
    - Ensure all equations are (bio)chemically plausible and structurally consistent with known kinetic modeling forms.
    - Briefly justify each proposed equation.
    - Do not provide numerical parameter values; use symbolic parameters (k1, k2, etc.).
    - Prefer algebraic (power-law or rational) functional forms; avoid logarithmic or exponential expressions unless strongly justified.
    - Consider whether additive or baseline terms may be required to capture limiting behavior or low-concentration regimes.

    * INPUTS:
    1) Global reaction: {info["balanced_reaction"]}.
    2) Products names: ({info["products"]}). Reactants names: ({info["reactants"]}).
    3) Additional information: {info["additional_info"]}.

    * EXAMPLE:
    The following example illustrates the expected output structure and level of detail. It is NOT intended to define the type of mechanism or domain-specific chemistry to be used in the actual problem.
    BEGINNING OF EXAMPLE
    1) Global reaction: CO + 1/2 O2 → CO2.

    2) Products names: ("Carbon dioxide (CO2): CO2").
       Reactants names: ("Carbon monoxide (CO): CO", "Oxygen (O2): O2").

    Answer:
    A plausible new kinetic equation is r = (k6 * KCO * CO * (KO2 * O2)^(1/2)) / (1 + KCO * CO + KO2 * O2)
    This expression differs from all previous candidates because:
    - The denominator is first order (not squared).
    - Oxygen appears with a fractional exponent (1/2).
    - Both adsorption terms are retained but modified.
    Mechanistically, this rate law could correspond to a scenario in which oxygen dissociatively adsorbs on the catalyst surface, generating two adsorbed oxygen atoms per O2 molecule. The square-root dependence on O2 reflects equilibrium dissociation before the rate-determining surface reaction. 
    The single-power denominator suggests moderate surface coverage where saturation effects exist but do not require a squared site balance term. This could represent a regime where only one surface site is involved in the rate-determining step, while competitive adsorption still influences the kinetics.
    Thus, this new equation integrates features of Langmuir-Hinshelwood behavior while introducing dissociative oxygen adsorption, making it mechanistically distinct from the previous candidates.
    END OF EXAMPLE

    * In the actual task, mechanisms and functional forms must be consistent with (bio)chemical and (bio)process kinetics.

    * NOW, GENERATE THE NEW 5 KINETIC EQUATIONS:
\end{Verbatim}

\newpage
\subsubsection*{S6.4. LLM-Only Discovery: Seeding Prompt (Iteration 1) -- With Data}

\begin{Verbatim}[
    breaklines=true,
    breakanywhere=true,
    fontsize=\small,
    frame=single
]

    You are an expert in (bio)chemical reaction kinetics. Your job is to propose plausible kinetic equations for a given global reaction.

    * INSTRUCTIONS:
    - You will receive the following inputs: 
        1) The global reaction.
        2) Products and reactants names (if known).
        3) Experimental data.
        4) Additional information.

    * TASK:
    - Propose five kinetic equations for the biomass change in time following global reaction, considering your (bio)chemical & mathematical knowledge.
    - Ensure all equations are (bio)chemically plausible and structurally consistent with known kinetic modeling forms.
    - Briefly justify each proposed equation.
    - Do not provide numerical parameter values; use symbolic parameters (k1, k2, etc.).
    - Prefer algebraic (power-law or rational) functional forms; avoid logarithmic or exponential expressions unless strongly justified.
    - Consider the provided experimental data and assess whether the proposed equations are consistent with the observed trends and system behavior.
    - Consider whether additive or baseline terms may be required to capture limiting behavior or low-concentration regimes.

    * INPUTS:
    1) Global reaction: {info["balanced_reaction"]}.
    2) Products names: ({info["products"]}). Reactants names: ({info["reactants"]}).
    3) Experimental data: {data}.
    4) Additional information: {info["additional_info"]}.
    
    * EXAMPLE:
    The following example illustrates the expected output structure and level of detail. It is NOT intended to define the type of mechanism or domain-specific chemistry to be used in the actual problem.
    BEGINNING OF EXAMPLE
    1) Global reaction: CO + 1/2 O2 → CO2.

    2) Products names: ("Carbon dioxide (CO2): CO2").
       Reactants names: ("Carbon monoxide (CO): CO", "Oxygen (O2): O2").

    Answer:
    A plausible new kinetic equation is r = (k6 * KCO * CO * (KO2 * O2)^(1/2)) / (1 + KCO * CO + KO2 * O2)
    This expression differs from all previous candidates because:
    - The denominator is first order (not squared).
    - Oxygen appears with a fractional exponent (1/2).
    - Both adsorption terms are retained but modified.
    Mechanistically, this rate law could correspond to a scenario in which oxygen dissociatively adsorbs on the catalyst surface, generating two adsorbed oxygen atoms per O2 molecule. The square-root dependence on O2 reflects equilibrium dissociation before the rate-determining surface reaction. 
    The single-power denominator suggests moderate surface coverage where saturation effects exist but do not require a squared site balance term. This could represent a regime where only one surface site is involved in the rate-determining step, while competitive adsorption still influences the kinetics.
    Thus, this new equation integrates features of Langmuir-Hinshelwood behavior while introducing dissociative oxygen adsorption, making it mechanistically distinct from the previous candidates.
    END OF EXAMPLE

    * In the actual task, mechanisms and functional forms must be consistent with (bio)chemical and (bio)process kinetics.

    * NOW, GENERATE THE NEW 5 KINETIC EQUATIONS:
    """
\end{Verbatim}

\subsubsection*{S6.5. LLM-Only Discovery: Subsequent Iterations -- With Data}

For iterations beyond the seeding step, the critique prompt (Step 1) is identical to that used in the full framework (Supplementary Information Section~\ref{subsec:analysis}), receiving only the top-performing candidates' symbolic structure, fitted parameters, and performance metrics. Only the proposal prompt (Step 2) is modified to additionally include the raw concentration-time data for all available experiments, shown below.

\begin{Verbatim}[
    breaklines=true,
    breakanywhere=true,
    fontsize=\small,
    frame=single
]

    You are an expert in (bio)chemical reaction kinetics. Your job is to propose plausible kinetic equations for a given global reaction.

    * INSTRUCTIONS:

    - You will receive the following inputs: 
        1) The global reaction.
        2) Products and reactants names (if known).
        3) Experimental data.
        4) The current 5 best candidate rate laws found by Symbolic Regression, with their AIC value and an explanation of why they could be plausible for the global reaction.

    * TASK:
    - Propose three kinetic equations for the following global reaction, considering the available experimental data, the current best candidates, and your (bio)chemical & mathematical knowledge.
    - The current candidates have valuable information about the reaction mechanism and the influence of each species on the rate, but they are not perfect. 
    - Modifications are allowed if they improve mechanistic consistency or reveal new dynamical behavior; avoid purely algebraic or equivalent reformulations.
    - Ensure all equations are (bio)chemically plausible and structurally consistent with known kinetic modeling forms.
    - Briefly justify each proposed equation.
    - Do not provide numerical parameter values; use symbolic parameters (k1, k2, etc.).
    - Consider the provided experimental data and assess whether the proposed equations are consistent with the observed trends and system behavior.
    - Consider whether the current candidate equations may be missing key structural effects that improve physical consistency, interpretability, or correct limiting behavior.
    - Prefer algebraic (power-law or rational) functional forms; avoid logarithmic or exponential expressions unless strongly justified.
    - Consider whether additive or baseline terms may be required to capture limiting behavior or low-concentration regimes.

    * INPUTS:
    1) Global reaction: {info["balanced_reaction"]}.
    2) Products names: ({info["products"]}). Reactants names: ({info["reactants"]}).
    3) Experimental data: {data}.
    4) Analysis of current best candidate rate laws: {info["best_candidates_analysis"]}.

    * EXAMPLE:
    The following example illustrates the expected output structure and level of detail. It is NOT intended to define the type of mechanism or domain-specific chemistry to be used in the actual problem.
    BEGINNING OF EXAMPLE
    1) Global reaction: CO + 1/2 O2 → CO2.

    2) Products names: ("Carbon dioxide (CO2): CO2").
       Reactants names: ("Carbon monoxide (CO): CO", "Oxygen (O2): O2").

    3) Analysis of current best candidate rate laws:
    - (k1 * KCO * CO * KO2 * O2) / (1 + KCO * CO + KO2 * O2)^2  (AIC = 95.2) -> Plausible Langmuir-Hinshelwood mechanism with competitive adsorption.

    - (k2 * CO * O2) / (1 + KCO * CO)  (AIC = 101.8)  → CO adsorption controls surface coverage.

    - k3 * CO * O2  (AIC = 128.4)  → Simple bimolecular power-law behavior.

    - (k4 * CO) / (1 + KCO * CO)  (AIC = 110.6)  → Oxygen in excess, CO adsorption limiting.

    - k5 * CO^2 * O2  (AIC = 140.3)  → Second-order in CO, less mechanistically justified.

    Answer:

    A plausible new kinetic equation is r = (k6 * KCO * CO * (KO2 * O2)^(1/2)) / (1 + KCO * CO + KO2 * O2)

    This expression differs from all previous candidates because:
    - The denominator is first order (not squared).
    - Oxygen appears with a fractional exponent (1/2).
    - Both adsorption terms are retained but modified.

    Mechanistically, this rate law could correspond to a scenario in which oxygen dissociatively adsorbs on the catalyst surface, generating two adsorbed oxygen atoms per O2 molecule. The square-root dependence on O2 reflects equilibrium dissociation before the rate-determining surface reaction. 
    The single-power denominator suggests moderate surface coverage where saturation effects exist but do not require a squared site balance term. This could represent a regime where only one surface site is involved in the rate-determining step, while competitive adsorption still influences the kinetics.
    Thus, this new equation integrates features of Langmuir-Hinshelwood behavior while introducing dissociative oxygen adsorption, making it mechanistically distinct from the previous candidates.
    END OF EXAMPLE

    In the actual task, mechanisms and functional forms must be consistent with (bio)chemical and bioprocess kinetics.

    * NOW, GENERATE THE NEW KINETIC EQUATION:
\end{Verbatim}

\end{document}